\documentclass{article}

\usepackage{microtype}
\usepackage{graphicx}
\usepackage{subfigure}
\usepackage{float}
\usepackage{hyperref}
\usepackage{fancyhdr}
\usepackage{booktabs} 
\usepackage{hyperref}

\usepackage[accepted]{icml2025}

\usepackage{amsmath}
\usepackage{amssymb}
\usepackage{placeins}
\usepackage{mathtools}
\usepackage{amsthm}
\usepackage{multirow}
\usepackage{bbding}
\usepackage[capitalize,noabbrev]{cleveref}

\theoremstyle{plain}

\theoremstyle{definition}

\theoremstyle{remark}

\newcommand{\markerlessfootnote}[1]{\begingroup
\renewcommand\thefootnote{}\footnote{#1}
\addtocounter{footnote}{-1}
\endgroup
}
\usepackage[textsize=tiny]{todonotes}

\begin{document}

\twocolumn[
\icmltitle{DeblurDiff: Real-World Image Deblurring with Generative Diffusion Models}



\icmlsetsymbol{inter}{*}

\begin{icmlauthorlist}
\icmlauthor{Lingshun Kong}{NJUST,Sense,inter}
\icmlauthor{Jiawei Zhang}{Sense}
\icmlauthor{Dongqing Zou}{Sense,PBVR}
\icmlauthor{Jimmy Ren}{Sense}
\icmlauthor{Xiaohe Wu}{Hit}
\icmlauthor{Jiangxin Dong}{NJUST}
\icmlauthor{Jinshan Pan}{NJUST} \\$^1$ Nanjing University of Science and Technology
$^2$ SenseTime Research
$^3$ PBVR
$^4$ Harbin Institute of Technology

\end{icmlauthorlist}

\icmlaffiliation{NJUST}{Nanjing University of Science and Technology}
\icmlaffiliation{Sense}{SenseTime Research}
\icmlaffiliation{Hit}{Harbin Institute of Technology}
\icmlaffiliation{PBVR}{PBVR}

\icmlkeywords{Machine Learning}

\vskip 0.3in
]




\begin{abstract}
Diffusion models have achieved significant progress in image generation.
The pre-trained Stable Diffusion (SD) models are helpful for image deblurring by providing clear image priors.
However, directly using a blurry image or pre-deblurred one as a conditional control for SD will either hinder accurate structure extraction or make the results overly dependent on the deblurring network.
In this work, we propose a Latent Kernel Prediction Network (LKPN) to achieve robust real-world image deblurring.
Specifically, we co-train the LKPN in latent space with conditional diffusion.
The LKPN learns a spatially variant kernel to guide the restoration of sharp images in the latent space.
By applying element-wise adaptive convolution (EAC), the learned kernel is utilized to adaptively process the input feature, effectively preserving the structural information of the input.
This process thereby more effectively guides the generative process of Stable Diffusion (SD), enhancing both the deblurring efficacy and the quality of detail reconstruction.
Moreover, the results at each diffusion step are utilized to iteratively estimate the kernels in LKPN to better restore the sharp latent by EAC.
This iterative refinement enhances the accuracy and robustness of the deblurring process.
Extensive experimental results demonstrate that the proposed method outperforms state-of-the-art image deblurring methods on both benchmark and real-world images.
The codes will be released in: \textit{\url{https://github.com/kkkls/DeblurDiff}}.

\end{abstract}

\markerlessfootnote{* This work was done during an intern at Sensetime Research.}

\section{Introduction}
\label{introduction}
Image deblurring aims to recover a sharp image from a blurry observation.
Blurring can be caused by various factors, such as camera shake and high-speed movement of the photographed objects.
%
This task is challenging as only the blurry images are available and the blur might be non-uniform.
Traditional deblurring methods~\cite{L0,dark_channel} have made significant progress by utilizing hand-crafted features and priors.  
However, these methods often struggle to handle complex blur patterns and may produce unsatisfactory results.
\begin{figure*}[!t]
\footnotesize
\centering
\vspace{-2mm}

    \begin{tabular}{c c c c c c c}
            \multicolumn{3}{c}{\multirow{5}*[45.6pt]{
            \hspace{-4mm} 
            \vspace{-25mm}
            \includegraphics[width=0.325\linewidth,height=0.236\linewidth]{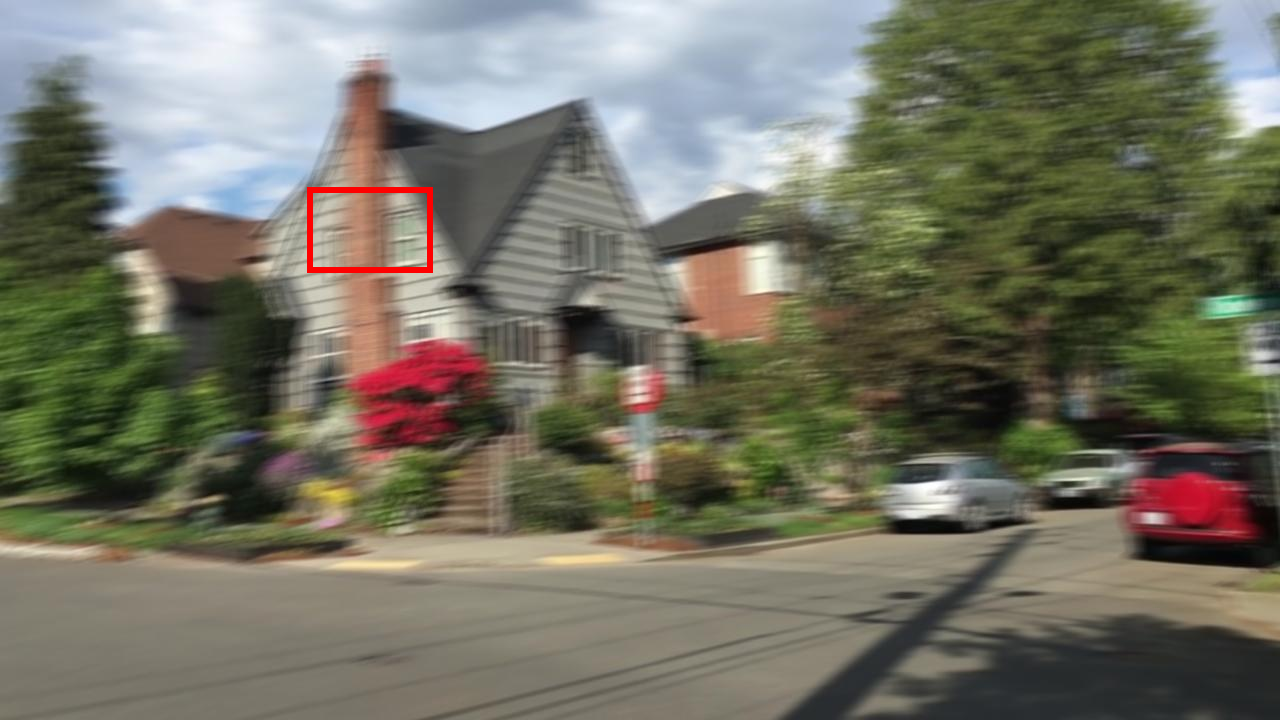}}}
            & \hspace{-4.0mm} \includegraphics[width=0.16\linewidth,height=0.105\linewidth]{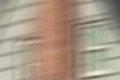}
            & \hspace{-4.0mm} \includegraphics[width=0.16\linewidth,height=0.105\linewidth]{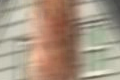}
            & \hspace{-4.0mm} \includegraphics[width=0.16\linewidth,height=0.105\linewidth]{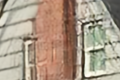}
            & \hspace{-4.0mm} \includegraphics[width=0.16\linewidth,height=0.105\linewidth]{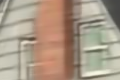}
              \\
    		\multicolumn{3}{c}{~}
            & \hspace{-4.0mm} (a) Blurred patch
            & \hspace{-4.0mm} (b) DBGAN
            & \hspace{-4.0mm} (c) FFTformer
            & \hspace{-4.0mm} (d) HI-Diff \\		
    	\multicolumn{3}{c}{~}
            & \hspace{-4.0mm} \includegraphics[width=0.16\linewidth,height=0.105\linewidth]{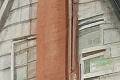}
            & \hspace{-4.0mm} \includegraphics[width=0.16\linewidth,height=0.105\linewidth]{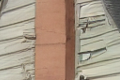}
            & \hspace{-4.0mm} \includegraphics[width=0.16\linewidth,height=0.105\linewidth]{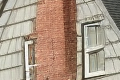}
            & \hspace{-4.0mm} \includegraphics[width=0.16\linewidth,height=0.105\linewidth]{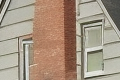}
            \\
    	\multicolumn{3}{c}{\hspace{-4.0mm} Blurred image from DVD dataset}
            & \hspace{-4.0mm} (e) ControlNet
            & \hspace{-4.0mm} (f) PASD
            & \hspace{-4.0mm} (g) DiffBIR
            & \hspace{-4.0mm} (h) Ours\\

    \end{tabular}
\vspace{-4mm}
\caption{Visual comparison with state-of-the-art image deblurring methods. 
The results of GAN-based method (b) and diffusion-based method without pretraining (d) still contain significant blur effects.
Directly using the blurry image as the conditional input (e) presents significant challenges in effectively extracting structural information. 
(f) is a method based on pre-trained SD that performs pre-deblurring on the input features, which alters the original information, leading to erratic generation.
For (g), it uses the result of the pre-trained FFTformer (c) as the condition. 
(g) is influenced by the erroneous structures in (c), resulting in generated outputs that retain erroneous artifacts and erroneous structures.
In contrast, our approach, guided by the clear structural information provided by LKPN, generates a more distinct and artifact-free image.}
\label{fig: Intro_vis}
\vspace{-6mm}
\end{figure*}
Recently, numerous learning-based approaches~\cite{SRN,MIMO,MPRNet,NAFNet} have been inclined to employ a variety of convolutional neural network (CNN) architectures.
Compared with traditional algorithms, CNNs have demonstrated remarkable success.
However, convolution operation is a spatially invariant local operation, which cannot effectively model the spatially variant characteristics of image content and the global contexts for image deblurring.
STAFN~\cite{stfan} employs a Kernel Prediction Network (KPN) to estimate a spatially variant kernel to assist with deblurring.
To address the limitations of CNNs, Transformers~\cite{Restormer,Uformer,GRL,fftformer} have been increasingly applied to image deblurring and have achieved commendable performance.
However, the above regression-based methods, which often use L2 and L1 reconstruction losses, tend to predict the mean rather than a plausible high-resolution (HR) solution~\cite{SRFlow}. 
This results in overly smooth outputs, leading to a significant loss of detail.
To address this problem, recent methods~\cite{DeblurGAN,DeblurGANv2,GAN_SR,physicgan} have adopted adversarial training and perceptual loss functions to generate more detailed results.
However, generative adversarial networks (GANs) suffer from training instability, mode collapse, and limited diversity in complex data generation, which may hinder the plausibility of the generated images.
Recently, diffusion models have demonstrated outstanding performance in image generation~\cite{DDPM,diffusion1}.
Some researchers have attempted to utilize Denoising Diffusion Probabilistic Models (DDPMs) for image restoration~\cite{cdm,resshift}, aiming to leverage their ability to capture complex data distributions and generate detailed images.
However, due to the lack of large-scale pre-training, these methods have not demonstrated satisfactory results, particularly when applied to out-of-distribution data, often leading to inferior performance and visually unpleasant artifacts.
With the surprising performance of large-scale pre-trained models like Stable Diffusion (SD) in image generation, it has been developed for image restoration~\cite{diffbir,pasd}.
SD models have strong priors on the structure and details of high-quality (HQ) images. 
%
%
However, directly utilizing blurry images as conditional inputs can hinder the extraction of effective structural information, especially in cases of severe blur, ultimately resulting in inaccurately generated structures (Figure~\ref{fig: Intro_vis}(e)).
While recent methods attempt to mitigate these limitations, their technical trajectories introduce new issues.
These methods~\cite{diffbir,pasd} require training an additional Degradation Removal Model (DRM) to first restore the clear images, and then enhance the details using a ControlNet~\cite{controlnet}.
%
%
This suggests that the restoration results of the entire method are notably influenced by the results of the degradation removal model. 
When the DRM(implemented via FFTformer~\cite{fftformer}) yields inaccurate results (Figure~\ref{fig: Intro_vis} (c)), it may adversely affect the diffusion process, potentially resulting in suboptimal performance (Figure~\ref{fig: Intro_vis} (g)).
Moreover, due to the poor generalization of existing degradation removal models across different datasets, these methods also tend to perform poorly when dealing with various types of blur.
%



%
In this paper, we investigate in-depth the problem of how to leverage pre-trained SD models to assist in real-world image deblurring while reconstructing realistic details and textures. 
This approach circumvents the direct use of blurry images as conditions, as severe blur can hinder the extraction of accurate structural information, leading to suboptimal final generated structures.
Unlike pre-training an additional degradation removal model, which uses restored images as conditions and can be problematic due to poor restoration quality introducing incorrect structures and resulting in erroneous generation, we jointly train a Latent Kernel Prediction Network (LKPN) with the diffusion model.
The LKPN, together with EAC, is designed to guide the conditional generation at each step of the diffusion process.
%

%
%
The effectiveness of the LKPN lies in its ability to predict dedicated convolution kernels for each latent pixel, dynamically adjusting the kernel weights based on local content (e.g., edges, textures, and flat regions). 
%
%
These pixel-specific kernels are then applied to the latent blurry image by the element-wise adaptive convolution (EAC), enabling better restoration of clear structures by adaptively addressing distinct blur characteristics at each latent pixel location. 
This adaptive mechanism not only preserves the necessary information in the input image but also avoids the destruction of structural details, making it particularly advantageous for recovering fine structures and textures.
By integrating the LKPN, the diffusion model uses the clear structures from the LKPN as conditional inputs to guide its generation process at each step. This produces more accurate results and enhances both deblurring and detail reconstruction.
Furthermore, the intermediate results obtained at each step of the diffusion process are utilized to refine the output of the LKPN, progressively improving the accuracy of the deblurred results by EAC. 
This iterative refinement generates increasingly clearer conditional inputs, which in turn guide the generation process of the Stable Diffusion (SD) model to achieve iterative improvements.
%
Additionally, the LKPN continuously benefits from the strong prior structures and detailed information provided by SD, enabling it to estimate better kernels to remove blur in latent space.
%
%

%
The main contributions are summarized as follows:
\begin{itemize}
    \item  We are the first to design an architecture based on pre-trained SD models to achieve effective deblurring while reconstructing realistic details and textures.

    \item We propose an LKPN architecture that estimates a spatially variant kernel, which is then utilized by the EAC to progressively generate clear structures and preserve the necessary information in the input image throughout the diffusion process. This guides SD to produce more accurate details and structures, thereby enhancing the overall deblurring performance.

    \item We quantitatively and qualitatively evaluate the proposed method on benchmark datasets and real-world images and show that our method outperforms state-of-the-art methods.
\end{itemize}

\begin{figure*}[!t]
\footnotesize
\centering
\vspace{-2mm}
\begin{tabular}{c|c|cc}

\includegraphics[width = 0.1\linewidth,height=0.1\linewidth]{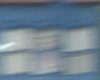}& 
\includegraphics[height=0.1\linewidth]{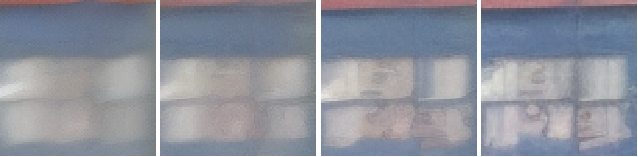}& 
\includegraphics[height=0.1\linewidth]{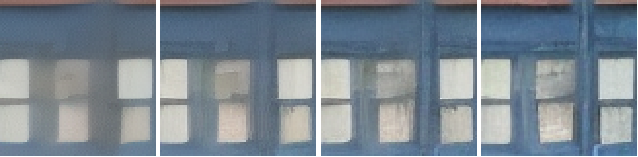} & \\
(a) Input&(c) ControlNet&(e) Deblurred image by LKPN and EAC\\
\includegraphics[width = 0.1\linewidth,height=0.1\linewidth]{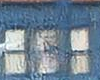}& 
\includegraphics[height=0.1\linewidth]{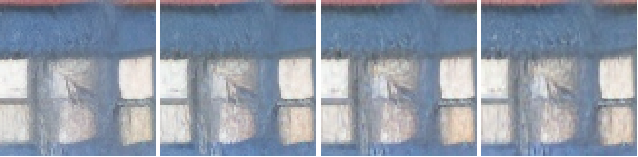}& 
\includegraphics[height=0.1\linewidth]{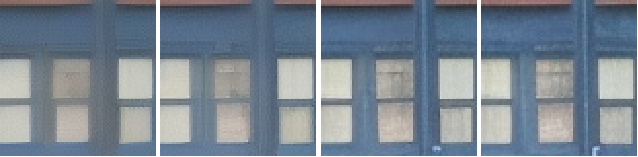} & \\
(b) FFTformer&(d) DiffBIR&(f) DeblurDiff\\

\vspace{-6mm}
\end{tabular}
\vspace{-6mm}
\hspace{17mm}
\includegraphics[width = 0.82\linewidth]{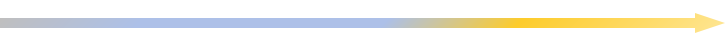}\hspace{10mm} \\

\caption{Iterative results of the diffusion model. The arrow represents the iterative diffusion process. To visualize this process, we decode the features deblurred by the LKPN and EAC to the image space using the VAE decoder in each time step. Using the blurry image directly as a conditional input (c) makes the diffusion model struggle to recover clear structures and fine details in (a). 
For (d), it uses the result of the pre-trained FFTformer (b) as the condition. However, the deblurring network can introduce incorrect structures, leading to erroneous content generation.
In contrast, the proposed LKPN can preserve the input information and restore the structure (e) by EAC, thereby guiding the diffusion model to generate better results (f).}
\label{fig: mov}
\vspace{-3mm}
\end{figure*}

\section{Related Work}
\label{Related Work}
{\flushleft \textbf{Image deblurring.}}
Due to the fact that image deblurring is an ill-posed problem, traditional methods~\cite{sparse,L0,dark_channel} often develop various effective priors to constrain the solution space.
These manually designed priors can help remove blur. 
However, they do not fully utilize the characteristics of clear image data, which leads to a struggle in handling complex blur patterns and may produce unsatisfactory results
%

%
With the development of deep learning, many learning-based methods have tended to use various CNN architectures for image deblurring. 
SRN~\cite{SRN} proposes a multi-scale structure that performs image deblurring from coarse to fine.
MIMOUnet~\cite{MIMO} redesigns the coarse-to-fine structure, significantly reducing the computational cost.
NAFNet~\cite{NAFNet} analyzes the baseline module and simplifies it by removing the activation function, which better facilitates image restoration.
Due to the excellent performance of Transformers in global context exploration and their great potential in many visual tasks, some methods have applied it to image deblurring.
Restormer~\cite{Restormer} simplifies the baseline module by estimating self-attention in the channel dimension, reducing the computational cost of self-attention
Uformer~\cite{Uformer} proposes a general U-shaped Transformer model, computing self-attention within local windows to address the image deblurring.
FFTformer~\cite{fftformer} proposes a frequency-domain based Transformer model and achieved state-of-the-art results.
Although these methods have achieved good deblurring effects, these regression-based methods tend to predict smooth results, with limited ability to depict details.

{\flushleft \textbf{Diffusion model.}}
Denoising Diffusion Probabilistic Models (DDPM)~\cite{DDPM} have shown remarkable capabilities in generating high-quality natural images.
Some methods~\cite{resshift,cdm,hidiff} have attempted to directly train a diffusion model for image restoration.
Rombach et al.~\cite{sd} extended the DDPM structure to the latent space and conducted large-scale pre-training, demonstrating impressive generative capabilities.
Recently, some researchers have utilized powerful pre-trained generative models, such as SD~\cite{sd}, to address image restoration problems.
DiffBIR~\cite{diffbir} proposes a two-stage approach, first restoring the degraded image and then using SD to generate details.
PASD~\cite{pasd} restores clear images through a Degradation Removal module to provide clear conditional inputs for SD.
However, these methods require training an additional image restoration model and then enhance the details through SD. 
This means that the final results of the SD largely depend on the outcomes of the restoration model. 
When the degradation removal model produces erroneous results, it may lead to poor performance in the diffusion process. 
Moreover, due to the poor generalization of existing restoration models, these methods also tend to perform poorly when dealing with various types of blur.
%


\section{Method}
\begin{figure*}[!t]
    \centering
     \vspace{-1mm}
 \includegraphics[width=0.98\textwidth]{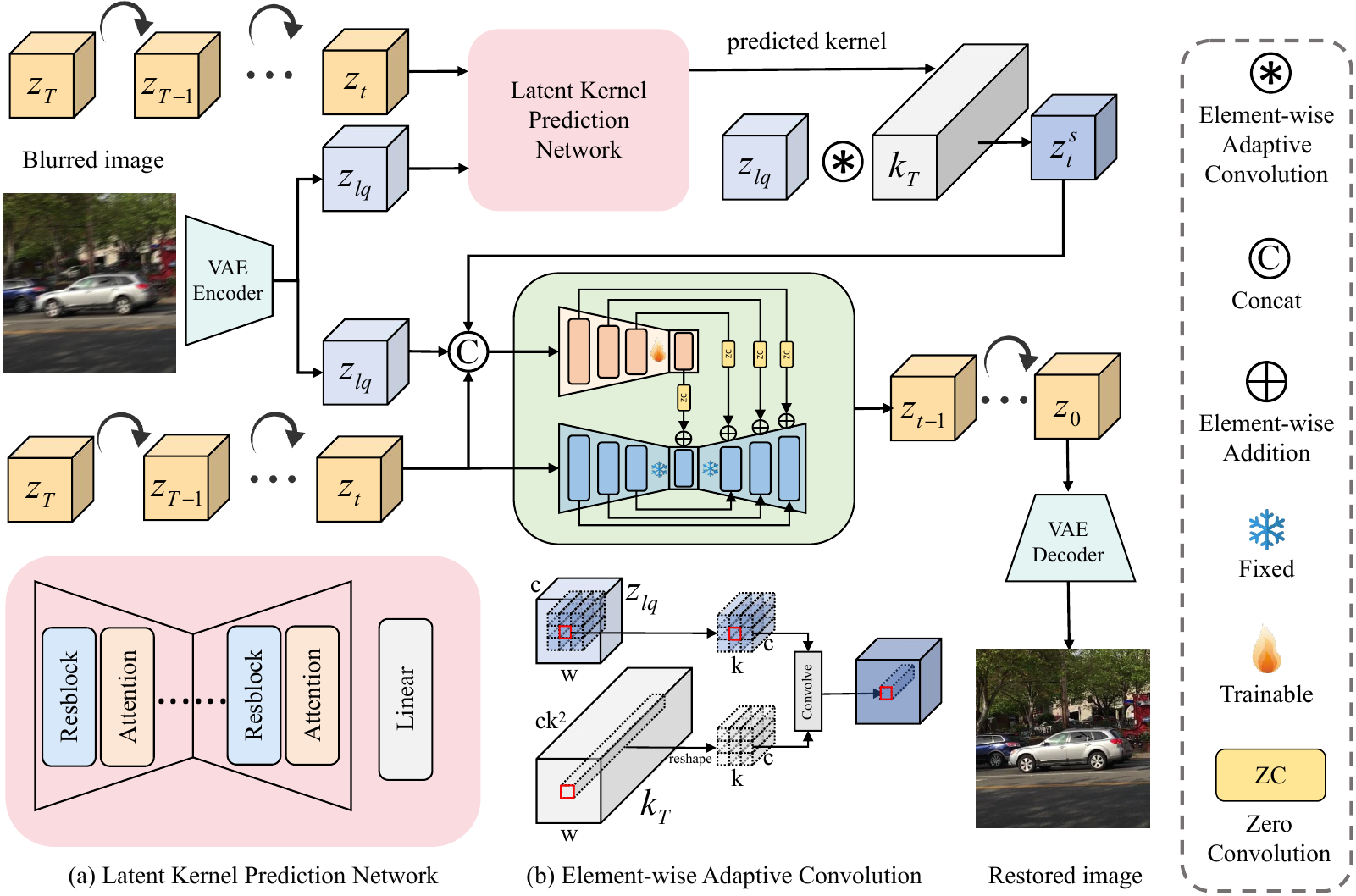}
 \vspace{-4mm}
 \caption{Overall architecture of the proposed DeblurDiff. It integrates a Latent Kernel Prediction Network (LKPN) with a generative diffusion model to address the challenges of real-world image deblurring. The LKPN progressively recovers clear structures from blurred images by estimating pixel-specific deblurring kernels at each step of the diffusion process. These kernels are adaptively adjusted based on local content and applied through Element-wise Adaptive Convolution (EAC). The refined clear $z^s$ is used as a condition to guide the diffusion process, enabling the model to effectively preserve the input information and structural integrity.}
 \vspace{-6mm}
 \label{fig: Network}
\end{figure*}

Our goal is to leverage the powerful priors of pre-trained Stable Diffusion models to achieve robust image deblurring, while simultaneously reconstructing realistic details and textures.
Specifically, we introduce a framework that integrates a Latent Kernel Prediction Network (LKPN) with the diffusion process. 
The LKPN estimates pixel-specific kernels in the latent space, enabling adaptive deblurring tailored to the distinct characteristics of each pixel. 
By iteratively refining the deblurred results throughout the diffusion process, our approach progressively restores clear structures and enhances fine details. 
Figure~\ref{fig: Network} shows the overview of the proposed method.
\subsection{Motivation and Preliminaries}
\textbf{Motivation.}
%
%
%
Directly using the blurry image as as the condition of SD leads to suboptimal deblurring, as it lacks sufficient structural information to effectively guide the generation process, making it difficult for diffusion models to recover clear structures and details as shown in Figure~\ref{fig: mov}(c).
When using a pre-trained network to process the blurry image, if the deblurring result is inaccurate, it can introduce incorrect structures (Figure~\ref{fig: mov}(b)). 
This, in turn, leads to the diffusion model generating erroneous structures during its iterative process (Figure~\ref{fig: mov}(d)).
In this paper, we utilize the image priors from Stable Diffusion (SD) and the blurry image together to help us predict an accurate clear structure, thereby more accurately guiding the generation process.

Our work is motivated by existing methods~\cite{noise_deblur, noise_deblur_2}, which have explored using paired noisy and blurry images to improve deblurring performance.
However, obtaining paired images is often quite difficult in practical applications, which limits the applicability of these methods.
In contrast, our approach leverages the intermediate noise representations generated during the diffusion process to guide deblurring. 
Our method generates a guidance with clear structures during the diffusion process (Figure~\ref{fig: mov}(e)), enabling the diffusion model to produce clearer and more accurate results while retaining more of the input information as shown in Figure~\ref{fig: mov}(f).

\textbf{Denoising Diffusion Probabilistic Models (DDPM)} are a type of probabilistic model that learn the data distribution $p(x)$ by progressively denoising a normally distributed variable. 
The training process involves a forward noise-adding process and a reverse denoising process. At the $t$-th step of the forward diffusion process, a noisy image $x_t$ is generated from the clear image $x_0$ by:
\begin{equation}
x_t = \sqrt{{\bar{\alpha}}_t} x_0 + \sqrt{1 - {\bar{\alpha}}_t} {\epsilon} , ,
\label{eq: ddpm}
\end{equation}
where $\epsilon$ denotes the noise sampled from the standard normal distribution $\mathcal{N}(0,I)$, and $\bar{\alpha}_t$ controls the amount of noise added at each step $t$, which is typically a predefined value. 
In the reverse denoising process, a denoising neural network $\epsilon_\theta(x_t,t)$ is used to predict the noise in $x_t$. 
After training, the denoising network can iteratively run for $T$ steps starting from noise sampled from a standard normal distribution to generate a clear image. 
To improve computational efficiency, the diffusion process is often transferred from the pixel space to the latent space, where high-frequency, imperceptible details are abstracted away via a pre-trained Variational Autoencoder (VAE). 
This latent space is more suitable for likelihood-based generative models as it focuses on the important and semantic parts of the data, operating in a lower-dimensional and computationally efficient space~\cite{sd}. 
Specifically, a pre-trained VAE is utilized to transform the clear image $x_0$ into a latent representation $z_0$ by $z_0=\mathcal{E}(x_0)$, and the forward noising and reverse denoising processes of DDPM are performed in this latent space. 
The corresponding optimization objective can be simplified to minimizing the denoising loss:
\begin{equation}
\mathcal{L}_{\text{denoise}} = \mathbb{E}_{z_t, \epsilon \sim \mathcal{N}(0, I), t} \left[ \left\| \epsilon - \epsilon_\theta(\mathbf{z}_t, t) \right\|^2_2 \right].
\label{eq: loss}
\end{equation}
where $z_t$ is the noisy latent representation at step $t$.

\begin{table*}[!t]\footnotesize
\vspace{-6mm}

    \caption{Quantitative evaluations of the proposed method against state-of-the-art ones on both synthetic and real-world benchmarks. The models marked with an asterisk $^*$ indicate that we retrain them on our own training set. The best and second performances are marked in \textcolor{red}{red} and \textcolor{blue}{blue}, respectively. For the RWBI and Real Blurry Images datasets, which lack ground truth (GT) data, we evaluate the performance using only no-reference metrics.
    }
    \label{tab:result}
\begin{tabular}{c|c|cc|cc|ccc|cc}
        \toprule
Dataset                      & Metrics             & FFTformer$^*$  & DBGAN      & ResShift & HI-Diff & ControlNet$^*$   & PASD$^*$  & DiffBIR$^*$ & Ours   \\
        \midrule 

\multirow{7}{*}{GoPro}         & PSNR~$\uparrow$     &    26.86     &    \textcolor{blue}{31.18}  &  29.03   &   \textcolor{red}{33.33}  &   22.31          &  22.82    &   23.86     &  24.32      \\
                             & SSIM~$\uparrow$       &    0.8357    &    \textcolor{blue}{0.9182}   &  0.8781  &   \textcolor{red}{0.9462} &   0.6547         &  0.6559   &  0.7173     &  0.7375      \\
                             & LPIPS~$\downarrow$    &    0.1538    &  0.1120    &   \textcolor{red}{0.0780}  &   \textcolor{blue}{0.0820} &   0.3292         &  0.2984   &  0.2772     &  0.2191      \\
                             & NIQE~$\downarrow$     &     4.1200   &   5.1988   & 4.8367   &   {4.6119} &   {3.5305}          &   \textcolor{red}{2.6567}  & {3.4193}      & \textcolor{blue}{3.1769}\\
                             & MUSIQ~$\uparrow$      &     52.2993   &    42.0985 & 44.2820  &  47.7791&  59.4246         &  \textcolor{blue}{61.5345}  &  56.3249    &  \textcolor{red}{61.6369}\\
                             & MANIQA~$\uparrow$     &     0.5454    &  0.4976    & 0.9419   &  0.6119 &  0.5746          &   \textcolor{blue}{0.5904}  & 0.5464      &  \textcolor{red}{0.6134}\\
                             & CLIP-IQA~$\uparrow$   &    0.4360     &    0.3788  &  0.4229  &  0.4841 &  \textcolor{blue}{0.5869}          &   0.5758  &   0.5260    &   \textcolor{red}{0.5966}\\
                                     \midrule 
\multirow{7}{*}{DVD}         & PSNR~$\uparrow$       &    27.07     &     \textcolor{blue}{27.78}   &   \textcolor{blue}{27.78}   &   \textcolor{red}{30.31}  &    22.03         &   22.23   &   23.49     &  23.74     \\
                             & SSIM~$\uparrow$       &    \textcolor{blue}{0.8534}    &    0.8356  &  \textcolor{red}{0.8420}  &  {0.8972} &    0.6409        &  0.6440   &   0.7114    &  0.7055    \\
                             & LPIPS~$\downarrow$    &    0.1628    &    0.2126  & \textcolor{red}{0.1249}   &  \textcolor{blue}{0.1363} &   0.3330         &  0.2968   &   0.2795    &  0.2501    \\
                             & NIQE~$\downarrow$     &    3.8562    &   4.8188   &  4.6422  &  5.1858 &   3.4037         &    3.2504 &   \textcolor{blue}{3.1357}    &  \textcolor{red}{2.7822}\\
                             & MUSIQ~$\uparrow$      &     60.1091  &    40.4781 & 52.7551  & 45.5395 &   65.3657        &  \textcolor{red}{68.3299}  &  61.6415    &  \textcolor{blue}{67.2447}\\
                             & MANIQA~$\uparrow$     &      0.6257  &  0.5548    & 0.5744   & 0.5360  &  0.6155          &  \textcolor{blue}{0.6409}   &  0.5773     &  \textcolor{red}{0.6480}\\
                             & CLIP-IQA~$\uparrow$   &     0.5271   &    0.4346  &  0.4831  & 0.4065  &   0.6606         &  \textcolor{blue}{0.6528}   &   0.5829    &  \textcolor{red}{0.6686}\\
                                     \midrule 
\multirow{7}{*}{Realblur}    & PSNR~$\uparrow$       &    \textcolor{blue}{26.94}          &     23.91  &    26.30 &  \textcolor{red}{30.18}  &     23.77        &   25.02   &    25.50    &  25.71     \\
                             & SSIM~$\uparrow$       &    \textcolor{blue}{0.8580}          &    0.7434  &   0.8140 &  \textcolor{red}{0.9049} &     0.6787       &   0.7642  &    0.7724   &  0.7705     \\
                             & LPIPS~$\downarrow$    &   0.1411           &    0.2945  &   \textcolor{blue}{0.1249} &  \textcolor{red}{0.0868} &    0.2565        &   0.2075  &    0.1951   &  0.1693     \\
                             & NIQE~$\downarrow$     &     4.3473         &  5.3228    &  5.2628  &5.1437   &  4.6525          &   \textcolor{red}{3.9192}  &  {4.3053}     &  \textcolor{blue}{4.2666}\\
                             & MUSIQ~$\uparrow$      &    61.5808          &    38.4866 &  49.3209 & 57.1640 &   \textcolor{red}{66.5174}        &   61.1498 &   58.7450   &  \textcolor{blue}{65.0557}\\
                             & MANIQA~$\uparrow$     &    0.6374          &   0.43222  &  0.5373  & 0.6218  &   \textcolor{blue}{0.6452}         &  0.5994   &  0.5869     &  \textcolor{red}{0.6538}\\
                             & CLIP-IQA~$\uparrow$   &    0.5336         &   0.3469   &  0.4521  &  0.5101 &    \textcolor{blue}{0.6041}        & 0.5457    & 0.5261      &  \textcolor{red}{0.6087}\\

                                     \midrule 

\multirow{4}{*}{RWBI}        & NIQE~$\downarrow$     &    4.4631          &  5.2905    &  5.4446  & 5.3373  & 5.0331           &    \textcolor{red}{4.1973} &  \textcolor{blue}{4.2742}     &   4.5171\\
                             & MUSIQ~$\uparrow$      &    59.6223          &   42.7631  &  51.0359 & 47.1820 &   \textcolor{blue}{62.5079}        &  62.1680  &  61.8865    &   \textcolor{red}{66.7505}\\
                             & MANIQA~$\uparrow$     &    0.5425          &   0.4852   & 0.4953   & 0.5082  &   \textcolor{blue}{0.5758}         &  0.5645   &   0.5618    &   \textcolor{red}{0.6260}\\
                             & CLIP-IQA~$\uparrow$   &    0.5413          &   0.3645   & 0.5032   & 0.3907  &   \textcolor{blue}{0.6199}         & 0.5820    &   0.6042    &   \textcolor{red}{0.6849}\\
                                     \midrule 

\multirow{4}{*}{Real Images} & NIQE~$\downarrow$     &    3.8520          & 4.9338     & 5.4704   &  4.7018 &  4.0978          & 4.4460    &  \textcolor{blue}{3.7964}     &   \textcolor{red}{3.6628}\\
                             & MUSIQ~$\uparrow$      &    52.9290          & 32.0568    & 48.8154  & 43.8702 &   51.5191        & \textcolor{red}{61.6320}   &  \textcolor{blue}{53.6088}    &   52.9263\\
                             & MANIQA~$\uparrow$     &     0.5170         & 0.4488     & 0.5345   &  0.5722 &   0.5544         &  \textcolor{blue}{0.5937}   &    0.5564   &   \textcolor{red}{0.5963}\\
                             & CLIP-IQA~$\uparrow$   &   0.5026           &  0.3501    & 0.4767   &  0.4545 &    0.5254        &  \textcolor{red}{0.5919}   &  0.5384     &   \textcolor{blue}{0.5496}\\
                                     \bottomrule

\end{tabular}
\vspace{-5mm}

\end{table*}

\subsection{LKPN for clear structure guidance.}
In the context of blind deblurring, directly using blurry images as conditions can impede the extraction of effective structural information, especially when dealing with significant blur, which can lead to inaccurate final generated structures. 
To address this challenge, we propose a method that incorporates an LKPN trained simultaneously with the diffusion model.  
The LKPN dynamically estimates pixel-specific deblurring kernels at each step of the diffusion process. 
These kernels are adaptively adjusted based on local content by the element-wise adaptive convolution(EAC), enabling the model to address spatially varying blur patterns effectively in the latent space.
By leveraging intermediate image priors generated during the diffusion process, the EAC can generate clear structural guidance, which is then used as conditional inputs to the diffusion model. 
This iterative refinement allows the LKPN to progressively improve the accuracy of deblurred results by EAC, leading to increasingly clearer intermediate results while preserving the information of the input image.
Specifically, we first employ a pre-trained and frozen VAE encoder initialized from SD to encode the blurry image and the clear one into the latent space, obtaining their corresponding latent representations $z_{lq}$ and $z_0$.
We follow the Eq.(\ref{eq: ddpm}) to add noise to $z_0$, obtaining $z_t$.
Then the LKPN, which is a U-Net architecture, predicts a spatially variant kernel in latent space at step $t$ :
\begin{equation}
k_t = \text{LKPN}(z_t, z_{lq}, t),\\
\label{eq: kpn}
\end{equation}
where $k_t$ is the predicted kernel at time step $t$.
Then $k_t$ is used to refine the blurry image in the latent space by :
\begin{equation}
z^s_t = \text{EAC}(\mathbf{z}_{lq}, k_t),\\
\label{eq: kpn}
\end{equation}
where $\text{EAC}$ is the element-wise adaptive convolution in Figure~\ref{fig: Network}(b) and $z^s_t$ is the deblurred latent. 
In the appendix, we provide a detailed explanation of the LKPN and EAC architecture.

During training, the LKPN is trained simultaneously with the diffusion model. 
The LKPN follows the framework of DDPM, imposing constraints simultaneously in both the latent space and the pixel space to progressively optimize the kernel estimation.
Specifically, our objective is to minimize the following objective:
\begin{equation}
\begin{split}
&\mathcal{L_{\text{LKPN}}} = \mathcal{L}_{latent} + \mathcal{L}_{pixel},\\
&\mathcal{L}_{latent} = \mathbb{E}_{z_0, z_{lq}, k_t}\left[ \left\| z_0 - \text{EAC}(\mathbf{z}_{lq}, k_t) \right\|^2_2 \right],\\
&\mathcal{L}_{pixel} = \mathbb{E}_{z_0, z_{lq}, k_t}\left[ \left\| \mathcal{D}(z_0) - \mathcal{D}(\mathcal{C}(\mathbf{z}_{lq}, k_t)) \right\|^2_2 \right],\\
\label{eq: loss_kpn}
\end{split}
\end{equation}

where $\text{EAC}$ denotes the element-wise adaptive convolution (EAC) in Figure~\ref{fig: Network}(b), and $\mathcal{D}$ denotes the pre-trained VAE decoder of SD.
The network architecture of the proposed LKPN is shown in Figure~\ref{fig: Network}.
%
%
\subsection{Conditional diffusion for image deblurring .}
For the conditional generation network, we follow the training methodology of ControlNet~\cite{controlnet}, given its demonstrated effectiveness in conditional image generation.
Specifically, we adopt the encoder of the UNet in SD as a trainable conditional control branch and initialize the control network by copying weights from the pre-trained SD model.
We concatenate $z^s$ recovered by the EAC and $z_{lq}$ as the input to the ControlNet, which is initialized with the weights of SD. We provide a detailed explanation of the network architecture in the appendix.
%
%
%
During training, the LKPN and ControlNet are jointly trained following the framework of DDPM~\cite{DDPM}, we minimize the following loss function:
\begin{equation}
\mathcal{L} = \mathcal{L}_{\text{denoise}} + \mathcal{L}_{\text{LKPN}}.\\
\label{eq: loss_kpn}
\end{equation}

\subsection{Sampling process of the DeblurDiff.}
During the inference stage, the LKPN first estimates the initial spatially variant kernel from random Gaussian noise and the blurred image, which is used to obtain an initial conditional ${z^s_T}$.
The controlled diffusion model uses ${z^s_T}$ and $z_{lq}$ as conditions to generate an initial clear result ${z_{T-1}}$.
Subsequently, ${z_{T-1}}$ is fed back into the LKPN to help estimate a more accurate kernel:
\begin{equation}
k_{T-1} = \text{LKPN}({z_{T-1}}, z_{lq}, T-1),
\label{eq: kpn_infer}
\end{equation}
thereby iteratively optimizing the generated results in subsequent steps.
Please see the appendix for more details.
%

%
%
This synergy between the LKPN and the diffusion model creates a mutually reinforcing cycle, where clear structural guidance from the LKPN improves the diffusion process, and intermediate results from the diffusion model further refine the deblurred results. 
%
\begin{figure*}[!t]
\footnotesize
 \vspace{-2mm}
\centering
    \begin{tabular}{c c c c c c c}
            \multicolumn{3}{c}{\multirow{5}*[45.6pt]{
            \hspace{-4mm} 
            \vspace{-25mm}
            \includegraphics[width=0.325\linewidth,height=0.236\linewidth]{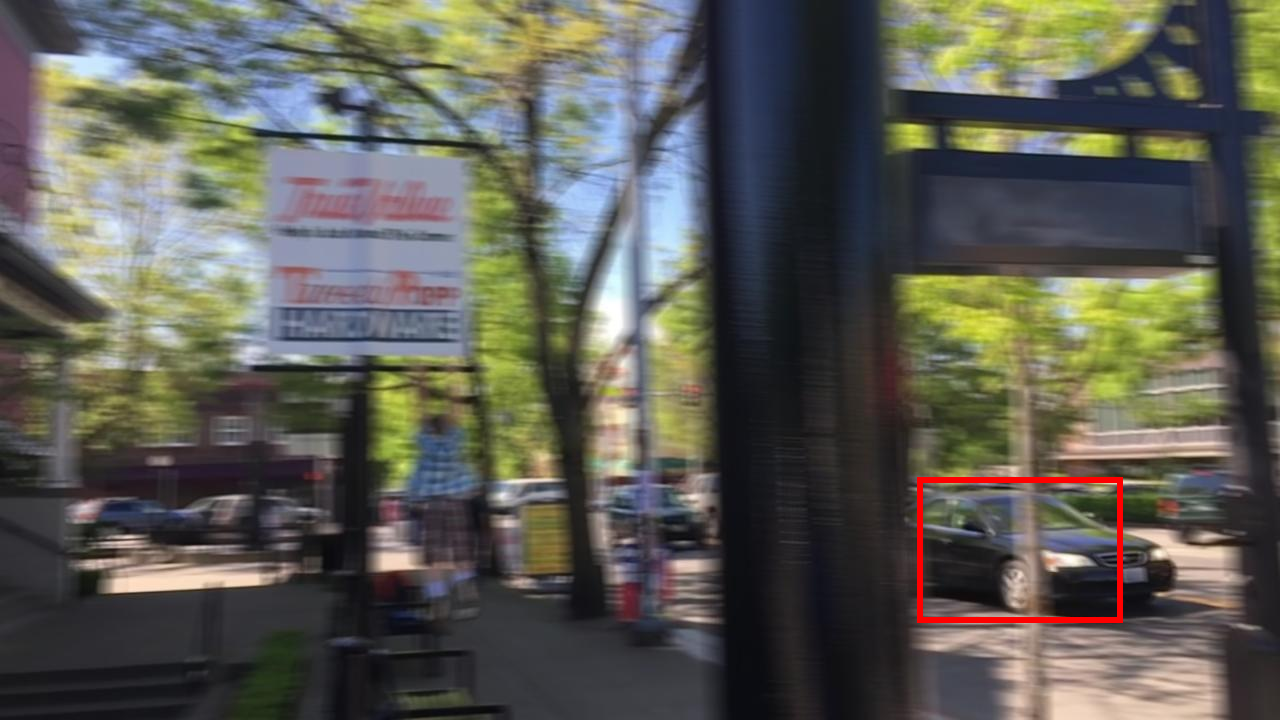}}}
            & \hspace{-4.0mm} \includegraphics[width=0.16\linewidth,height=0.105\linewidth]{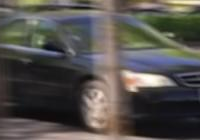}
            & \hspace{-4.0mm} \includegraphics[width=0.16\linewidth,height=0.105\linewidth]{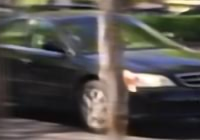}
            & \hspace{-4.0mm} \includegraphics[width=0.16\linewidth,height=0.105\linewidth]{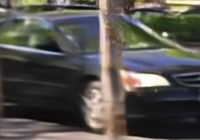}
            & \hspace{-4.0mm} \includegraphics[width=0.16\linewidth,height=0.105\linewidth]{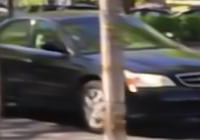}
              \\
    		\multicolumn{3}{c}{~}
            & \hspace{-4.0mm} (a) Blurred patch
            & \hspace{-4.0mm} (b) DBGAN
            & \hspace{-4.0mm} (c) FFTformer
            & \hspace{-4.0mm} (d) HI-Diff \\		
    	\multicolumn{3}{c}{~}
            & \hspace{-4.0mm} \includegraphics[width=0.16\linewidth,height=0.105\linewidth]{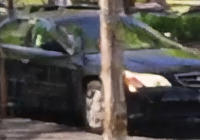}
            & \hspace{-4.0mm} \includegraphics[width=0.16\linewidth,height=0.105\linewidth]{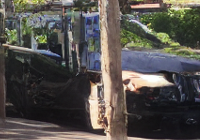}
            & \hspace{-4.0mm} \includegraphics[width=0.16\linewidth,height=0.105\linewidth]{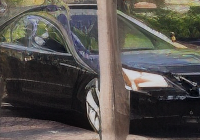}
            & \hspace{-4.0mm} \includegraphics[width=0.16\linewidth,height=0.105\linewidth]{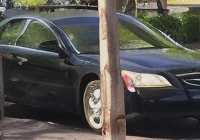}
            \\
    	\multicolumn{3}{c}{\hspace{-4.0mm} Blurred image from DVD dataset}
            & \hspace{-4.0mm} (e) ResShift
            & \hspace{-4.0mm} (f) PASD
            & \hspace{-4.0mm} (g) DiffBIR
            & \hspace{-4.0mm} (h) Ours\\

    \end{tabular}
\vspace{-5mm}
\caption{Deblurred results on the DVD dataset~\cite{dvd}. Existing methods struggle to effectively restore clear images. In contrast, our approach not only removes blur but also recovers sharp structures and fine details.}
\label{fig: dvd}
\vspace{-2mm}
\end{figure*}

\begin{figure*}[!t]
\footnotesize
\vspace{0mm}
\centering
    \begin{tabular}{c c c c c c c}
            \multicolumn{3}{c}{\multirow{5}*[45.6pt]{
            \hspace{-4mm} 
            \vspace{-25mm}
            \includegraphics[width=0.325\linewidth,height=0.236\linewidth]{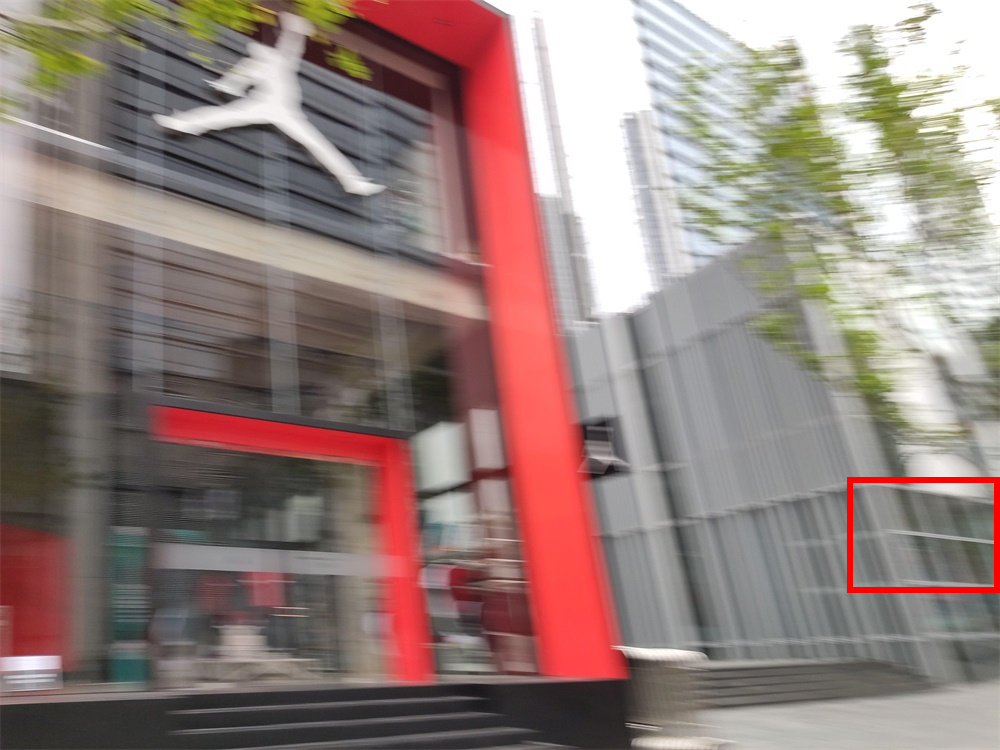}}}
            & \hspace{-4.0mm} \includegraphics[width=0.16\linewidth,height=0.105\linewidth]{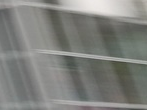}
            & \hspace{-4.0mm} \includegraphics[width=0.16\linewidth,height=0.105\linewidth]{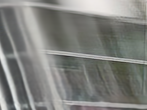}
            & \hspace{-4.0mm} \includegraphics[width=0.16\linewidth,height=0.105\linewidth]{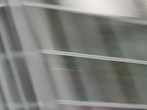}
            & \hspace{-4.0mm} \includegraphics[width=0.16\linewidth,height=0.105\linewidth]{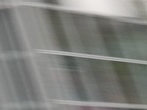}
              \\
    		\multicolumn{3}{c}{~}
            & \hspace{-4.0mm} (a) Blurred patch
            & \hspace{-4.0mm} (b) DBGAN
            & \hspace{-4.0mm} (c) FFTformer
            & \hspace{-4.0mm} (d) HI-Diff \\		
    	\multicolumn{3}{c}{~}
            & \hspace{-4.0mm} \includegraphics[width=0.16\linewidth,height=0.105\linewidth]{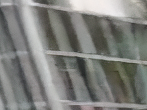}
            & \hspace{-4.0mm} \includegraphics[width=0.16\linewidth,height=0.105\linewidth]{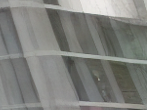}
            & \hspace{-4.0mm} \includegraphics[width=0.16\linewidth,height=0.105\linewidth]{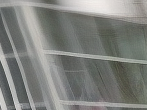}
            & \hspace{-4.0mm} \includegraphics[width=0.16\linewidth,height=0.105\linewidth]{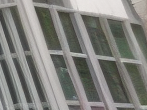}
            \\
    	\multicolumn{3}{c}{\hspace{-4.0mm} Blurred image from RWBI dataset}
            & \hspace{-4.0mm} (e) ResShift
            & \hspace{-4.0mm} (f) PASD
            & \hspace{-4.0mm} (g) DiffBIR
            & \hspace{-4.0mm} (h) Ours\\

    \end{tabular}
\vspace{-5mm}
\caption{Deblurred results on the RWBI dataset~\cite{dbgan}. The structures are not recovered well in (b)-(g). The proposed method generates an image with much clearer structures.}
\label{fig: rwbi}
\vspace{-2mm}
\end{figure*}
%

%
\begin{figure*}[!t]
\footnotesize
\vspace{-2mm}
\centering
    \begin{tabular}{c c c c c c c}
            \multicolumn{3}{c}{\multirow{5}*[45.6pt]{
            \hspace{-4mm} 
            \vspace{-25mm}
            \includegraphics[width=0.325\linewidth,height=0.236\linewidth]{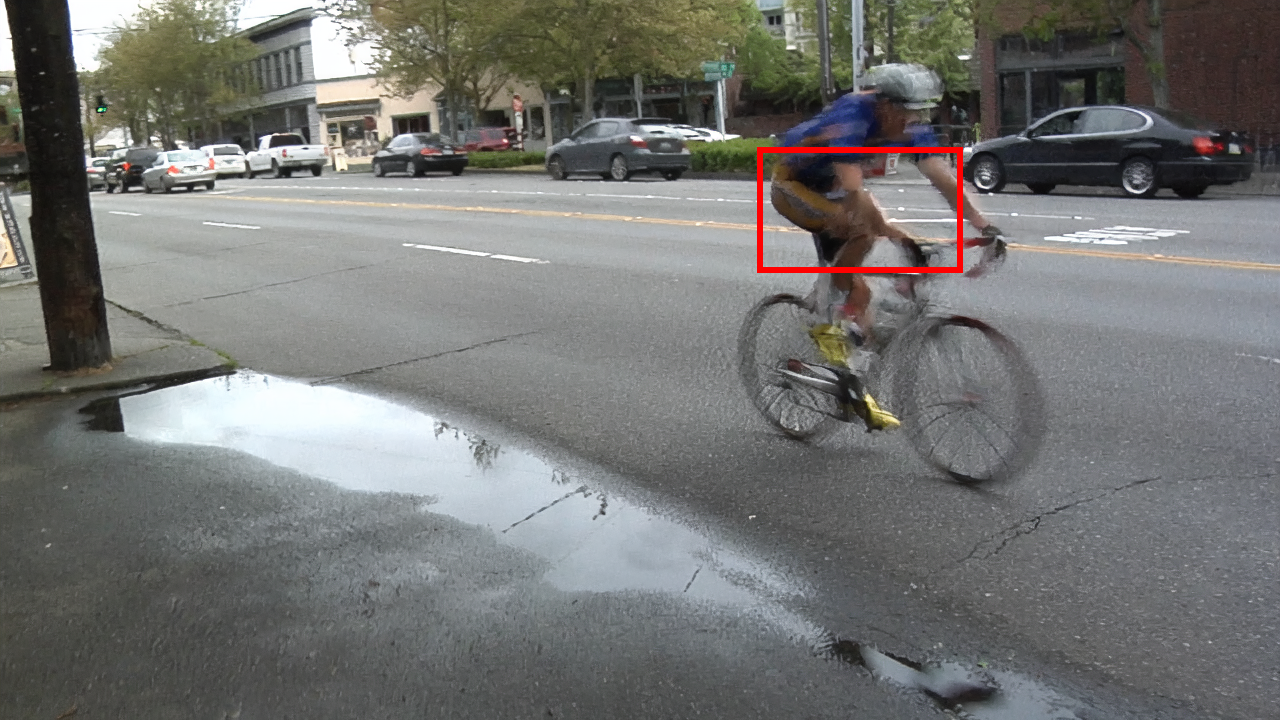}}}
            & \hspace{-4.0mm} \includegraphics[width=0.16\linewidth,height=0.105\linewidth]{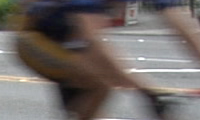}
            & \hspace{-4.0mm} \includegraphics[width=0.16\linewidth,height=0.105\linewidth]{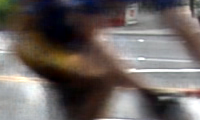}
            & \hspace{-4.0mm} \includegraphics[width=0.16\linewidth,height=0.105\linewidth]{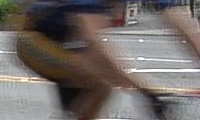}
            & \hspace{-4.0mm} \includegraphics[width=0.16\linewidth,height=0.105\linewidth]{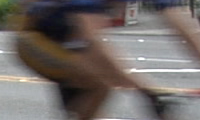}
              \\
    		\multicolumn{3}{c}{~}
            & \hspace{-4.0mm} (a) Blurred patch
            & \hspace{-4.0mm} (b) DBGAN
            & \hspace{-4.0mm} (c) FFTformer
            & \hspace{-4.0mm} (d) HI-Diff \\		
    	\multicolumn{3}{c}{~}
            & \hspace{-4.0mm} \includegraphics[width=0.16\linewidth,height=0.105\linewidth]{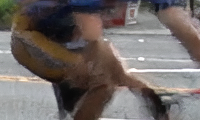}
            & \hspace{-4.0mm} \includegraphics[width=0.16\linewidth,height=0.105\linewidth]{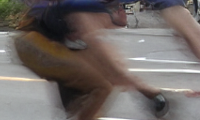}
            & \hspace{-4.0mm} \includegraphics[width=0.16\linewidth,height=0.105\linewidth]{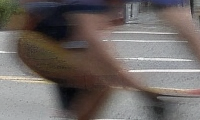}
            & \hspace{-4.0mm} \includegraphics[width=0.16\linewidth,height=0.105\linewidth]{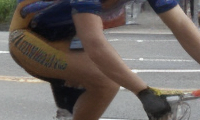}
            \\
    	\multicolumn{3}{c}{\hspace{-4.0mm} Blurred image from Real Blurry Images}
            & \hspace{-4.0mm} (e) ResShift
            & \hspace{-4.0mm} (f) PASD
            & \hspace{-4.0mm} (g) DiffBIR
            & \hspace{-4.0mm} (h) Ours\\

    \end{tabular}
\vspace{-5mm}
\caption{Deblurred results on real blurry images dataset~\cite{real_image}. The deblurred results in (b)-(g) still contain significant blur effects. The proposed method generates a clear image. }
\label{fig: real_blurry}
\vspace{-6mm}
\end{figure*}

\section{Experimental Results}
\subsection{Experimental Settings}
{\flushleft \textbf{Training Datasets.}} 
Current existing deblurring datasets are generally small in scale and low in resolution, which are insufficient for effectively training diffusion models. 
Therefore, we do not use common deblurring datasets such as GoPro~\cite{GoPro} as our training set.
Instead 
We have collected and created a large-scale dataset containing approximately 500,000 data pairs.
Our training dataset consists of three parts: (1) Existing deblurring datasets (including MC-Blur~\cite{MC-blur} and RSBlur~\cite{RSBlur}).
(2) We collect and capture some high-definition video clips, generating blurred and clear data pairs using the same strategy as REDS~\cite{REDS}.
(3) We collect a large number of high-definition images and generate various motion blur kernels to synthesize corresponding blurred images.
We provide more details about the training dataset in the appendix.
{\flushleft \textbf{Test Datasets.}} 
We evaluate the proposed DeblurDiff on commonly used image deblurring dataset, including synthetic datasets (GoPro~\cite{GoPro}, DVD~\cite{dvd}) and Real Blurry Images~\cite{real_image}, RealBlur~\cite{Realblur}, RWBI~\cite{dbgan}.
{\flushleft \textbf{Implementation Details.}} 
We employ the Adam optimizer~\cite{Adam} to train DeblurDiff with a batch size of 128. 
The learning rate is set to a fixed value of \(5 \times 10^{-5}\). 
The model is trained for 100K iterations using 8 NVIDIA 80G-A100 GPUs, each with 80GB of memory.
{\flushleft \textbf{Evaluation Metrics.}} 
We employ a range of reference-based and no-reference metrics to provide a comprehensive evaluation of the deblurring performance.
For fidelity and perceptual quality assessment, we employ reference-based metrics, including PSNR, SSIM, and LPIPS~\cite{lpips}.
For no-reference evaluation, we include NIQE~\cite{NIQE}, MANIQA~\cite{MANIQA}, MUSIQ~\cite{MUSIQ}, and CLIPIQA~\cite{CLIP-IQA}, which assess image quality based on statistical and learning-based approaches. 
This diverse set of metrics ensures a thorough analysis of both fidelity and perceptual quality.
{\flushleft \textbf{Compared Methods.}} 
We compare our DeblurDiff with several state-of-the-art image deblurring methods, which are categorized into two groups: (1) non-diffusion-based methods, including FFTformer (a transformer-based regression approach)~\cite{fftformer} and DBGAN (a GAN-based method)~\cite{dbgan}, and (2) diffusion-based methods, including HiDiff~\cite{hidiff}, ResShift~\cite{resshift}, ControlNet~\cite{controlnet}, PASD~\cite{pasd}, and DiffBIR~\cite{diffbir}.
To ensure a fair comparison, we retrain FFTformer on our training dataset.
We employ the pre-trained FFTformer as the Degradation Removal Module (DRM) in DiffBIR and retrain DiffBIR, as the original DiffBIR framework is not designed for deblurring tasks.
We also train a ControlNet and PASD for image deblurring tasks.
All the aforementioned retrained methods are trained and tested using the publicly released codes and models of the competing methods, ensuring a fair and consistent comparison under the same dataset and training settings as our proposed method.

\subsection{Comparisons with the state of the arts}
We evaluate our approach on the synthetic and real-world datasets.
Table~\ref{tab:result} shows the quantitative results.
Our method shows strong performance in no-reference metrics, achieving higher scores compared to existing approaches. This indicates that our method excels in perceptual quality and realism, which are critical for real-world applications where ground truth images are often unavailable.
Our method achieves lower scores in reference-based metrics compared to HI-DIff. 
However, HI-Diff does not utilize a pre-trained SD model, thus focusing more on reference-based metrics while lacking generative capabilities, which results in relatively worse performance in no-reference metrics.
%
%
\begin{table}[!t]
\vspace{-6mm}
  \caption{Effectiveness of each component in the proposed method on the real blurry images~\cite{real_image}.
  }
   \label{tab: ablation}
\footnotesize
\resizebox{0.48\textwidth}{!}{
 \centering
 \begin{tabular}{lccc|cccc}
    \toprule
                 &   LKPN           & EAC             & SD prior for LKPN    &NIQE&    MANIQA\\
    \hline
  ControlNet       &\XSolidBrush    &\XSolidBrush     &\XSolidBrush        &4.0978& 0.5544             \\
  w/o EAC        &\CheckmarkBold    &\XSolidBrush     &\CheckmarkBold        &4.0230& 0.5874 \\
  w/o SD for LKPN         &\CheckmarkBold    &\CheckmarkBold   &\XSolidBrush      &3.8087 &0.5882    \\
  DeblurDiff     &\CheckmarkBold    &\CheckmarkBold   &\CheckmarkBold      &\textbf{3.6628} &\textbf{0.5963}& \\
 \bottomrule

  \end{tabular}
}
 \vspace{-9mm}
\end{table}
Figure~\ref{fig: dvd} shows visual comparisons on the synthetic dataset of DVD~\cite{dvd}.
The GAN-based method~\cite{dbgan} exhibits significantly inferior deblurring performance, failing to restore clear structures and fine details effectively.
Existing diffusion-based methods, such as HiDiff~\cite{hidiff} and ResShift~\cite{resshift}, fail to achieve satisfactory results in Figure~\ref{fig: dvd} (d) and (e) due to their lack of SD priors, which leads to suboptimal generation quality in terms of both structural clarity and detail fidelity.
Leveraging the pre-trained FFTformer~\cite{fftformer} to preprocess blurred images Figure~\ref{fig: dvd} (c) and subsequently applying diffusion-based restoration can partially remove blur. 
However, since FFTformer sometimes cannot eliminate blur and tends to introduce undesirable artifacts during preprocessing, the final results generated by DiffBIR often exhibit unnatural structures and are inconsistent with the input.
In contrast, our method, guided by the LKPN that progressively generates clear structures, produces better results with sharper and more accurate structures.

Figure~\ref{fig: rwbi} and Figure~\ref{fig: real_blurry} present the visual comparison results on the real-world dataset. 
DBGAN~\cite{dbgan} and FFTformer~\cite{fftformer} struggle to recover clear structures from severe blur, while Hi-Diff~\cite{hidiff} and ResShift~\cite{resshift} fail to reconstruct fine details due to the lack of pre-trained image priors. 
PASD~\cite{pasd} and DiffBIR~\cite{diffbir}, which rely on degradation removal models for pre-deblurring, often produce suboptimal results with noticeable artifacts, as these models cannot effectively handle complex blur patterns. 
In contrast, our method demonstrates better performance in both structural recovery and detail reconstruction. 
The LKPN progressively generates clear structures and adaptively refines the deblurring process, enabling our approach to achieve high-quality results with minimal artifacts.
\subsection{Analysis experiments.}
\begin{figure}[!t]\tiny
\centering
\vspace{-2mm}
\begin{tabular}{cccccc}
\hspace{-3mm}
\includegraphics[width=0.09\textwidth,height=0.19\linewidth]{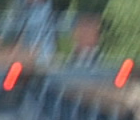} &\hspace{-4mm}
\includegraphics[width=0.09\textwidth,height=0.19\linewidth]{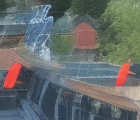} &\hspace{-4mm}
\includegraphics[width=0.09\textwidth,height=0.19\linewidth]{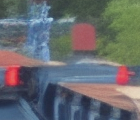} &\hspace{-4mm}
\includegraphics[width=0.09\textwidth,height=0.19\linewidth]{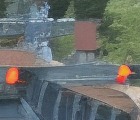} &\hspace{-4mm}
\includegraphics[width=0.09\textwidth,height=0.19\linewidth]{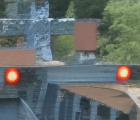} \\
\hspace{-3mm}(a) Blurred image &\hspace{-4mm} (b) ControlNet &\hspace{-4mm} (c) w/o EAC&\hspace{-4mm}(d) w/o SD &\hspace{-4mm}(e) DeblurDiff\\
\end{tabular}
\vspace{-3mm}
\caption{Effectiveness of the proposed DeblurDiff on image deblurring. We provide more detailed results in the appendix. }
\label{fig: ablation}
\vspace{-7mm}
\end{figure}

The proposed LKPN is used to leverage intermediate clear priors generated during the diffusion process, providing clear structural guidance as conditional inputs to the diffusion model.
When the blurry image is directly used as the conditional input (ControlNet for short), the diffusion model struggles to recover clear structures and fine details (Figure~\ref{fig: ablation}(b)), particularly in cases of severe blur. 
This is because the blurry image lacks sufficient structural information to guide the generation process effectively, leading to suboptimal deblurring performance.
Table~\ref{tab: ablation} shows the quantitative evaluation results on the real blurry images dataset.
When the LKPN is modified to directly predict the deblurred result in the latent space to serve as the conditional input for diffusion, instead of estimating element-wise adaptive kernels (w/o EAC for short), the performance is worse compared to our proposed method that predicts a kernel and uses EAC to generate a clear guidance. 
This approach fails to adaptively address distinct blur characteristics at each pixel location, resulting in artifacts and inconsistencies in the generated images. Additionally, it cannot effectively preserve the input information and structural integrity, leading to a loss of important details and coherence in the deblurred results(Figure~\ref{fig: ablation}(c)). 
In contrast, our full method, which estimates pixel-specific deblurring kernels, dynamically adjusts to local content, enabling more accurate deblurring.
To further validate the effectiveness of leveraging clear priors from the diffusion process in the LKPN, we compare our method with the baseline that directly estimates the kernel from the blurry image without utilizing the intermediate results from the diffusion process (w/o SD for LKPN for short).
In this baseline, the estimated kernel is applied to the blurry image to generate a deblurred result, which is then used as a conditional input to the diffusion model.
However, this approach lacks the iterative refinement provided by intermediate image priors from the diffusion process. 
As a result, the LKPN struggles to provide accurate structural guidance, leading to structural errors and inconsistencies in the generated results, as shown in Figure~\ref{fig: ablation}(d).

\section{Conclusion}
In this paper, we propose DeblurDiff, a framework for real-world image deblurring that integrates a Latent Kernel Prediction Network (LKPN) with a generative diffusion model. 
Our approach addresses the limitations of existing methods by leveraging the priors of pre-trained SD models and introducing an adaptive mechanism to estimate pixel-specific kernels.
These kernels are applied through Element-wise Adaptive Convolution (EAC), which adaptively adjusts to local content, enabling the model to preserve input information and structural integrity effectively. 
Through extensive experiments on both synthetic and real-world benchmarks, we demonstrated that DeblurDiff performs favorably against state-of-the-art methods in terms of structural fidelity and visual quality.
%
\nocite{langley00}

\bibliography{example_paper}

\begin{thebibliography}{46}
\providecommand{\natexlab}[1]{#1}
\providecommand{\url}[1]{\texttt{#1}}
\expandafter\ifx\csname urlstyle\endcsname\relax
  \providecommand{\doi}[1]{doi: #1}\else
  \providecommand{\doi}{doi: \begingroup \urlstyle{rm}\Url}\fi

\bibitem[Boracchi \& Foi(2012)Boracchi and Foi]{PSF}
Boracchi, G. and Foi, A.
\newblock Modeling the performance of image restoration from motion blur.
\newblock \emph{Image Processing, IEEE Transactions on}, 21\penalty0 (8):\penalty0 3502 --3517, aug. 2012.

\bibitem[Chen et~al.(2022)Chen, Chu, Zhang, and Sun]{NAFNet}
Chen, L., Chu, X., Zhang, X., and Sun, J.
\newblock Simple baselines for image restoration.
\newblock In \emph{ECCV}, 2022.

\bibitem[Chen et~al.(2023)Chen, Zhang, Ding, Bin, Gu, Kong, and Yuan]{hidiff}
Chen, Z., Zhang, Y., Ding, L., Bin, X., Gu, J., Kong, L., and Yuan, X.
\newblock Hierarchical integration diffusion model for realistic image deblurring.
\newblock In \emph{NeurIPS}, 2023.

\bibitem[Cho et~al.(2012)Cho, Wang, and Lee]{real_image}
Cho, S., Wang, J., and Lee, S.
\newblock Video deblurring for hand-held cameras using patch-based synthesis.
\newblock \emph{ACM Transactions on Graphics (TOG)}, 31:\penalty0 1 -- 9, 2012.

\bibitem[Cho et~al.(2021)Cho, Ji, Hong, Jung, and Ko]{MIMO}
Cho, S.-J., Ji, S.-W., Hong, J.-P., Jung, S.-W., and Ko, S.-J.
\newblock Rethinking coarse-to-fine approach in single image deblurring.
\newblock In \emph{ICCV}, 2021.

\bibitem[Dhariwal \& Nichol(2021)Dhariwal and Nichol]{diffusion1}
Dhariwal, P. and Nichol, A.
\newblock Diffusion models beat gans on image synthesis.
\newblock In \emph{NeurIPS}, 2021.

\bibitem[He et~al.(2016)He, Zhang, Ren, and Sun]{resnet}
He, K., Zhang, X., Ren, S., and Sun, J.
\newblock Deep residual learning for image recognition.
\newblock In \emph{CVPR}, 2016.

\bibitem[Ho et~al.(2020)Ho, Jain, and Abbeel]{DDPM}
Ho, J., Jain, A., and Abbeel, P.
\newblock Denoising diffusion probabilistic models.
\newblock In \emph{NeurIPS}, 2020.

\bibitem[Ke et~al.(2021)Ke, Wang, Wang, Milanfar, and Yang]{MUSIQ}
Ke, J., Wang, Q., Wang, Y., Milanfar, P., and Yang, F.
\newblock Musiq: Multi-scale image quality transformer.
\newblock In \emph{ICCV}, 2021.

\bibitem[Kingma \& Ba(2015)Kingma and Ba]{Adam}
Kingma, D.~P. and Ba, J.
\newblock Adam: A method for stochastic optimization.
\newblock In \emph{ICLR}, 2015.

\bibitem[Kong et~al.(2023)Kong, Dong, Ge, Li, and Pan]{fftformer}
Kong, L., Dong, J., Ge, J., Li, M., and Pan, J.
\newblock Efficient frequency domain-based transformers for high-quality image deblurring.
\newblock In \emph{CVPR}, 2023.

\bibitem[Krishnan et~al.(2011)Krishnan, Tay, and Fergus]{sparse}
Krishnan, D., Tay, T., and Fergus, R.
\newblock Blind deconvolution using a normalized sparsity measure.
\newblock In \emph{CVPR}, 2011.

\bibitem[Kupyn et~al.(2018)Kupyn, Budzan, Mykhailych, Mishkin, and Matas]{DeblurGAN}
Kupyn, O., Budzan, V., Mykhailych, M., Mishkin, D., and Matas, J.
\newblock Deblurgan: Blind motion deblurring using conditional adversarial networks.
\newblock In \emph{CVPR}, 2018.

\bibitem[Kupyn et~al.(2019)Kupyn, Martyniuk, Wu, and Wang]{DeblurGANv2}
Kupyn, O., Martyniuk, T., Wu, J., and Wang, Z.
\newblock Deblurgan-v2: Deblurring (orders-of-magnitude) faster and better.
\newblock In \emph{ICCV}, 2019.

\bibitem[Ledig et~al.(2017)Ledig, Theis, Husz{\'a}r, Caballero, Cunningham, Acosta, Aitken, Tejani, Totz, Wang, et~al.]{GAN_SR}
Ledig, C., Theis, L., Husz{\'a}r, F., Caballero, J., Cunningham, A., Acosta, A., Aitken, A., Tejani, A., Totz, J., Wang, Z., et~al.
\newblock Photo-realistic single image super-resolution using a generative adversarial network.
\newblock In \emph{CVPR}, 2017.

\bibitem[Li et~al.(2023)Li, Fan, Xiang, Demandolx, Ranjan, Timofte, and Van~Gool]{GRL}
Li, Y., Fan, Y., Xiang, X., Demandolx, D., Ranjan, R., Timofte, R., and Van~Gool, L.
\newblock Efficient and explicit modelling of image hierarchies for image restoration.
\newblock In \emph{CVPR}, 2023.

\bibitem[Lin et~al.(2025)Lin, He, Chen, Lyu, Dai, Yu, Qiao, Ouyang, and Dong]{diffbir}
Lin, X., He, J., Chen, Z., Lyu, Z., Dai, B., Yu, F., Qiao, Y., Ouyang, W., and Dong, C.
\newblock Diffbir: Toward blind image restoration with generative diffusion prior.
\newblock In \emph{ECCV}, 2025.

\bibitem[Lugmayr et~al.(2020)Lugmayr, Danelljan, Van~Gool, and Timofte]{SRFlow}
Lugmayr, A., Danelljan, M., Van~Gool, L., and Timofte, R.
\newblock Srflow: Learning the super-resolution space with normalizing flow.
\newblock In \emph{ECCV}, 2020.

\bibitem[Nah et~al.(2017)Nah, Kim, and Lee]{GoPro}
Nah, S., Kim, T.~H., and Lee, K.~M.
\newblock Deep multi-scale convolutional neural network for dynamic scene deblurring.
\newblock In \emph{CVPR}, 2017.

\bibitem[Nah et~al.(2019)Nah, Baik, Hong, Moon, Son, Timofte, and Mu~Lee]{REDS}
Nah, S., Baik, S., Hong, S., Moon, G., Son, S., Timofte, R., and Mu~Lee, K.
\newblock Ntire 2019 challenge on video deblurring and super-resolution: Dataset and study.
\newblock In \emph{CVPRW}, 2019.

\bibitem[Pan et~al.(2016)Pan, Sun, Pfister, and Yang]{dark_channel}
Pan, J., Sun, D., Pfister, H., and Yang, M.-H.
\newblock Blind image deblurring using dark channel prior.
\newblock In \emph{CVPR}, 2016.

\bibitem[Pan et~al.(2017)Pan, Hu, Su, and Yang]{L0}
Pan, J., Hu, Z., Su, Z., and Yang, M.-H.
\newblock $l_0$ -regularized intensity and gradient prior for deblurring text images and beyond.
\newblock \emph{IEEE Transactions on Pattern Analysis and Machine Intelligence}, 39\penalty0 (2):\penalty0 342--355, 2017.
\newblock \doi{10.1109/TPAMI.2016.2551244}.

\bibitem[Pan et~al.(2021)Pan, Dong, Liu, Zhang, Ren, Tang, Tai, and Yang]{physicgan}
Pan, J., Dong, J., Liu, Y., Zhang, J., Ren, J. S.~J., Tang, J., Tai, Y.-W., and Yang, M.-H.
\newblock Physics-based generative adversarial models for image restoration and beyond.
\newblock \emph{IEEE TPAMI}, 43\penalty0 (7):\penalty0 2449--2462, 2021.

\bibitem[Ravi et~al.(2024)Ravi, Gabeur, Hu, Hu, Ryali, Ma, Khedr, R{\"a}dle, Rolland, Gustafson, Mintun, Pan, Alwala, Carion, Wu, Girshick, Doll{\'a}r, and Feichtenhofer]{sam2}
Ravi, N., Gabeur, V., Hu, Y.-T., Hu, R., Ryali, C., Ma, T., Khedr, H., R{\"a}dle, R., Rolland, C., Gustafson, L., Mintun, E., Pan, J., Alwala, K.~V., Carion, N., Wu, C.-Y., Girshick, R., Doll{\'a}r, P., and Feichtenhofer, C.
\newblock Sam 2: Segment anything in images and videos.
\newblock \emph{arXiv preprint arXiv:2408.00714}, 2024.

\bibitem[Ren et~al.(2023)Ren, Delbracio, Talebi, Gerig, and Milanfar]{cdm}
Ren, M., Delbracio, M., Talebi, H., Gerig, G., and Milanfar, P.
\newblock Multiscale structure guided diffusion for image deblurring.
\newblock In \emph{CVPR}, 2023.

\bibitem[Rim et~al.(2020)Rim, Lee, Won, and Cho]{Realblur}
Rim, J., Lee, H., Won, J., and Cho, S.
\newblock Real-world blur dataset for learning and benchmarking deblurring algorithms.
\newblock In \emph{ECCV}, 2020.

\bibitem[Rim et~al.(2022)Rim, Kim, Kim, Lee, Lee, and Cho]{RSBlur}
Rim, J., Kim, G., Kim, J., Lee, J., Lee, S., and Cho, S.
\newblock Realistic blur synthesis for learning image deblurring.
\newblock In \emph{ECCV}, 2022.

\bibitem[Rombach et~al.(2022)Rombach, Blattmann, Lorenz, Esser, and Ommer]{sd}
Rombach, R., Blattmann, A., Lorenz, D., Esser, P., and Ommer, B.
\newblock High-resolution image synthesis with latent diffusion models.
\newblock In \emph{CVPR}, 2022.

\bibitem[Su et~al.(2017)Su, Delbracio, Wang, Sapiro, Heidrich, and Wang]{dvd}
Su, S., Delbracio, M., Wang, J., Sapiro, G., Heidrich, W., and Wang, O.
\newblock Deep video deblurring for hand-held cameras.
\newblock In \emph{CVPR}, 2017.

\bibitem[Tao et~al.(2018)Tao, Gao, Shen, Wang, and Jia]{SRN}
Tao, X., Gao, H., Shen, X., Wang, J., and Jia, J.
\newblock Scale-recurrent network for deep image deblurring.
\newblock In \emph{CVPR}, 2018.

\bibitem[Vaswani et~al.(2017)Vaswani, Shazeer, Parmar, Uszkoreit, Jones, Gomez, Kaiser, and Polosukhin]{Transformer}
Vaswani, A., Shazeer, N., Parmar, N., Uszkoreit, J., Jones, L., Gomez, A.~N., Kaiser, L., and Polosukhin, I.
\newblock Attention is all you need.
\newblock In \emph{NIPS}, 2017.

\bibitem[Wang et~al.(2023)Wang, Chan, and Loy]{CLIP-IQA}
Wang, J., Chan, K.~C., and Loy, C.~C.
\newblock Exploring clip for assessing the look and feel of images.
\newblock In \emph{AAAI}, 2023.

\bibitem[Wang et~al.(2022)Wang, Cun, Bao, Zhou, Liu, and Li]{Uformer}
Wang, Z., Cun, X., Bao, J., Zhou, W., Liu, J., and Li, H.
\newblock Uformer: A general u-shaped transformer for image restoration.
\newblock In \emph{CVPR}, 2022.

\bibitem[Yang et~al.(2022)Yang, Wu, Shi, Lao, Gong, Cao, Wang, and Yang]{MANIQA}
Yang, S., Wu, T., Shi, S., Lao, S., Gong, Y., Cao, M., Wang, J., and Yang, Y.
\newblock Maniqa: Multi-dimension attention network for no-reference image quality assessment.
\newblock In \emph{CVPR)}, 2022.

\bibitem[Yang et~al.(2024)Yang, Wu, Ren, Xie, and Zhang]{pasd}
Yang, T., Wu, R., Ren, P., Xie, X., and Zhang, L.
\newblock Pixel-aware stable diffusion for realistic image super-resolution and personalized stylization.
\newblock In \emph{ECCV}, 2024.

\bibitem[Yuan et~al.(2007)Yuan, Sun, Quan, and Shum]{noise_deblur}
Yuan, L., Sun, J., Quan, L., and Shum, H.-Y.
\newblock Image deblurring with blurred/noisy image pairs.
\newblock \emph{ACM Trans. Graph.}, 26\penalty0 (3), 2007.

\bibitem[Yue et~al.(2024)Yue, Wang, and Loy]{resshift}
Yue, Z., Wang, J., and Loy, C.~C.
\newblock Resshift: Efficient diffusion model for image super-resolution by residual shifting.
\newblock In \emph{NeurIPS}, 2024.

\bibitem[Zamir et~al.(2021)Zamir, Arora, Khan, Hayat, Khan, Yang, and Shao]{MPRNet}
Zamir, S.~W., Arora, A., Khan, S., Hayat, M., Khan, F.~S., Yang, M.-H., and Shao, L.
\newblock Multi-stage progressive image restoration.
\newblock In \emph{CVPR}, 2021.

\bibitem[Zamir et~al.(2022)Zamir, Arora, Khan, Hayat, Khan, and Yang]{Restormer}
Zamir, S.~W., Arora, A., Khan, S., Hayat, M., Khan, F.~S., and Yang, M.-H.
\newblock Restormer: Efficient transformer for high-resolution image restoration.
\newblock In \emph{CVPR}, 2022.

\bibitem[Zhang et~al.(2020)Zhang, Luo, Zhong, Ma, Stenger, Liu, and Li]{dbgan}
Zhang, K., Luo, W., Zhong, Y., Ma, L., Stenger, B., Liu, W., and Li, H.
\newblock Deblurring by realistic blurring.
\newblock In \emph{CVPR}, 2020.

\bibitem[Zhang et~al.(2024)Zhang, Wang, Luo, Ren, Stenger, Liu, Li, and Yang]{MC-blur}
Zhang, K., Wang, T., Luo, W., Ren, W., Stenger, B., Liu, W., Li, H., and Yang, M.-H.
\newblock Mc-blur: A comprehensive benchmark for image deblurring.
\newblock \emph{IEEE Transactions on Circuits and Systems for Video Technology}, 34\penalty0 (5):\penalty0 3755--3767, 2024.

\bibitem[Zhang et~al.(2015)Zhang, Zhang, and Bovik]{NIQE}
Zhang, L., Zhang, L., and Bovik, A.~C.
\newblock A feature-enriched completely blind image quality evaluator.
\newblock \emph{IEEE Transactions on Image Processing}, 24\penalty0 (8):\penalty0 2579--2591, 2015.

\bibitem[Zhang et~al.(2023)Zhang, Rao, and Agrawala]{controlnet}
Zhang, L., Rao, A., and Agrawala, M.
\newblock Adding conditional control to text-to-image diffusion models.
\newblock In \emph{ICCV}, 2023.

\bibitem[Zhang et~al.(2018)Zhang, Isola, Efros, Shechtman, and Wang]{lpips}
Zhang, R., Isola, P., Efros, A.~A., Shechtman, E., and Wang, O.
\newblock The unreasonable effectiveness of deep features as a perceptual metric.
\newblock In \emph{CVPR}, 2018.

\bibitem[Zhao et~al.(2024)Zhao, Po, Ye, Xu, and Yan]{noise_deblur_2}
Zhao, Y., Po, L.-M., Ye, X., Xu, Y., and Yan, Q.
\newblock Modeling dual-exposure quad-bayer patterns for joint denoising and deblurring.
\newblock \emph{IEEE Transactions on Image Processing}, 2024.

\bibitem[Zhou et~al.(2019)Zhou, Zhang, Pan, Xie, Zuo, and Ren]{stfan}
Zhou, S., Zhang, J., Pan, J., Xie, H., Zuo, W., and Ren, J.
\newblock Spatio-temporal filter adaptive network for video deblurring.
\newblock In \emph{ICCV}, 2019.

\end{thebibliography}
\bibliographystyle{icml2025}

\newpage
\appendix
\section{Training datasets.}
As mentioned in the main paper, existing deblurring datasets such as GoPro~\cite{GoPro} and DVD~\cite{dvd} have several limitations. These datasets typically contain a relatively small number of images (usually only a few thousand) and a limited variety of scenes (often only a few dozen). This insufficiency makes it difficult to train a robust deblurring diffusion model. 
Moreover, for the GoPro dataset, which uses multi-frame video sequences to synthesize blur, the low frame rate of the videos results in blur patterns that significantly differ from real-world blur.
Therefore, we do not train our model on these problematic datasets. 
Instead, we constructed a large-scale dataset to train our DeblurDiff model. This dataset addresses the limitations of existing datasets by providing a richer and more diverse set of image pairs, thereby better supporting the learning and generalization capabilities of the model.
%

\begin{figure*}[!ht]
\footnotesize
\centering
    \begin{tabular}{cccc}
    \includegraphics[width=0.24\textwidth]{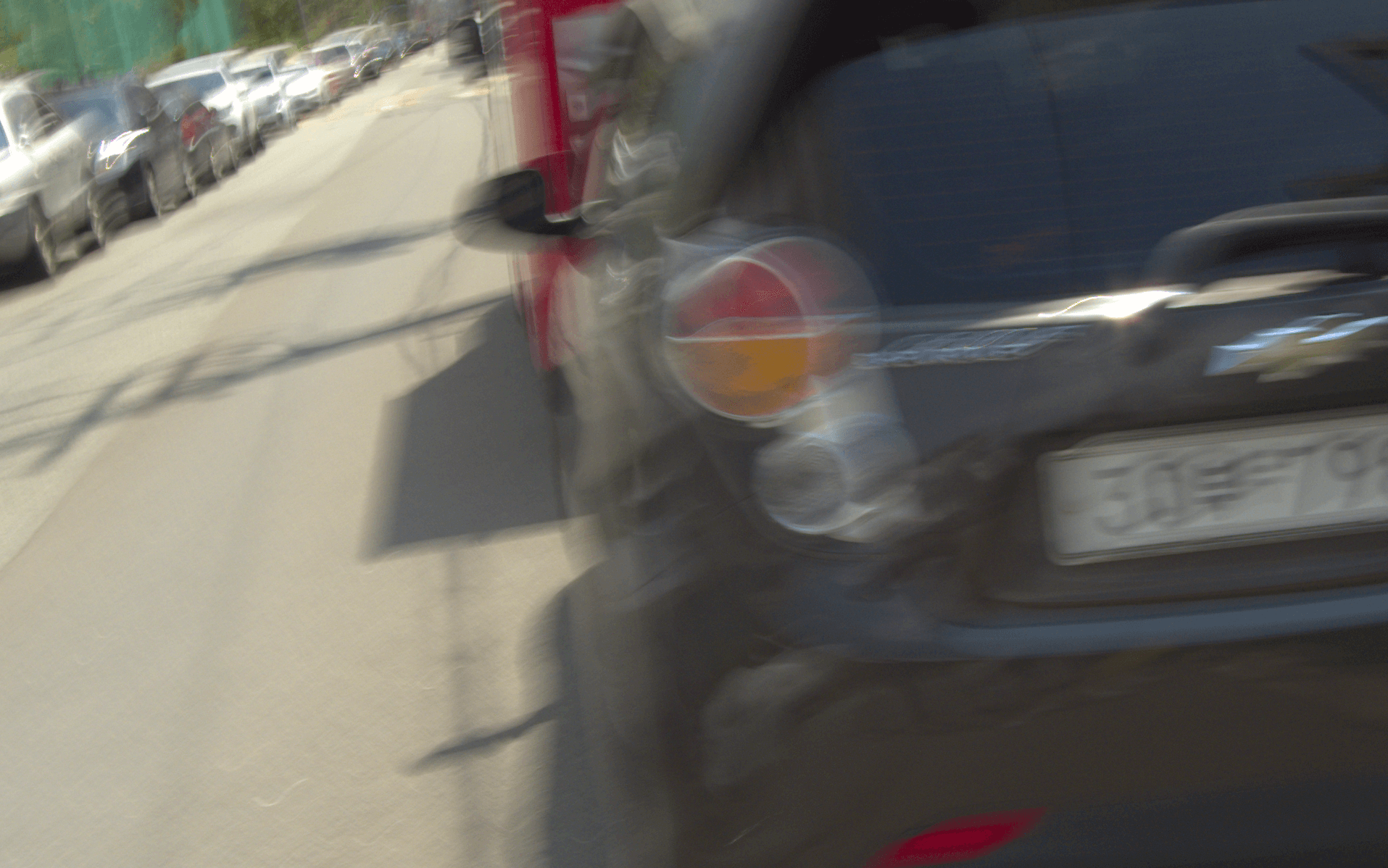}&\hspace{-3.5mm}
    \includegraphics[width=0.24\textwidth]{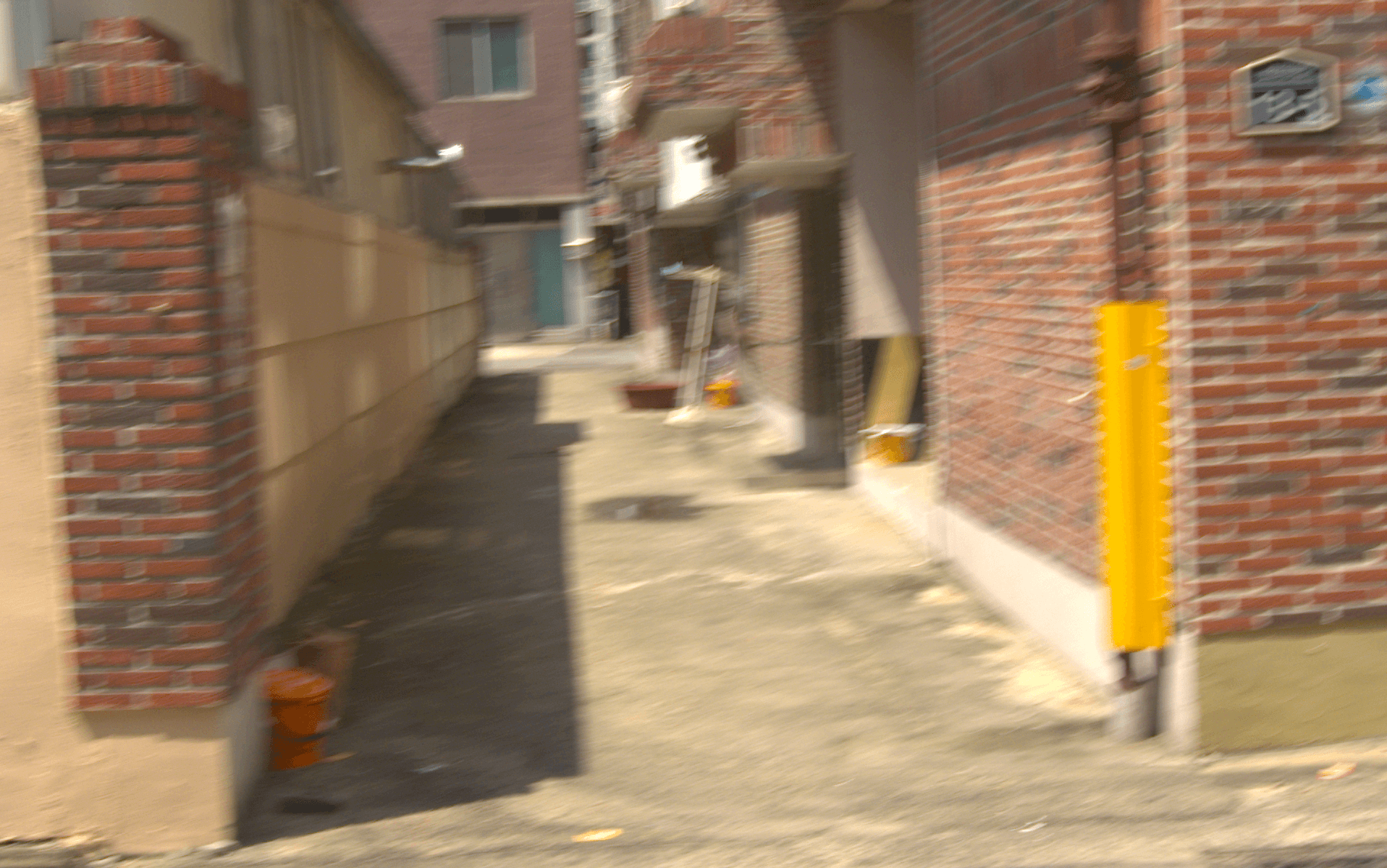}&\hspace{-4.5mm}
    \includegraphics[width=0.24\textwidth]{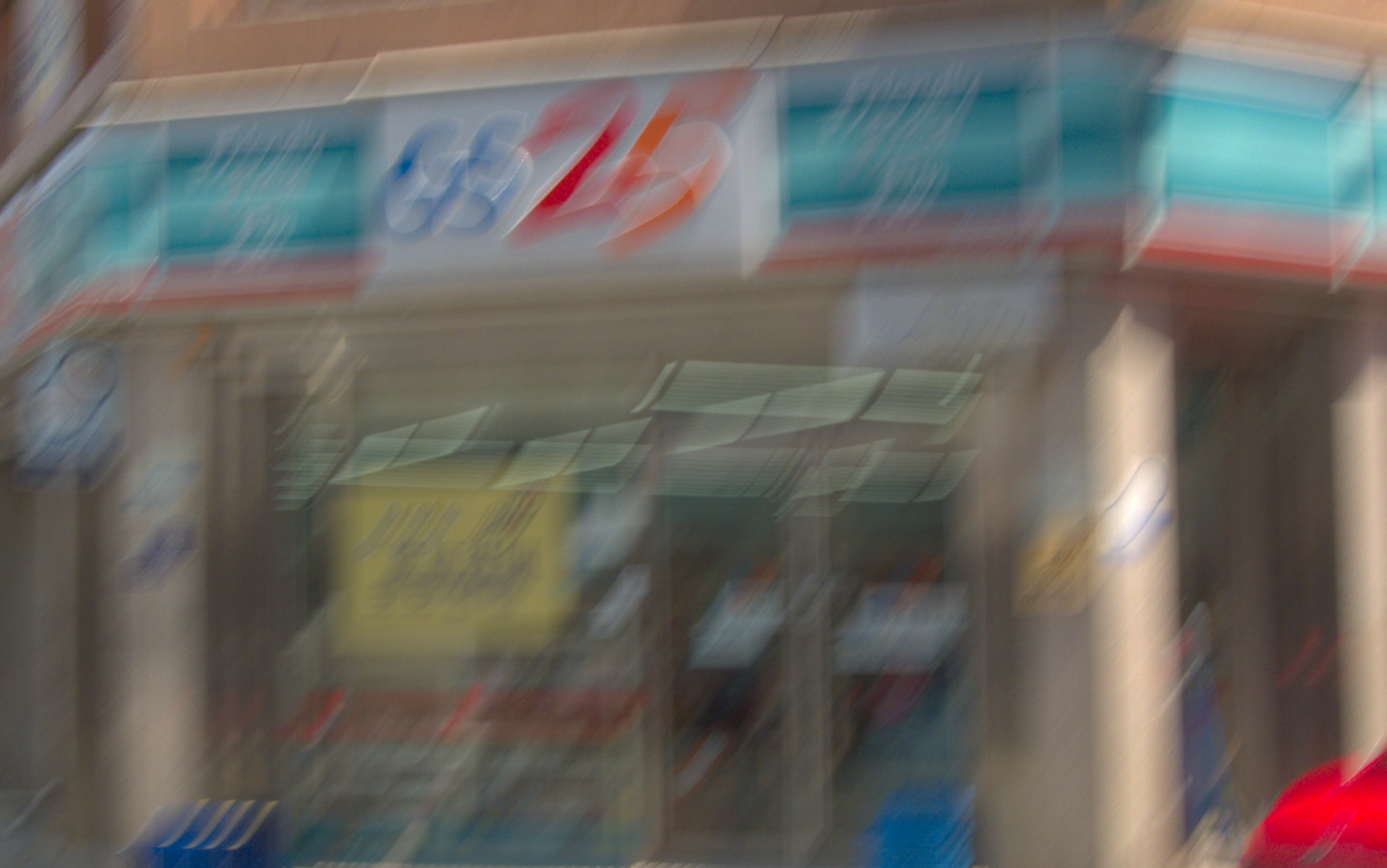}&\hspace{-3.5mm}
    \includegraphics[width=0.24\textwidth]{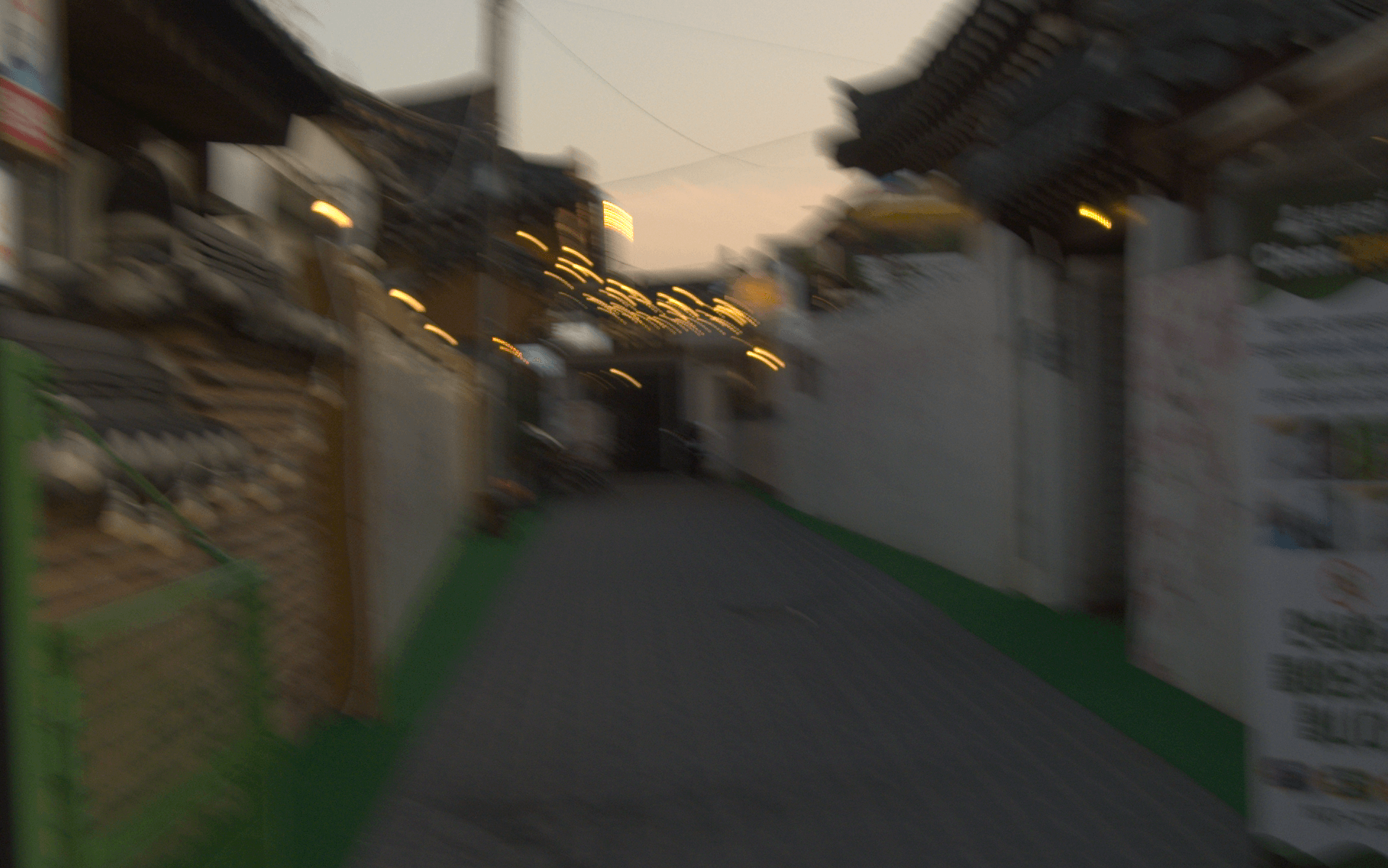}\\
    \includegraphics[width=0.24\textwidth]{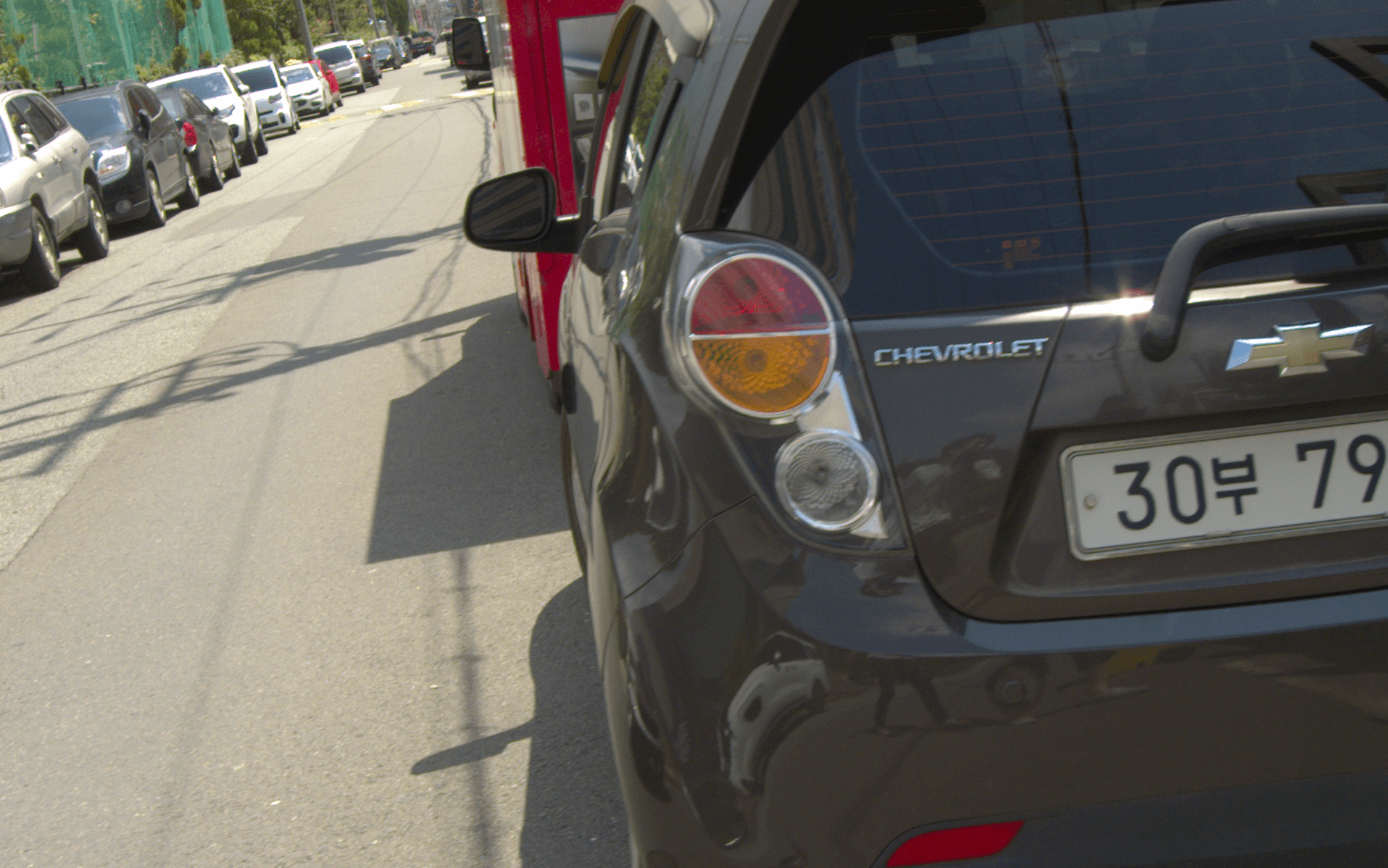}&\hspace{-3.5mm}
    \includegraphics[width=0.24\textwidth]{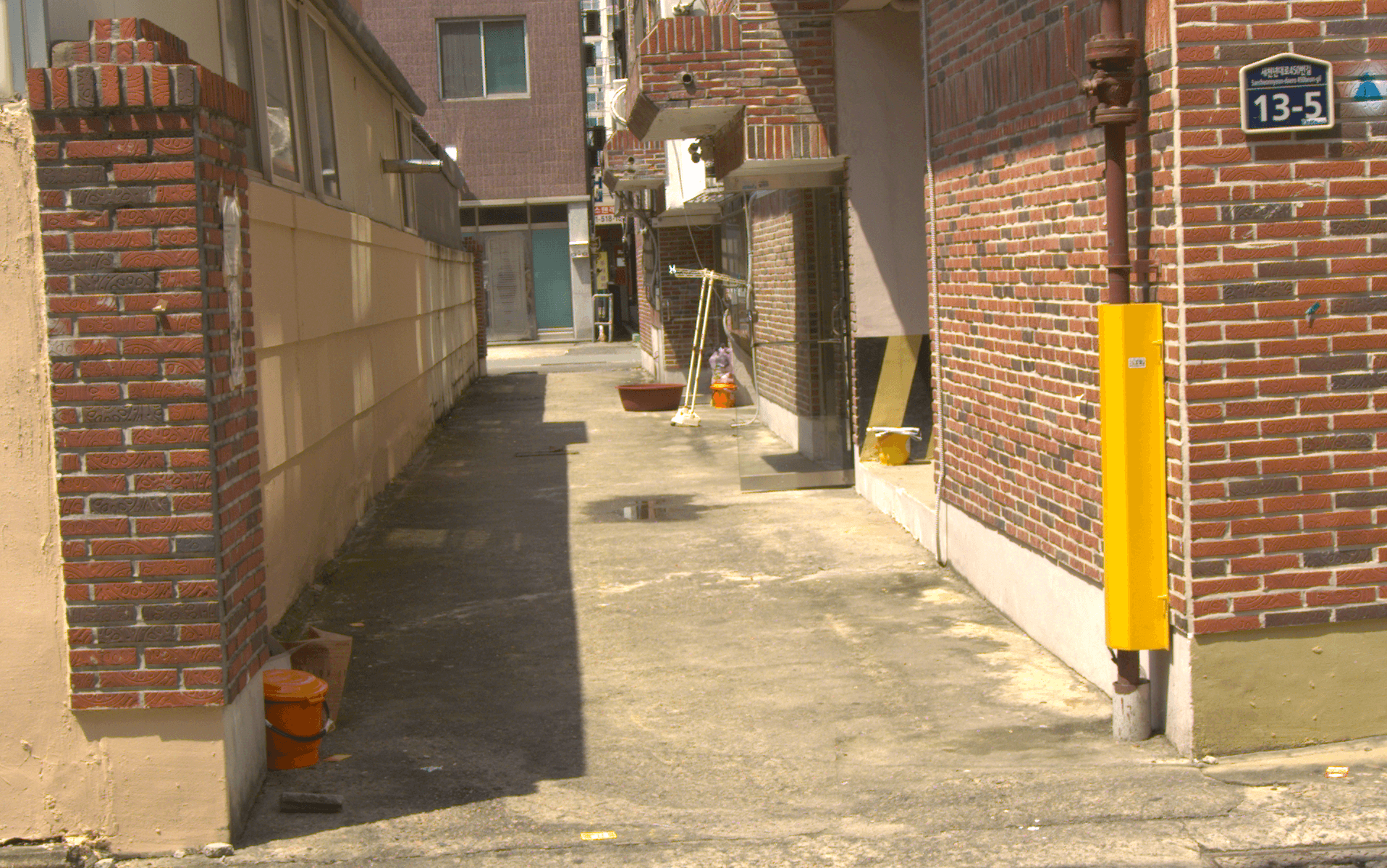}&\hspace{-4.5mm}
    \includegraphics[width=0.24\textwidth]{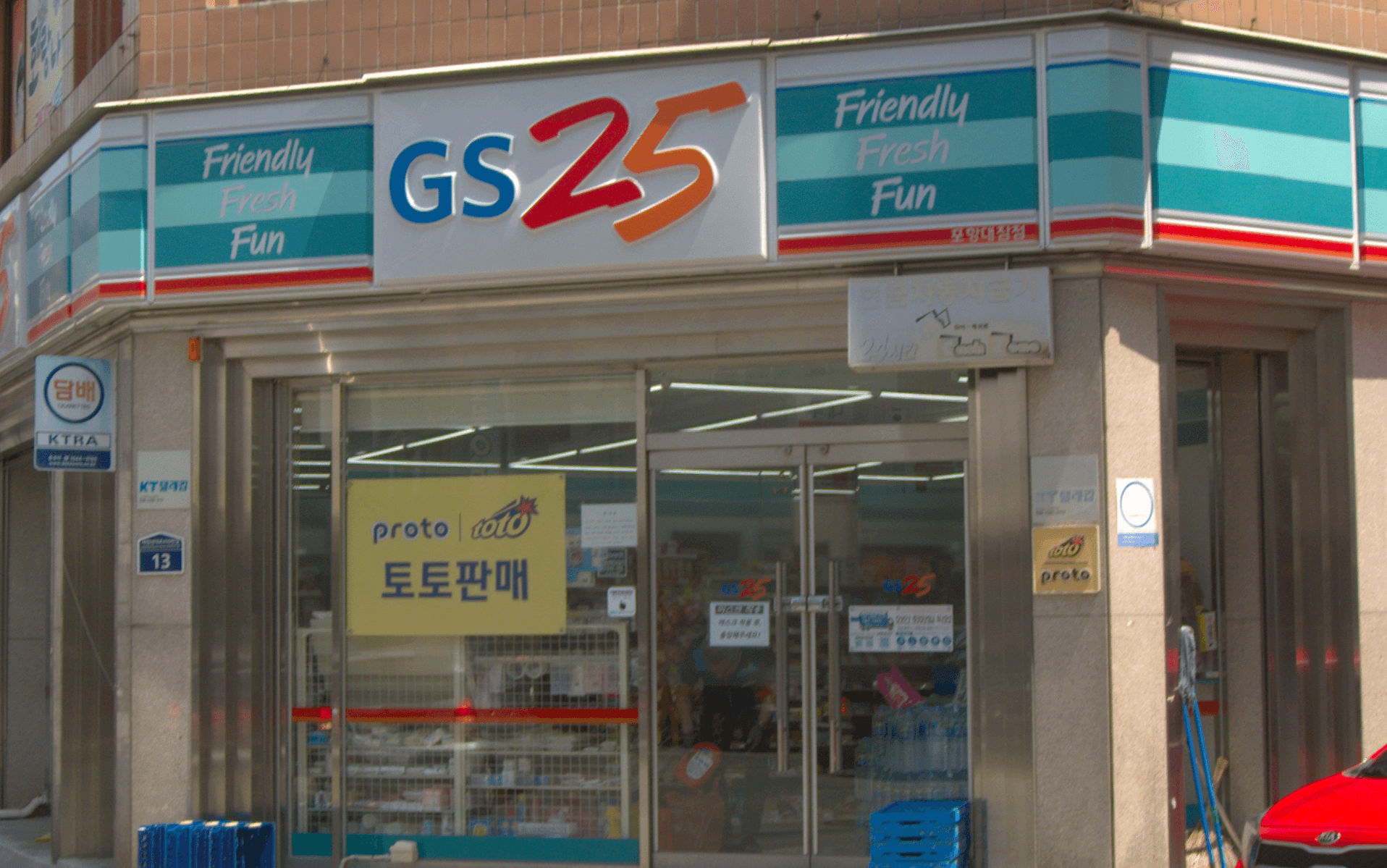}&\hspace{-3.5mm}
    \includegraphics[width=0.24\textwidth]{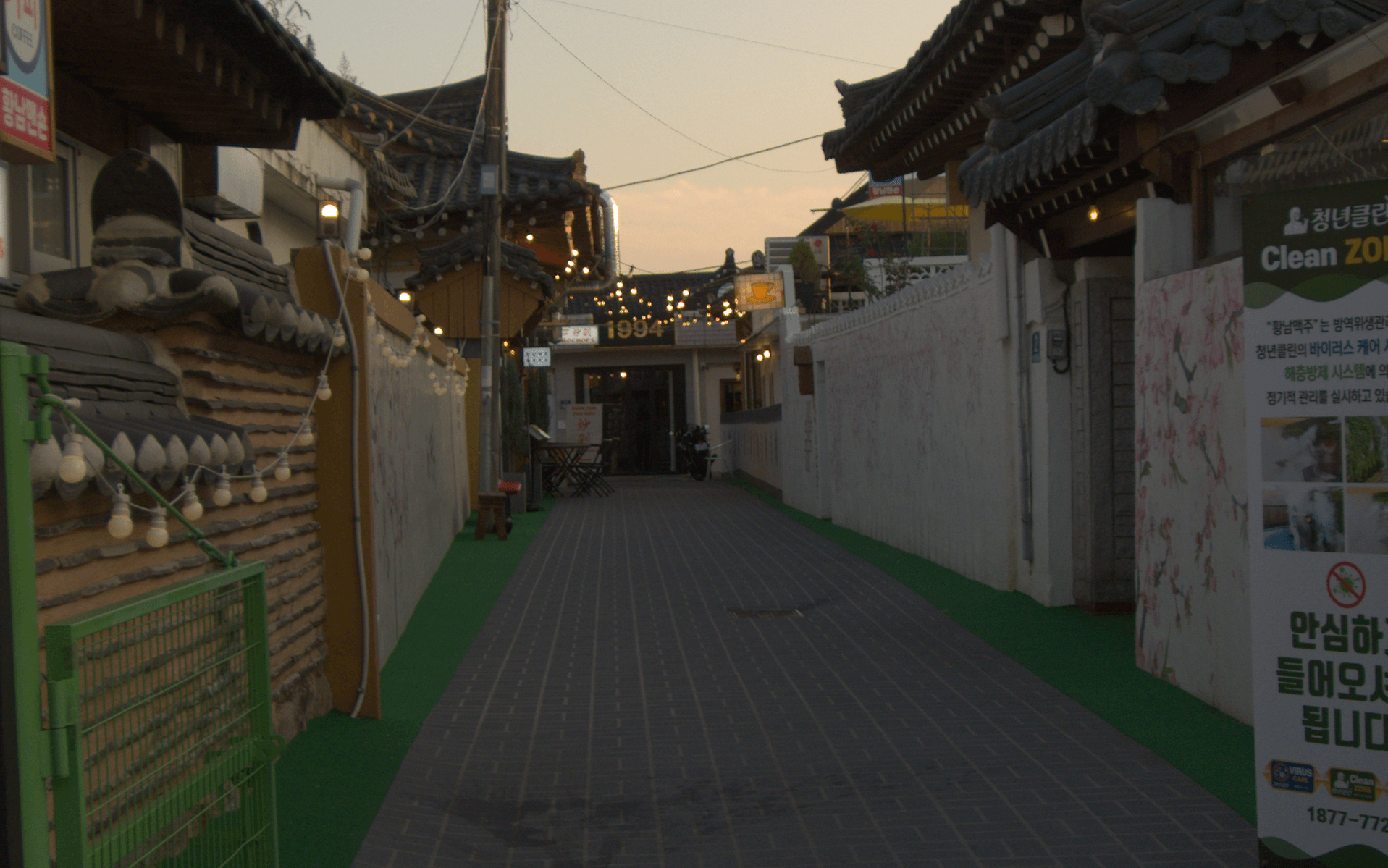}\\

    \end{tabular}
   \caption{Blurred and clear training data pairs in the RSBlur dataset~\cite{RSBlur}.}
    \label{fig:dataset_1}
\vspace{-3mm}
\end{figure*}

\begin{figure*}[!ht]
\footnotesize
\centering
    \begin{tabular}{cccc}
    \includegraphics[width=0.24\textwidth]{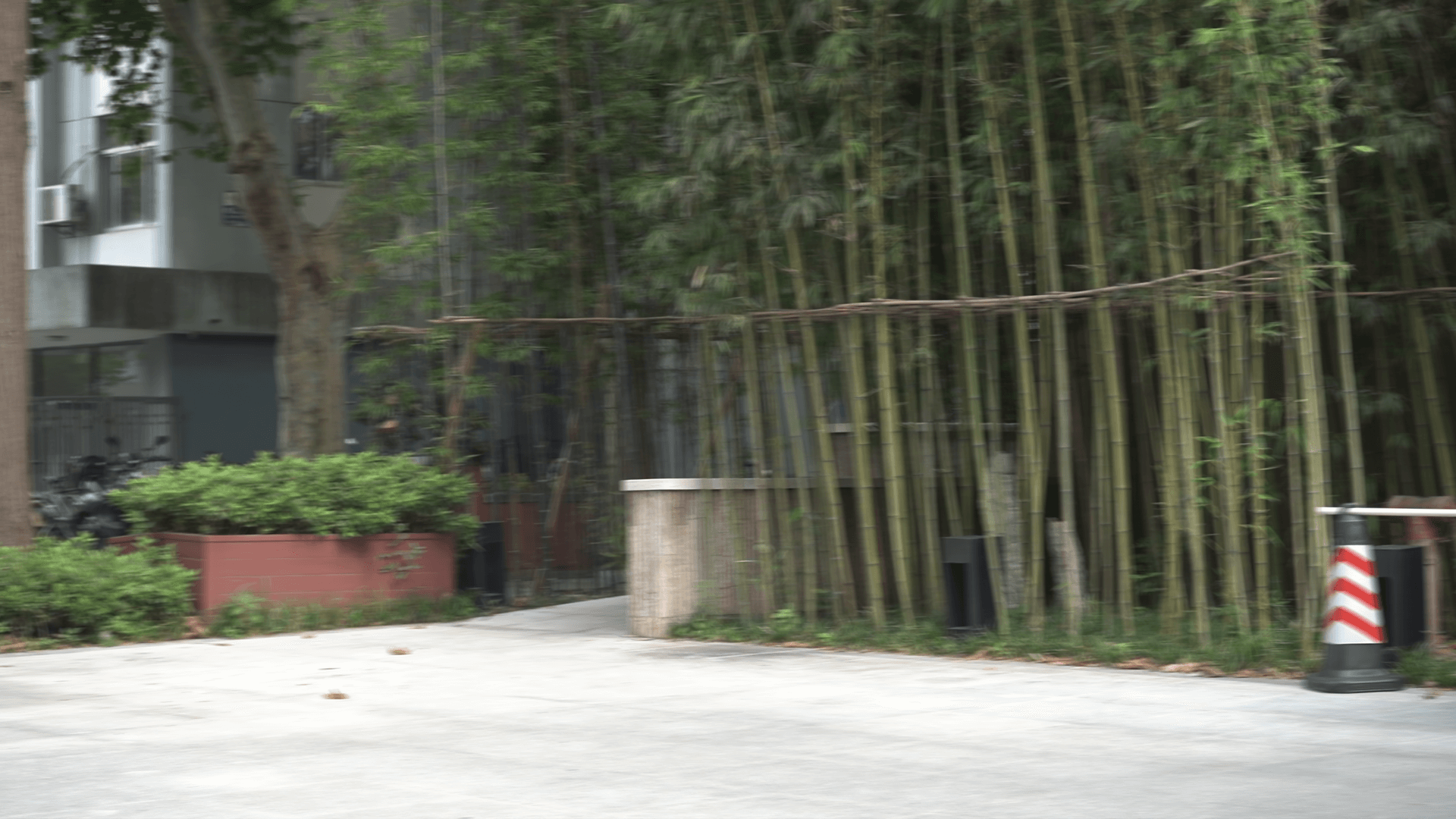}&\hspace{-3.5mm}
    \includegraphics[width=0.24\textwidth]{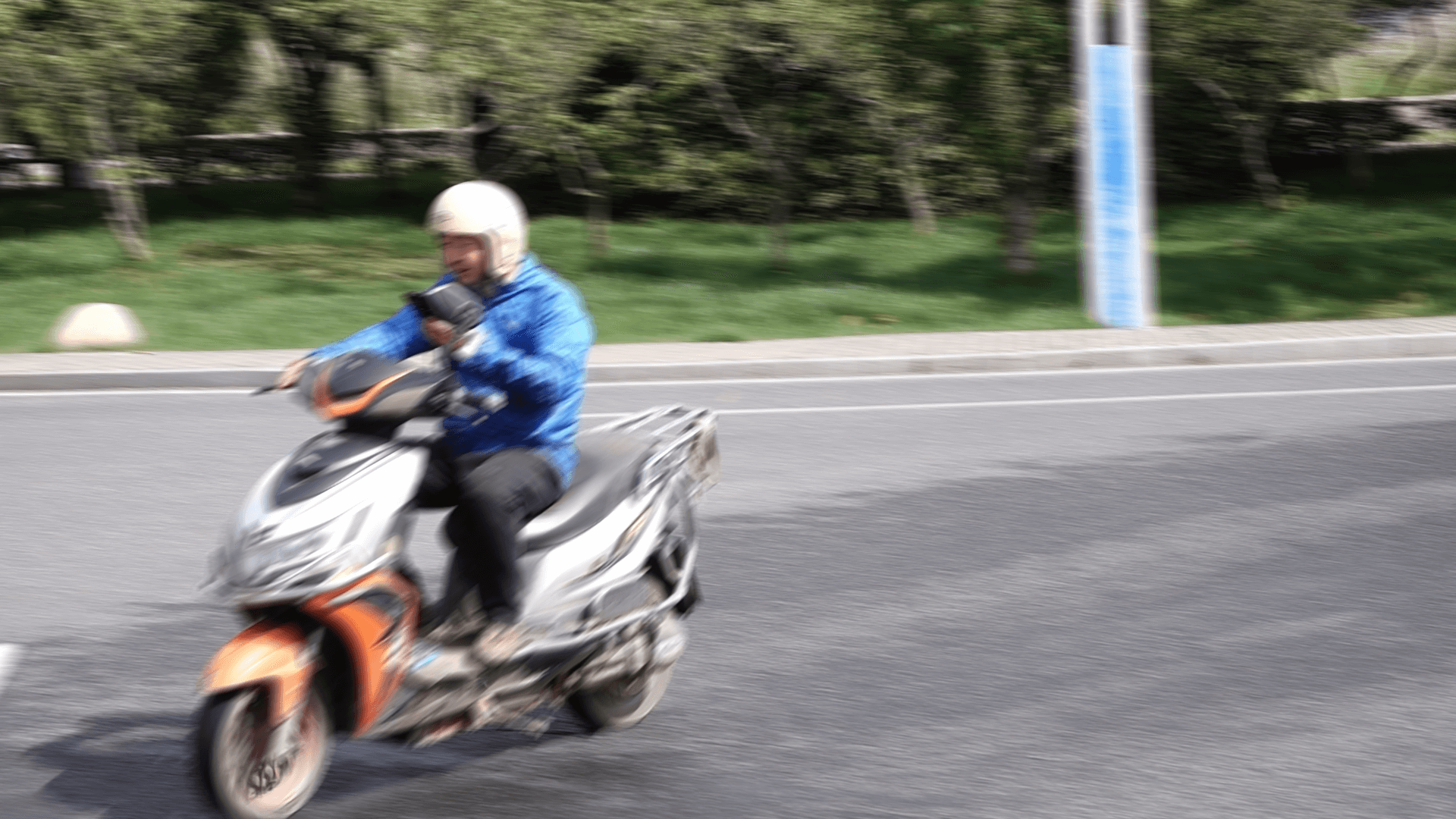}&\hspace{-4.5mm}
    \includegraphics[width=0.24\textwidth]{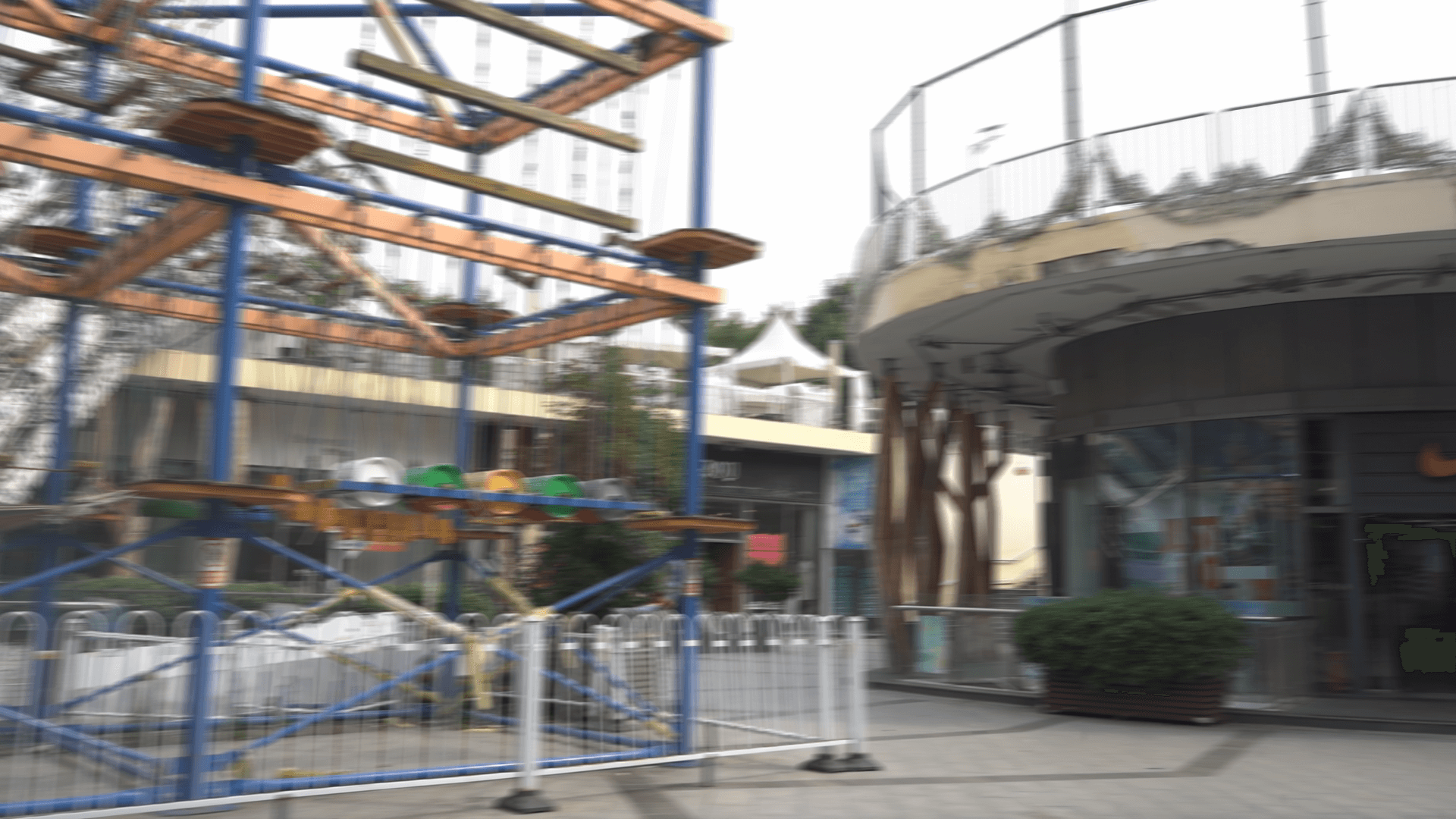}&\hspace{-3.5mm}
    \includegraphics[width=0.24\textwidth]{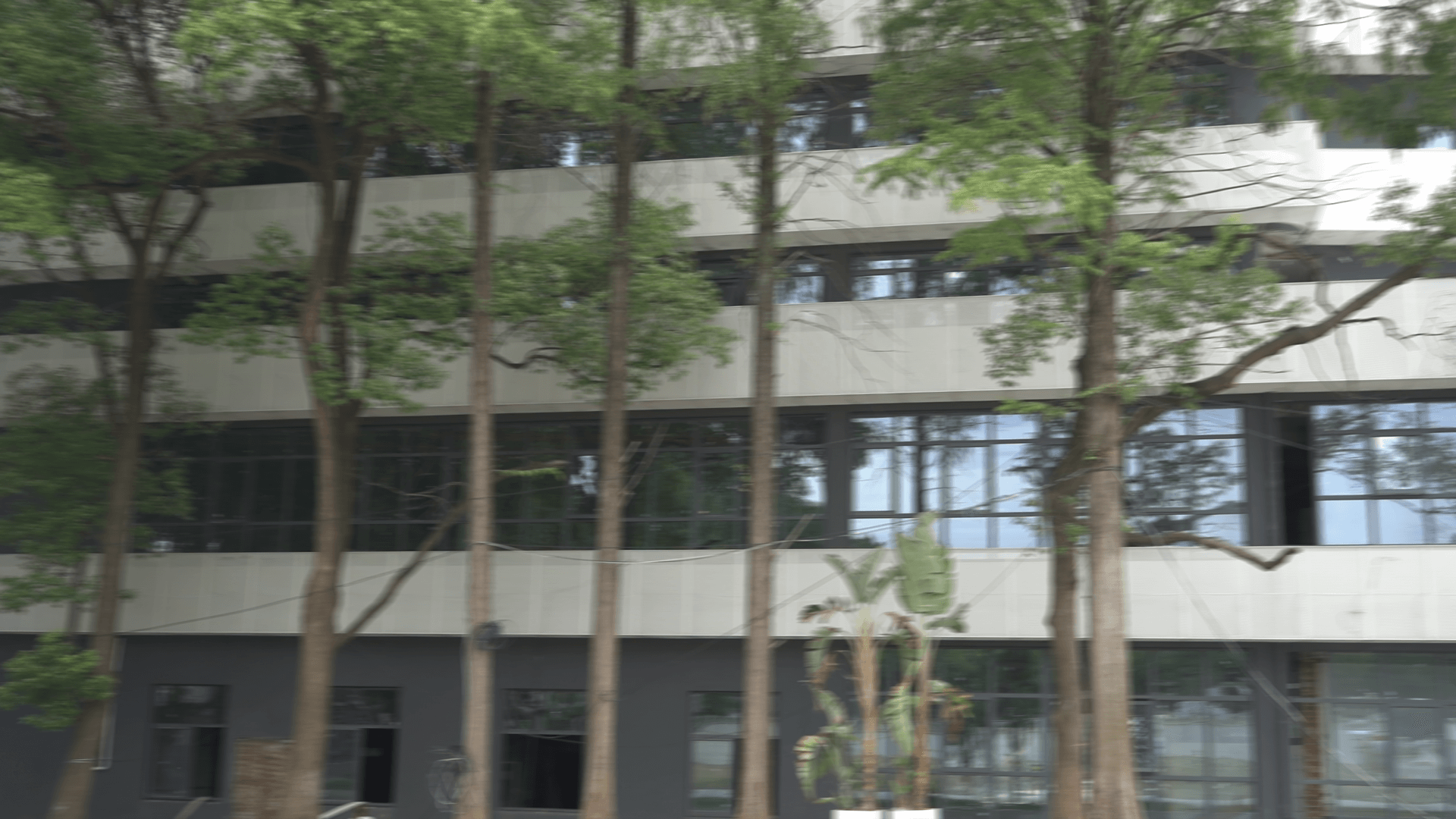}\\
    \includegraphics[width=0.24\textwidth]{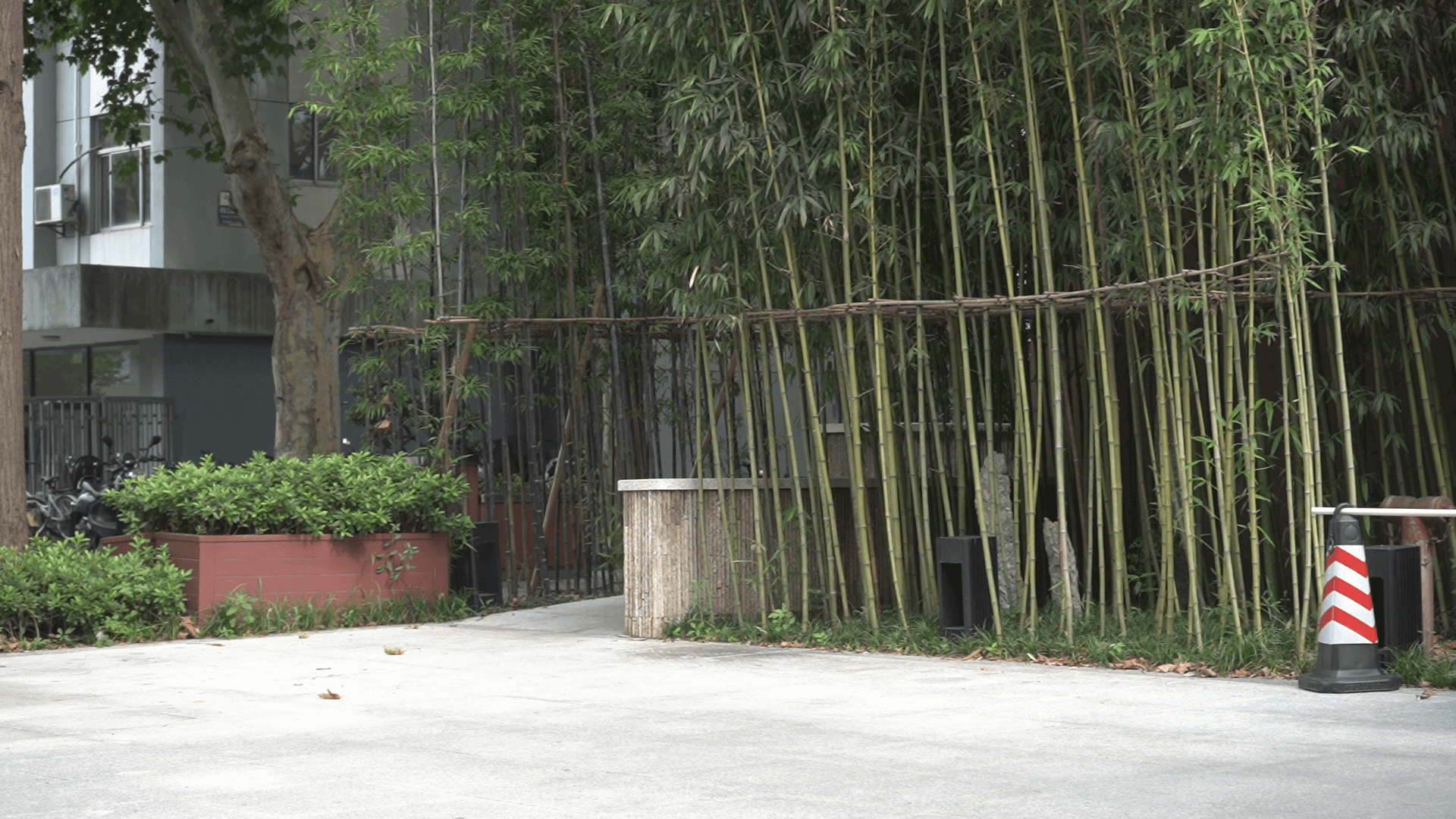}&\hspace{-3.5mm}
    \includegraphics[width=0.24\textwidth]{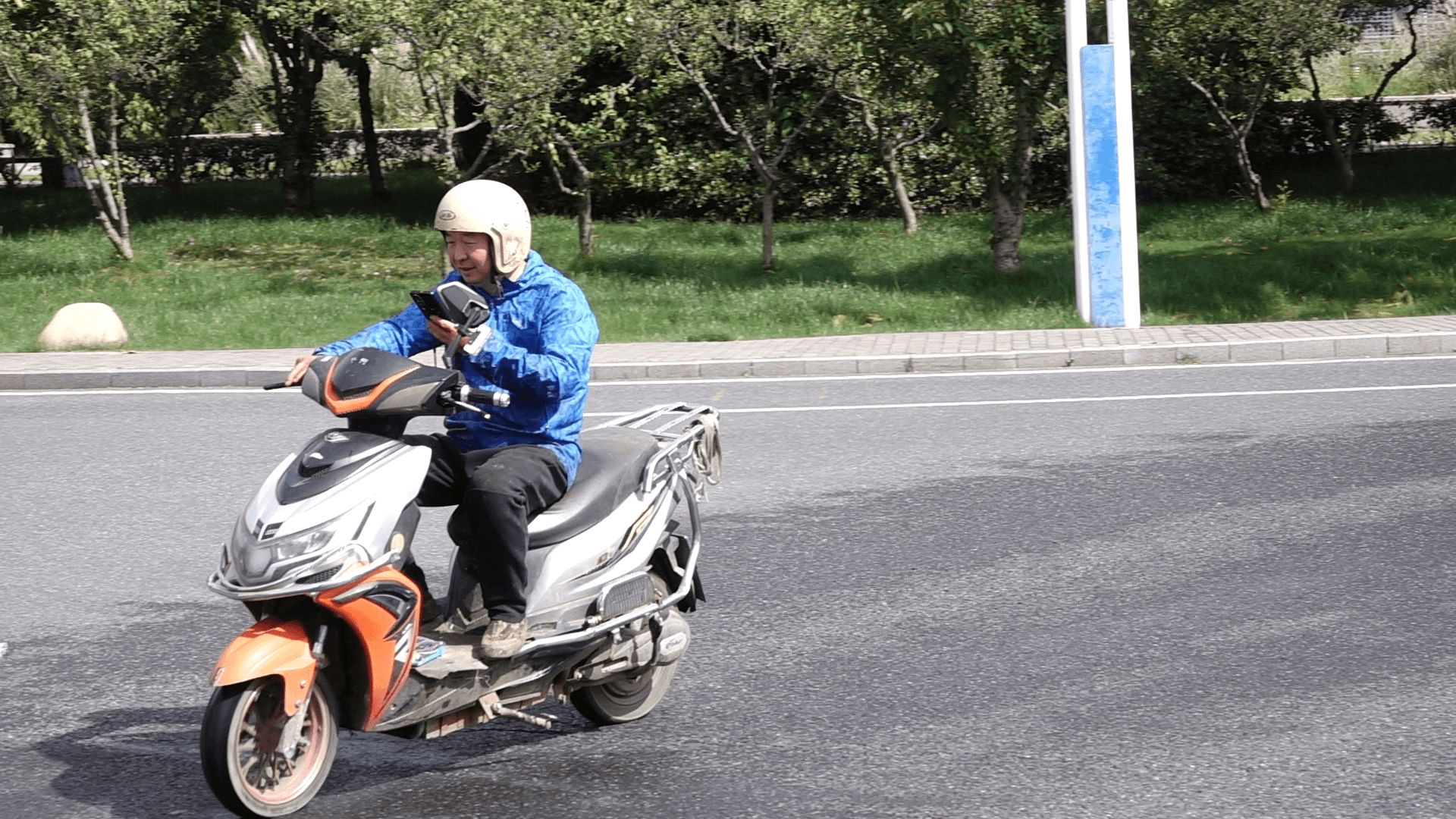}&\hspace{-4.5mm}
    \includegraphics[width=0.24\textwidth]{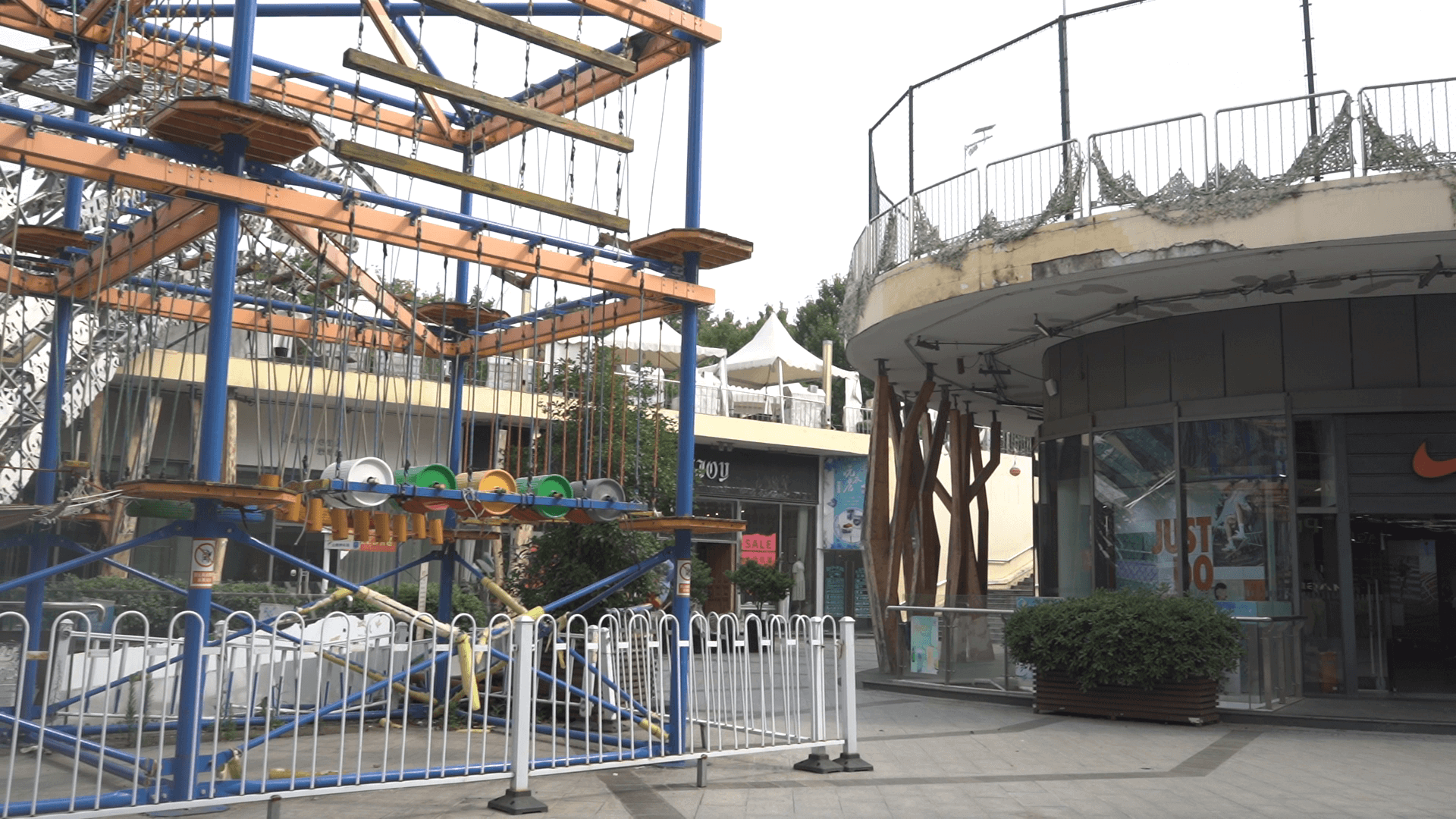}&\hspace{-3.5mm}
    \includegraphics[width=0.24\textwidth]{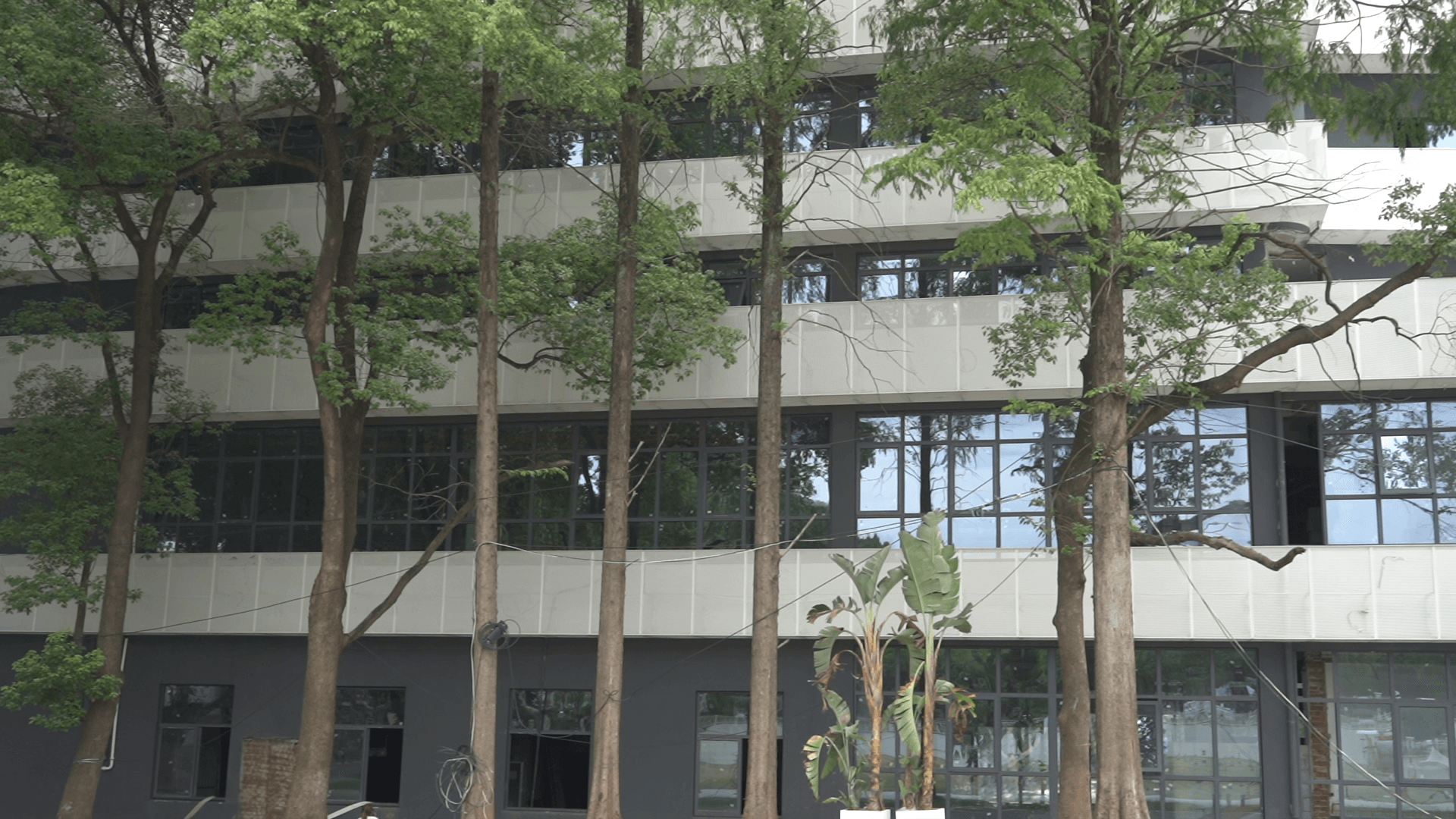}\\

    \end{tabular}
\vspace{-2mm}
   \caption{Blurred and clear training data pairs synthesized using consecutive video frames.}
    \label{fig:dataset_2}
\vspace{-3mm}
\end{figure*}

\begin{figure*}[!ht]
\footnotesize
\centering
    \begin{tabular}{cccccccc}
    \includegraphics[width=0.123\textwidth]{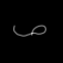}&\hspace{-4.5mm}
    \includegraphics[width=0.123\textwidth]{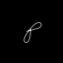}&\hspace{-4.5mm}
    \includegraphics[width=0.123\textwidth]{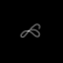}&\hspace{-4.5mm}
    \includegraphics[width=0.123\textwidth]{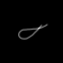}&\hspace{-4.5mm}
    \includegraphics[width=0.123\textwidth]{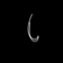}&\hspace{-4.5mm}
    \includegraphics[width=0.123\textwidth]{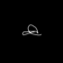}&\hspace{-4.5mm}
    \includegraphics[width=0.123\textwidth]{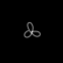}&\hspace{-4.5mm}
    \includegraphics[width=0.123\textwidth]{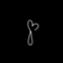}\\
    \includegraphics[width=0.123\textwidth]{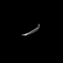}&\hspace{-4.5mm}
    \includegraphics[width=0.123\textwidth]{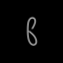}&\hspace{-4.5mm}
    \includegraphics[width=0.123\textwidth]{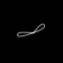}&\hspace{-4.5mm}
    \includegraphics[width=0.123\textwidth]{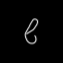}&\hspace{-4.5mm}
    \includegraphics[width=0.123\textwidth]{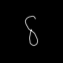}&\hspace{-4.5mm}
    \includegraphics[width=0.123\textwidth]{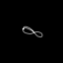}&\hspace{-4.5mm}
    \includegraphics[width=0.123\textwidth]{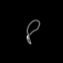}&\hspace{-4.5mm}
    \includegraphics[width=0.123\textwidth]{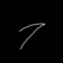}\\
        \hspace{-3mm}
    \end{tabular}
\vspace{-5mm}
   \caption{Motion blur kernels generated by~\cite{PSF}.}
    \label{fig:kernel}
\vspace{-3mm}
\end{figure*}

\begin{figure*}[!ht]
\footnotesize
\centering
    \begin{tabular}{ccccccccc}
    \includegraphics[width=0.123\textwidth]{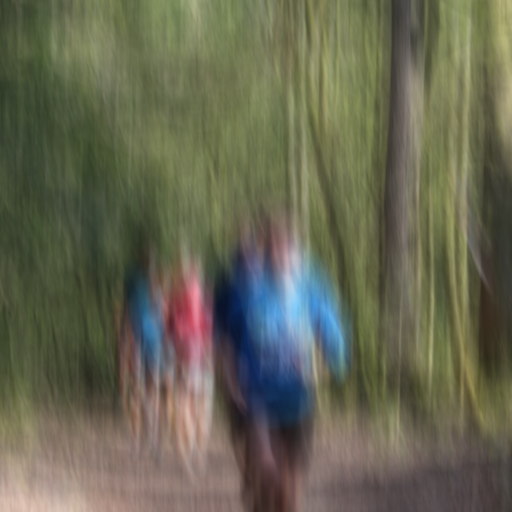}&\hspace{-4.5mm}
    \includegraphics[width=0.123\textwidth]{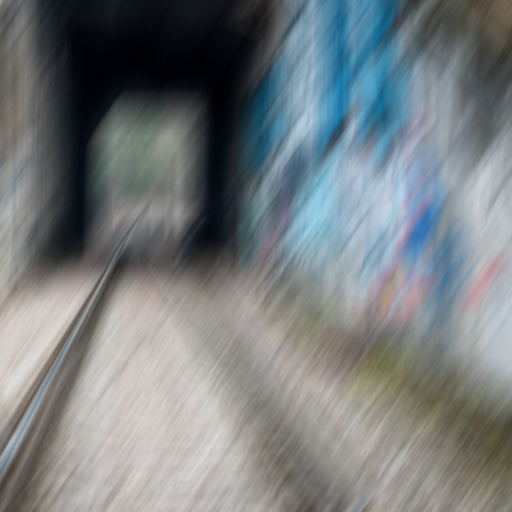}&\hspace{-4.5mm}
    \includegraphics[width=0.123\textwidth]{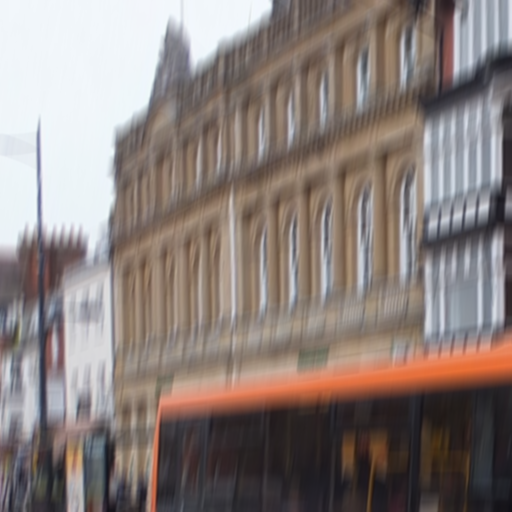}&\hspace{-4.5mm}
    \includegraphics[width=0.123\textwidth]{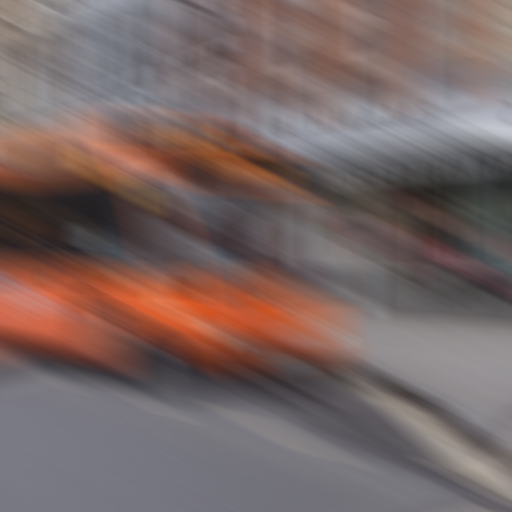}&\hspace{-4.5mm}
    \includegraphics[width=0.123\textwidth]{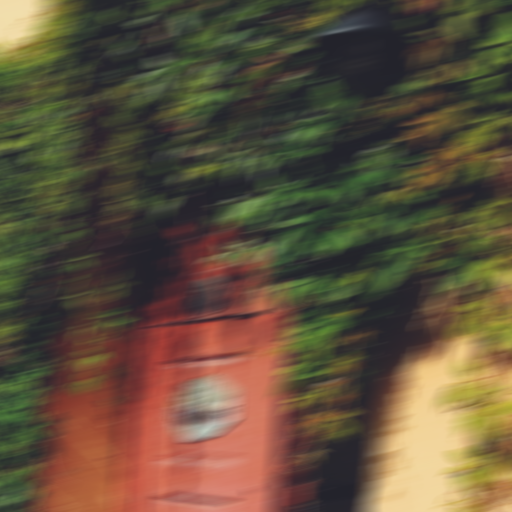}&\hspace{-4.5mm}
    \includegraphics[width=0.123\textwidth]{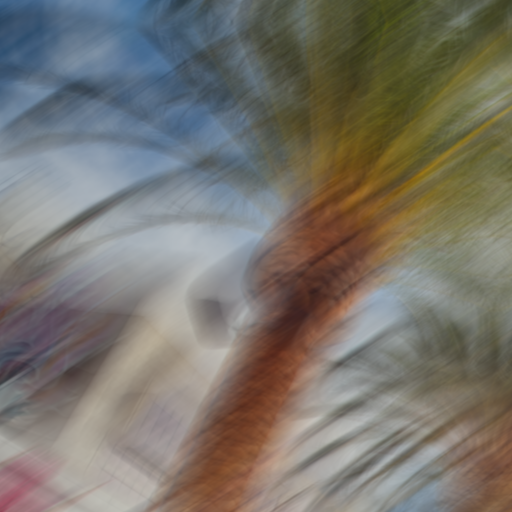}&\hspace{-4.5mm}
    \includegraphics[width=0.123\textwidth]{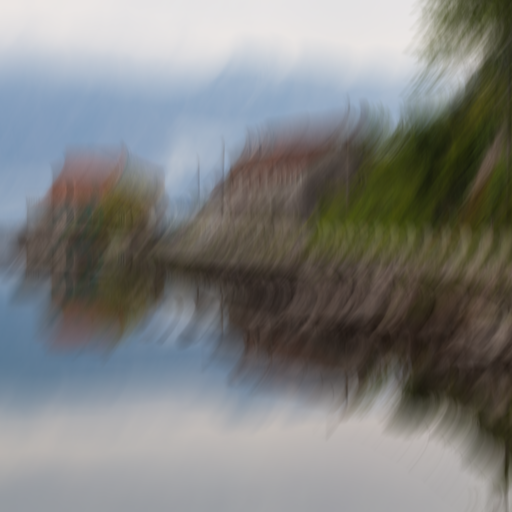}&\hspace{-4.5mm}
    \includegraphics[width=0.123\textwidth]{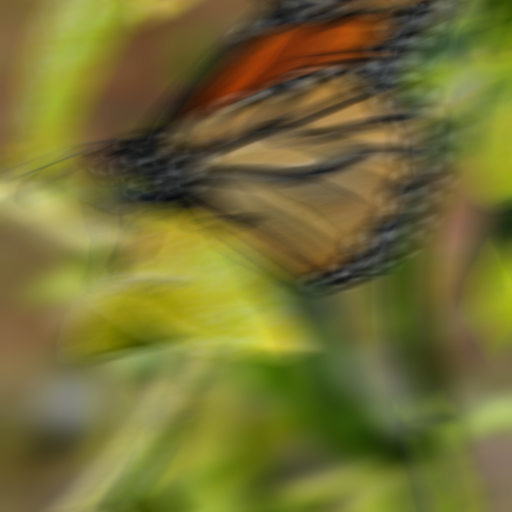}\\
    \includegraphics[width=0.123\textwidth]{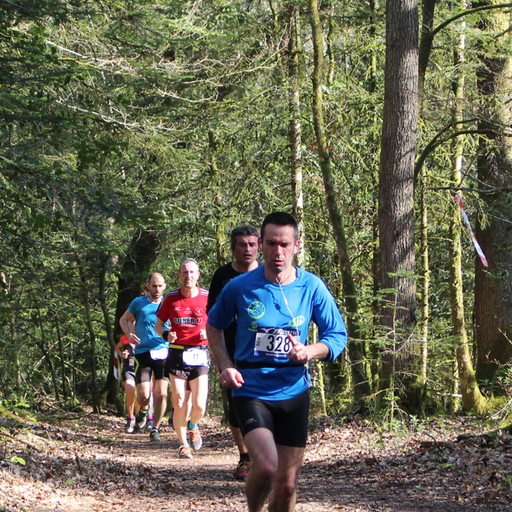}&\hspace{-4.5mm}
    \includegraphics[width=0.123\textwidth]{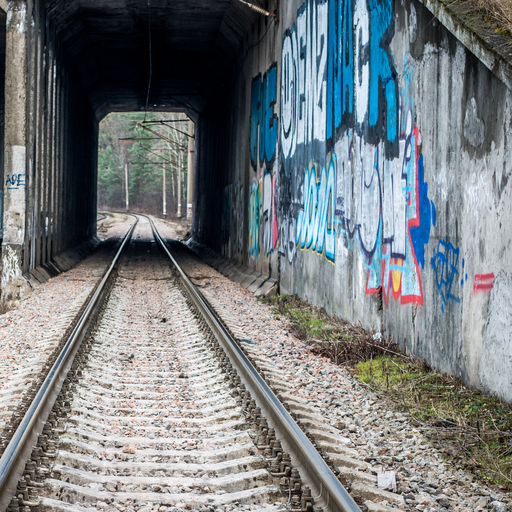}&\hspace{-4.5mm}
    \includegraphics[width=0.123\textwidth]{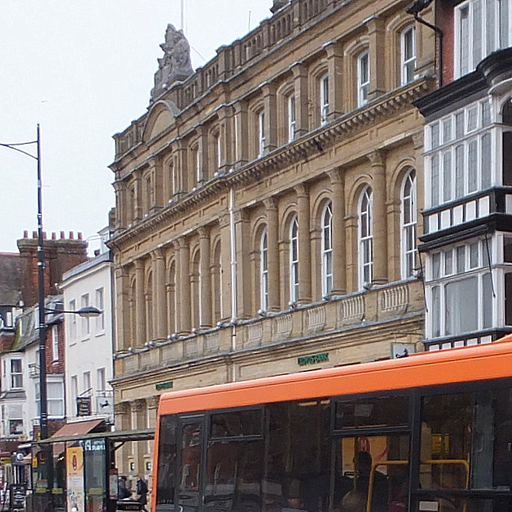}&\hspace{-4.5mm}
    \includegraphics[width=0.123\textwidth]{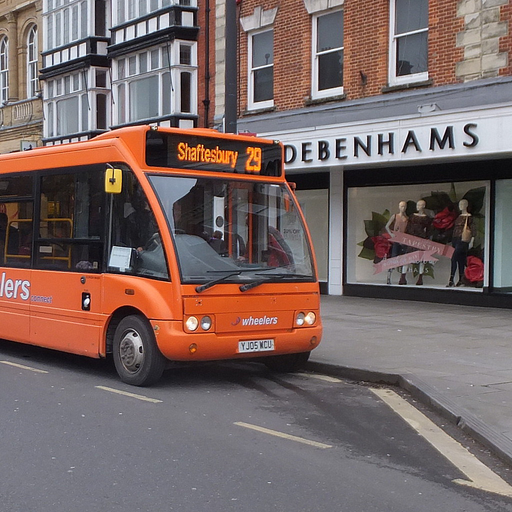}&\hspace{-4.5mm}
    \includegraphics[width=0.123\textwidth]{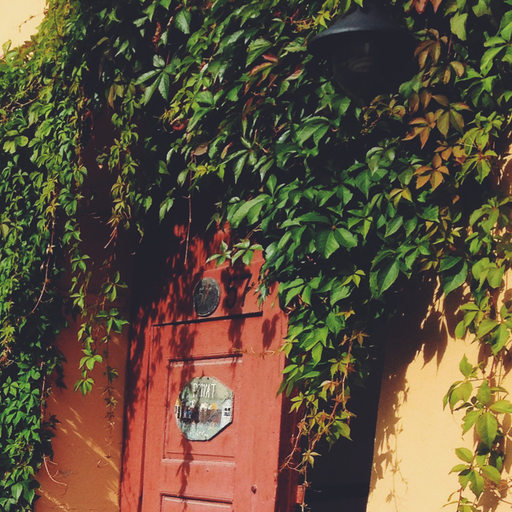}&\hspace{-4.5mm}
    \includegraphics[width=0.123\textwidth]{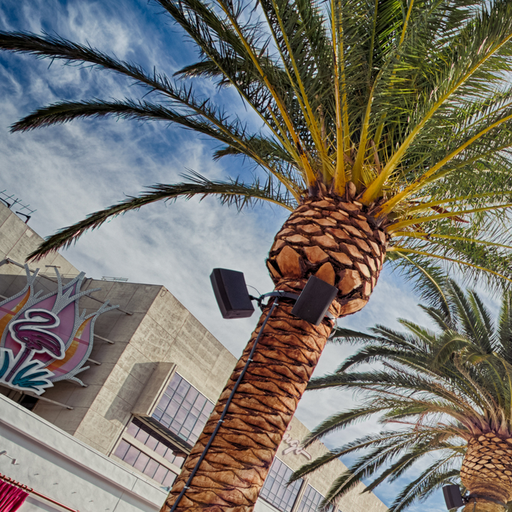}&\hspace{-4.5mm}
    \includegraphics[width=0.123\textwidth]{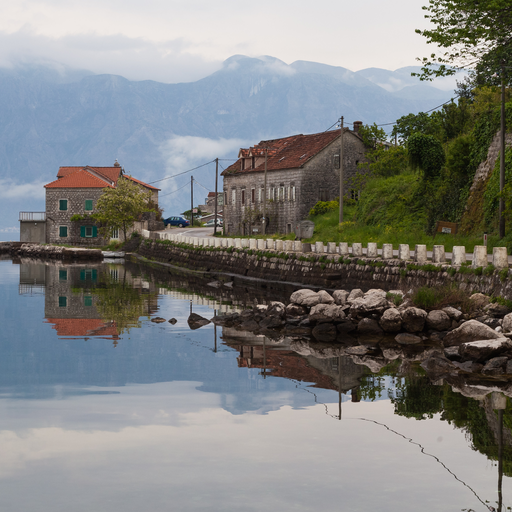}&\hspace{-4.5mm}
    \includegraphics[width=0.123\textwidth]{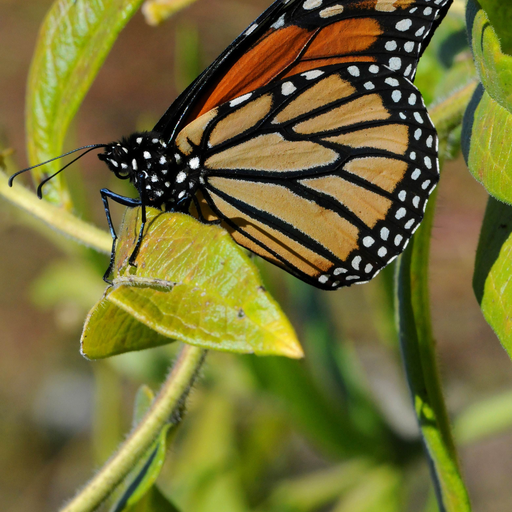}\\
    \end{tabular}
\vspace{-2mm}
   \caption{Training data pairs of globally uniform blurred images and clear images.}
    \label{fig:Uniform}
\vspace{-3mm}
\end{figure*}

\begin{figure*}[!ht]
\footnotesize
\centering
    \begin{tabular}{ccccccccc}
    \includegraphics[width=0.123\textwidth]{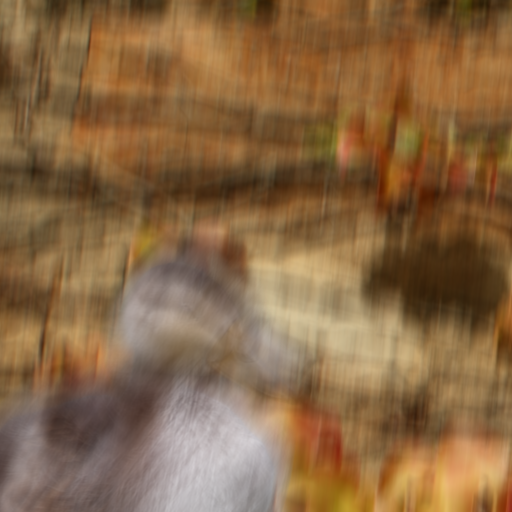}&\hspace{-4.5mm}
    \includegraphics[width=0.123\textwidth]{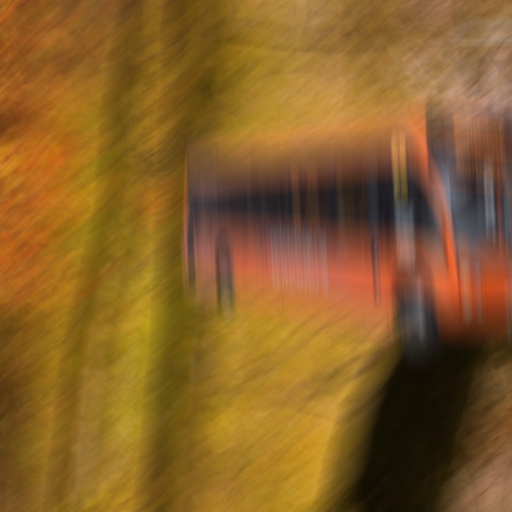}&\hspace{-4.5mm}
    \includegraphics[width=0.123\textwidth]{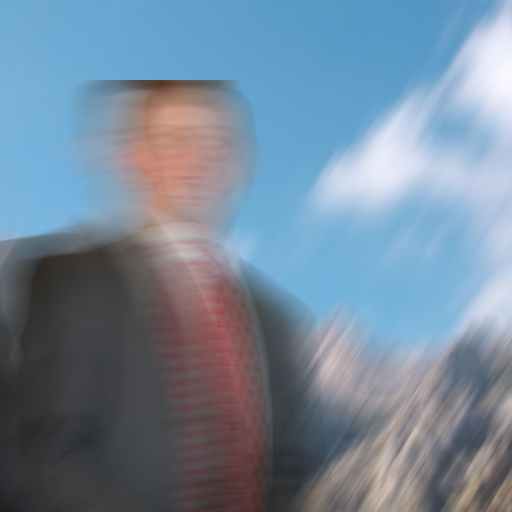}&\hspace{-4.5mm}
    \includegraphics[width=0.123\textwidth]{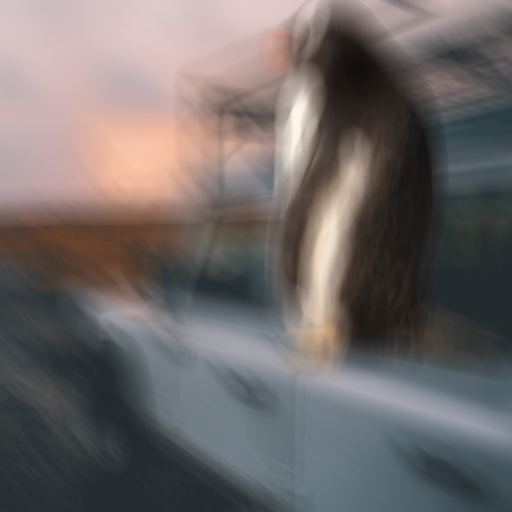}&\hspace{-4.5mm}
    \includegraphics[width=0.123\textwidth]{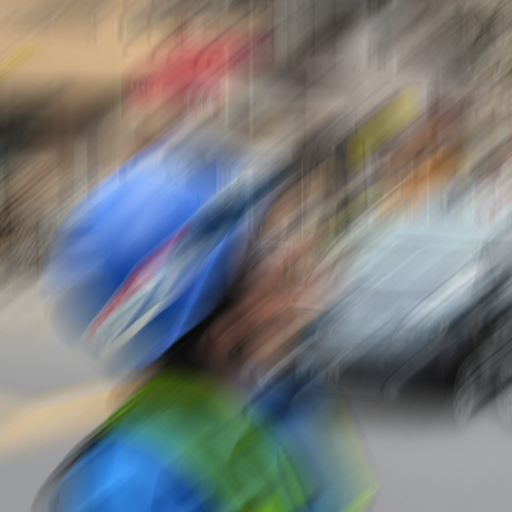}&\hspace{-4.5mm}
    \includegraphics[width=0.123\textwidth]{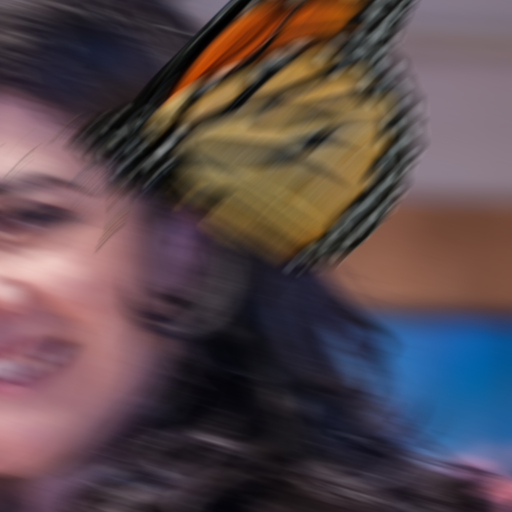}&\hspace{-4.5mm}
    \includegraphics[width=0.123\textwidth]{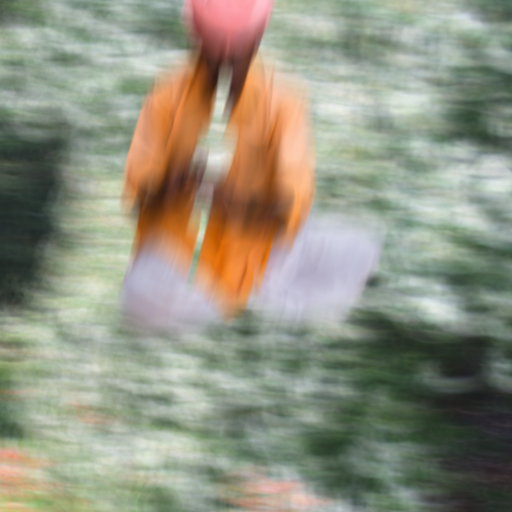}&\hspace{-4.5mm}
    \includegraphics[width=0.123\textwidth]{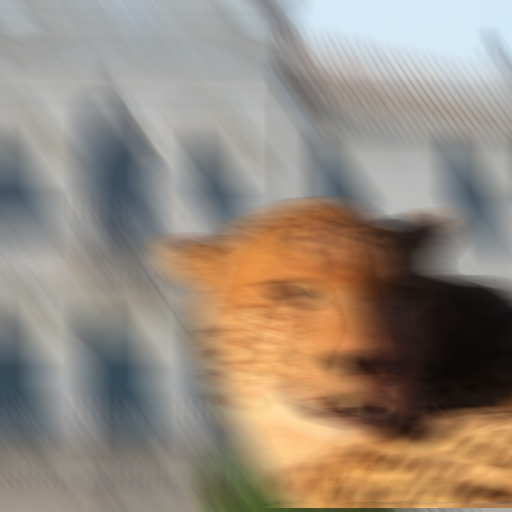}\\
    \includegraphics[width=0.123\textwidth]{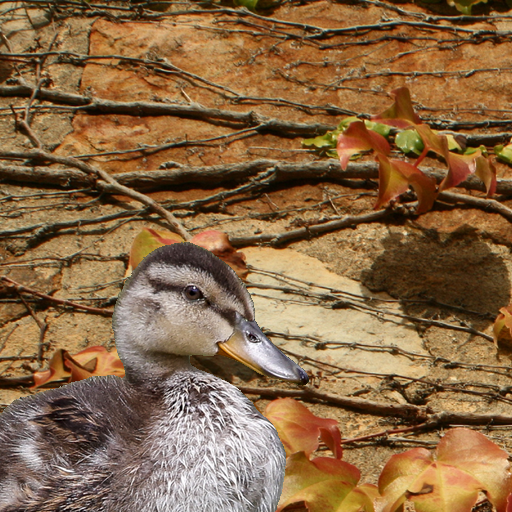}&\hspace{-4.5mm}
    \includegraphics[width=0.123\textwidth]{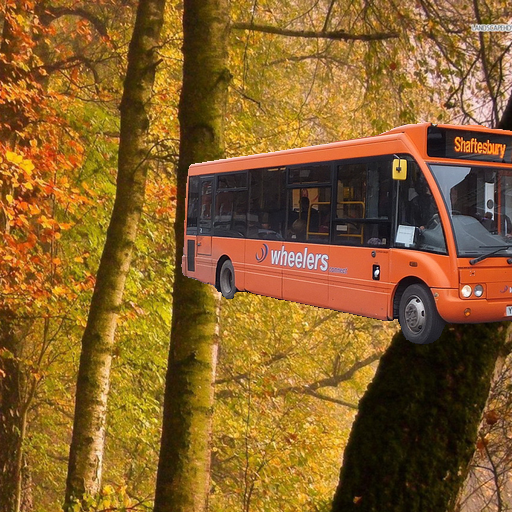}&\hspace{-4.5mm}
    \includegraphics[width=0.123\textwidth]{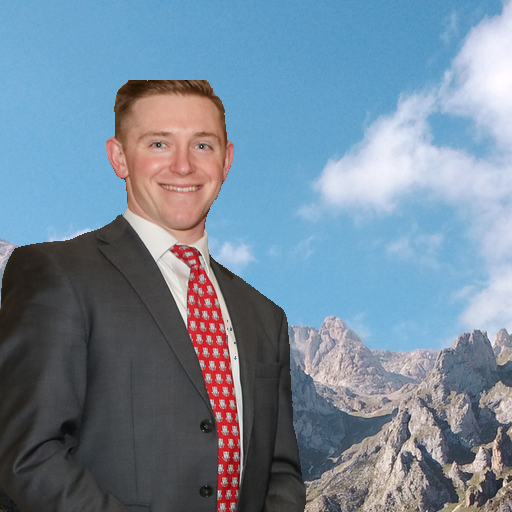}&\hspace{-4.5mm}
    \includegraphics[width=0.123\textwidth]{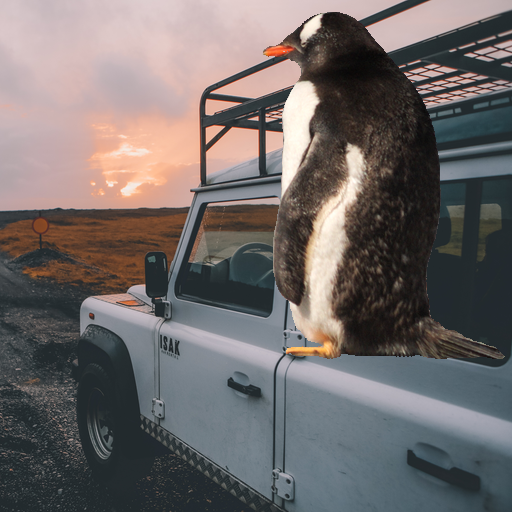}&\hspace{-4.5mm}
    \includegraphics[width=0.123\textwidth]{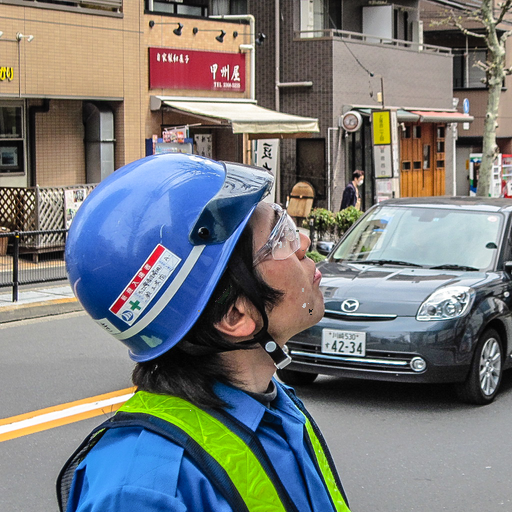}&\hspace{-4.5mm}
    \includegraphics[width=0.123\textwidth]{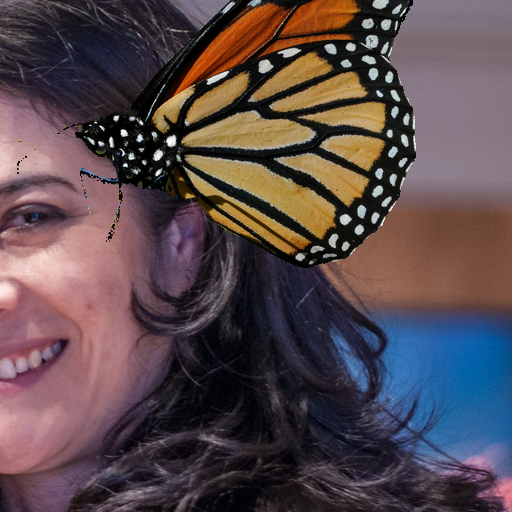}&\hspace{-4.5mm}
    \includegraphics[width=0.123\textwidth]{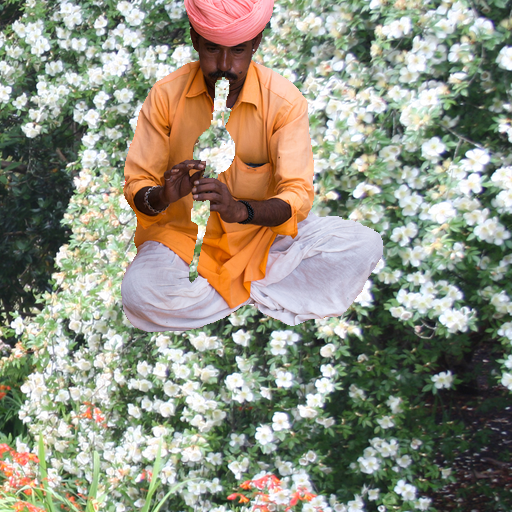}&\hspace{-4.5mm}
    \includegraphics[width=0.123\textwidth]{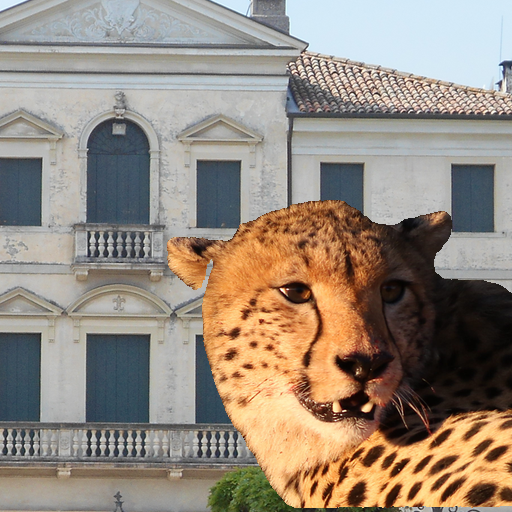}\\
    \end{tabular}
\vspace{-4mm}
   \caption{Training data pairs of globally non-uniform blurred images and clear images.}
    \label{fig:SAM}
\vspace{-3mm}
\end{figure*}

%
Our training dataset consists of three parts: (1) Existing deblurring datasets, including MC-Blur~\cite{MC-blur} and RSBlur~\cite{RSBlur}, which contain high-resolution blur-sharp data pairs(approximately 100,000 images). Some of their training data pairs are illustrated in Figure~\ref{fig:dataset_1}.
(2) We capture some high-frame-rate high-definition video clips, generating blurred and clear data pairs(approximately 200,000 images) using the same strategy as REDS~\cite{REDS}. We present some of our clear and blurred data pairs in Figure~\ref{fig:dataset_2}.
(3) We collected a large number of high-definition images as ground truth (approximately 200,000 images) and generated various motion blur kernels to synthesize corresponding blurred images. 
Specifically, using the method proposed by~\cite{PSF}, we generated 100,000 motion blur kernels (some of which are presented in Figure~\ref{fig:kernel}). 
The size of the motion kernel is $63\times63$, and we randomly generate motion trajectories within this range, with the length of the trajectory being a random number between 1 and 1000.
We use these kernels and high-definition images to synthesize two types of blurred and clear data pairs: globally uniform blur and non-uniform blur.
For globally uniform blur, we directly use the generated kernel as the blur kernel to convolve with the clear image, obtaining the corresponding blurred image.
For non-uniform blur, we use the SAM2~\cite{sam2} to segment objects in the sharp image and apply different motion blur kernels to different objects to simulate the inconsistent blur caused by the motion of different objects in the real world.
We present the training data pairs for globally uniform blur and non-uniform blur in Figures~\ref{fig:Uniform} and Figure~\ref{fig:SAM}, respectively.

\section{Comparison with More Methods.}
For a fair comparison, we provide a comparison of the results using the official FFTformer training weights on the test dataset and the results after retraining on our own training dataset, as shown in Table~\ref{tab:supp_fftformer}.
Apart from the significant difference on the GoPro dataset, which is expected since the FFTFormer was trained on the GoPro training set while our training dataset does not include GoPro data, our retrained method achieves comparable performance to the original FFTFormer on most metrics across other datasets.
\begin{table}[!t]\footnotesize
\vspace{-2mm}

    \caption{Quantitative evaluations of the original FFTFormer and the FFTformer$^*$ we retrained on both synthetic and real-world benchmarks. The models marked with an asterisk $^*$ indicate that we retrain them on our own training set. The best is marked in \textcolor{red}{red}. For the RWBI and Real Blurry Images datasets, which lack ground truth (GT) data, we evaluate the performance using only no-reference metrics.
    }
    \label{tab:supp_fftformer}
\begin{tabular}{c|c|cccc|ccc|cc}
        \toprule
Dataset                      & Metrics             & FFTformer  & FFTformer$^*$         \\
        \midrule 

\multirow{7}{*}{GoPro}         & PSNR~$\uparrow$   &  \textcolor{red}{34.21}     &    26.86           \\
                             & SSIM~$\uparrow$     & \textcolor{red}{0.9536}     &    0.8357            \\
                             & LPIPS~$\downarrow$  & \textcolor{red}{0.0725}     &    0.1538          \\
                & NIQE~$\downarrow$   &5.0338      &     \textcolor{red}{4.1200}    \\
                             & MUSIQ~$\uparrow$    &\textcolor{red}{46.0739}     &     52.2993   \\
                             & MANIQA~$\uparrow$   &0.5442      &     \textcolor{red}{0.5454}     \\
                             & CLIP-IQA~$\uparrow$ &0.4139      &    \textcolor{red}{0.4360}      \\
                                     \midrule 
\multirow{7}{*}{DVD}         & PSNR~$\uparrow$     & \textcolor{red}{27.29}      &    27.07           \\
                             & SSIM~$\uparrow$     & 0.8426     &    \textcolor{red}{0.8534}      \\
                             & LPIPS~$\downarrow$  & 0.1993     &    \textcolor{red}{0.1628}      \\
                             & NIQE~$\downarrow$   &4.7344      &    \textcolor{red}{3.8562}       \\
                             & MUSIQ~$\uparrow$    &40.0924     &     \textcolor{red}{60.1091} \\
                             & MANIQA~$\uparrow$   &0.5653      &      \textcolor{red}{0.6257}    \\
                             & CLIP-IQA~$\uparrow$ & 0.4454     &     \textcolor{red}{0.5271}    \\
                                     \midrule 
\multirow{7}{*}{Realblur}    & PSNR~$\uparrow$     &\textcolor{red}{29.97}       &    {26.94}                 \\
                             & SSIM~$\uparrow$     &\textcolor{red}{0.9036}      &    {0.8580}                \\
                             & LPIPS~$\downarrow$  &\textcolor{red}{0.0906}      &   0.1411                \\
                             & NIQE~$\downarrow$   & 5.1833     &    \textcolor{red}{ 4.3473}            \\
                             & MUSIQ~$\uparrow$    & 52.2186    &    \textcolor{red}{61.5808}          \\
                             & MANIQA~$\uparrow$   & 0.6151     &    \textcolor{red}{0.6374}          \\
                             & CLIP-IQA~$\uparrow$ &0.5077      &    \textcolor{red}{0.5336}          \\

                                     \midrule 

\multirow{4}{*}{RWBI}        & NIQE~$\downarrow$   &4.9704      &    \textcolor{red}{4.4631}             \\
                             & MUSIQ~$\uparrow$    &43.0404     &    \textcolor{red}{59.6223}           \\
                             & MANIQA~$\uparrow$   &0.5201      &    \textcolor{red}{0.5425}           \\
                             & CLIP-IQA~$\uparrow$ &0.3749      &    \textcolor{red}{0.5413}           \\
                                     \midrule 

\multirow{4}{*}{Real Images} & NIQE~$\downarrow$   & 4.4877    &    \textcolor{red}{3.8520}             \\
                             & MUSIQ~$\uparrow$    & 38.1174   &    \textcolor{red}{52.9290}          \\
                             & MANIQA~$\uparrow$   &\textcolor{red}{ 0.5487}    &     0.5170           \\
                             & CLIP-IQA~$\uparrow$ & 0.4222    & \textcolor{red}{0.5026}            \\
                                     \bottomrule

\end{tabular}
\vspace{-5mm}

\end{table}

\newpage
\section{The specific architecture of the DeblurDiff.}
\textbf{LKPN} is a UNet structure with 4 layers of encoders and decoders, each layer comprising two blocks, which consist of a ResBlock~\cite{resnet} and a self-attention~\cite{Transformer} layer as shown in Figure~\ref{fig: Network}.
The input to the LKPN is formed by concatenating $z_t$ and $z_{lq}$.
Additionally, the time step $t$ is embedded into the same dimension using a linear layer and added before the out\_layer of the ResBlock.
Given an input $z_{lq}$ with a shape of $(c,h,w)$ after encoding with the VAE encoder.
After passing through the LKPN's UNet, the shape of the output features remains $(c,h,w)$.
At the end of the LKPN, a linear layer is used to transform the shape into $(c\times k\times k,h,w)$,
where k represents the size of the spatially variant kernel we estimate. In this paper, we use $k=5$.
\textbf{EAC} is used to restore the latent of the clear image while preserving the input information by utilizing the spatially-variant kernel estimated by the LKPN.
The EAC first reshapes the kernel estimated by the LKPN into $(h,w,ck^2)$.
For each position $(h_i,w_i,c_i)$ of $z_{lq}\in {\mathbb{R}}^{h\times w}$, a spatially kernel $\mathcal{F}_{h_i,w_i,c_i} \in {\mathbb{R}}^{k\times k}$ is applied to the region contered around ${z_{lq}}_{(h_i,w_i,c_i)}$ as follows:
\begin{equation}
\begin{split}
{\hat{z}_{lq}(h_i,w_i,c_i)} &= \mathcal{F}_{h_i,w_i,c_i} * z_{lq,(h_i,w_i,c_i)} \\
&= \sum_{n=-r}^{r}\sum_{m=-r}^{r}\mathcal{F}(h_i,w_i,k^2c_i+kn+m) \\
&\quad \times {z}_{lq}(h_i-n,w_i-m,c_i),
\end{split}
\label{eq: kpn}
\end{equation}
where r=$\dfrac{k-1}{2}$, $*$ denotes convolution operation, $\mathcal{F}$ is the generated filter, ${z}_{lq}(h_i-n,w_i-m,c_i)$ and $\hat{z}_{lq}(h_i,w_i,c_i)$ denote the input features and transformed features, respectively.
\textbf{ControlNet} is used to enhance the capabilities of pre-trained SD models by introducing additional conditional inputs to precisely control the image generation process as shown in Figure~\ref{fig: Network}.
Specifically, ControlNet guides the diffusion process by integrating a conditioning input (e.g., a blurry image ) with the noisy latent representation from the diffusion model. 
Initialized with pre-trained diffusion model weights, ControlNet uses ZeroConv layers (1x1 convolutions initialized with zero weights) to gradually incorporate the conditioning input without disrupting the pre-trained knowledge. 
This allows the model to iteratively refine the latent representation, generating high-quality outputs that align with the provided guidance.
We use the concatenated $z_{lq}$ and $z^s$ obtained from EAC as conditional inputs.
For the first layer, we do not use the weights from the original SD model for initialization, since our input channels differ from those of the original SD. Instead, we used random initialization.
The sampling process of DeblurDiff is shown in Algorithm~\ref{alg:example}.
\begin{algorithm}[tb]
   \caption{DeblurDiff Sampling}
   \label{alg:example}
\begin{algorithmic}
   \STATE {\bfseries Input:} Blurry image $X_B$
   \STATE {\bfseries Output:} Deblurred image $X_D$
\STATE $\mathbf{z}_{lq}=\mathcal{E}(\mathbf{X}_{B})$
\STATE $\mathbf{z}_T\sim \mathcal{N}(0, I)$
   \FOR{$t=T,...,1$}
   \STATE $\mathbf{z} \sim \mathcal{N}(\mathbf{0}, \mathbf{I})$ if $t>1$, else $\mathbf{z}=\mathbf{0}$
   \STATE $k_t = \text{LKPN}(\mathbf{z}_t,\mathbf{Z}_{lq},t)$ 
 \STATE $z^s_t = \text{EAC}(\mathbf{z}_{lq},k_t)$ 
\STATE $\boldsymbol{\epsilon}_{pred,t}=\text{ControlNet}(z^s_t,{z}_{lq},z_t,t)$
\STATE $\mathbf{z}_{t-1}=\frac{1}{\sqrt{\alpha_t}}\left(\mathbf{z}_t-\frac{1-\alpha_t}{\sqrt{1-\bar{\alpha}_t}} \epsilon_{pred,t}\right)+\sigma_t \mathbf{z}$
   \ENDFOR
\end{algorithmic}
\end{algorithm}
\section{Visual comparison of the ablation study.}
\begin{figure*}[!t]
    \centering

 \includegraphics[width=0.98\textwidth]{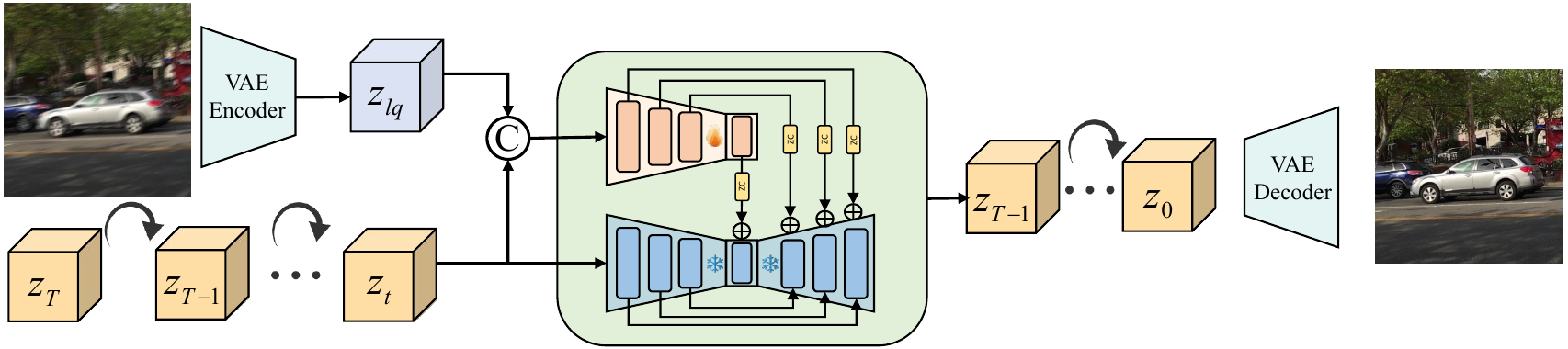}
 \vspace{-2mm}
 \caption{Overall architecture of ControlNet.}
 \label{fig: Network_control}
\end{figure*}

\begin{figure*}[!t]
    \centering
 \includegraphics[width=0.98\textwidth]{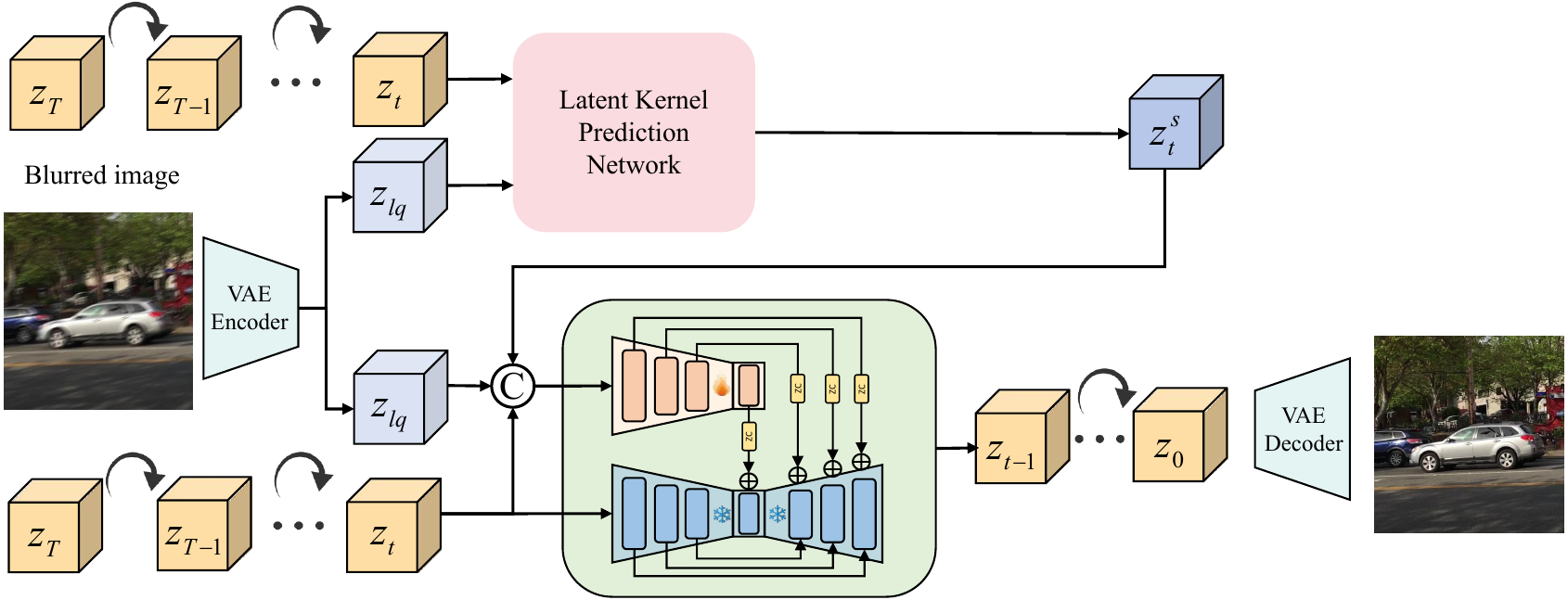}
 \vspace{-2mm}
 \caption{Overall architecture of the baseline ``without\_EAC''.}
 \vspace{-2mm}
 \label{fig: Network_without_EAC}
\end{figure*}

\begin{figure*}[!t]
    \centering
     \vspace{-2mm}
 \includegraphics[width=0.98\textwidth]{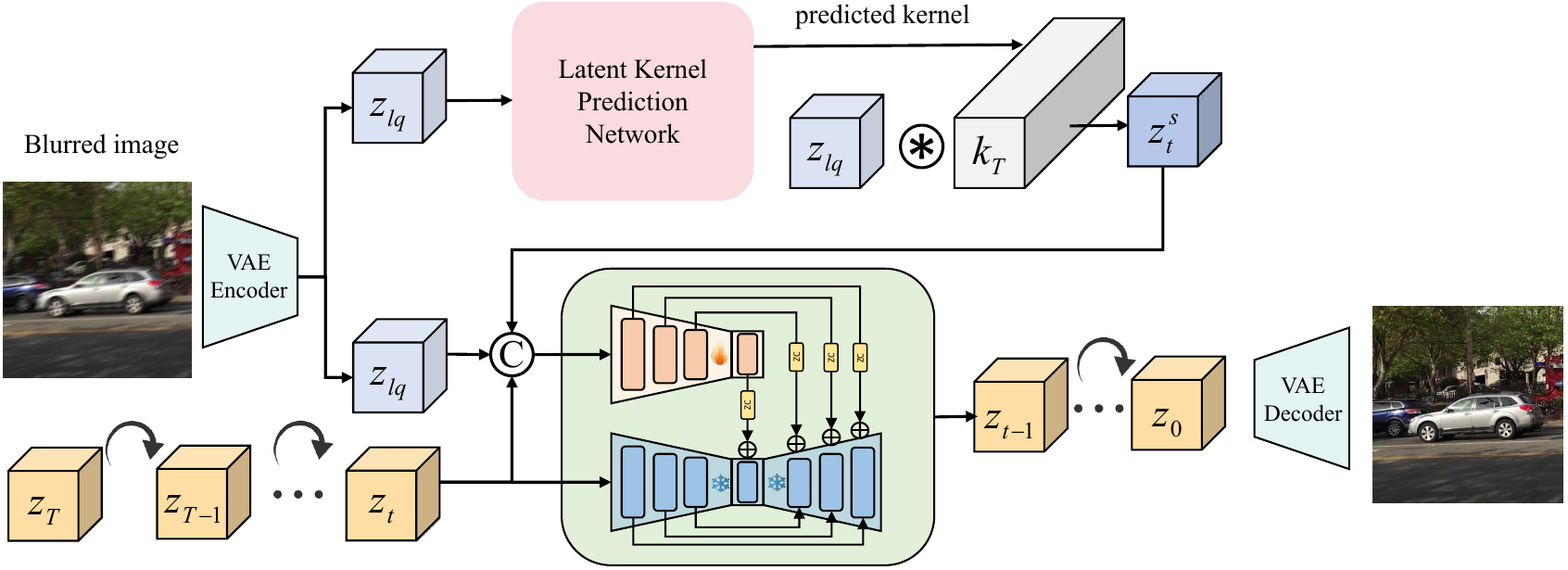}
 \vspace{-2mm}
 \caption{Overall architecture of the baseline ``without\_SD''.}
 \vspace{-2mm}
 \label{fig: Network_without_SD}
\end{figure*}

We provide the specific structural diagrams of the corresponding baselines in the ablation study in Figures\cref{fig: Network_control,fig: Network_without_EAC,fig: Network_without_SD}.
When the LKPN does not utilize the intermediate clear priors from the diffusion process (referred to as w/o SD for LKPN)~\ref{fig: Network_without_SD}, it estimates the deblurring result directly from the blurry image in the latent space throughout the entire iterative process. 

\newpage

\begin{figure*}[!ht]
\footnotesize
\centering
\vspace{-2mm}
\begin{tabular}{cccccc}

\includegraphics[width = 0.245\linewidth]{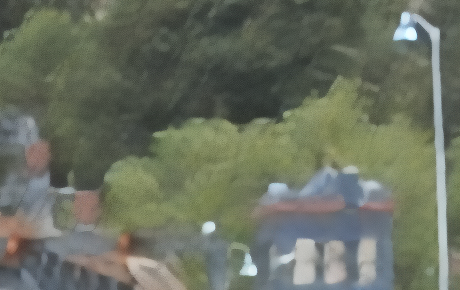}& \hspace{-4mm}
\includegraphics[width = 0.245\linewidth]{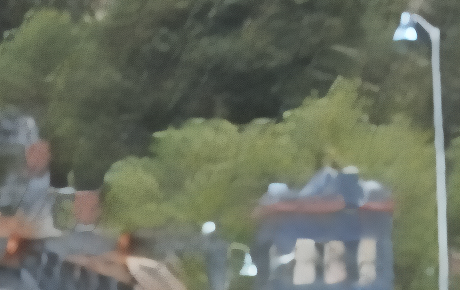}& \hspace{-4mm}
\includegraphics[width = 0.245\linewidth]{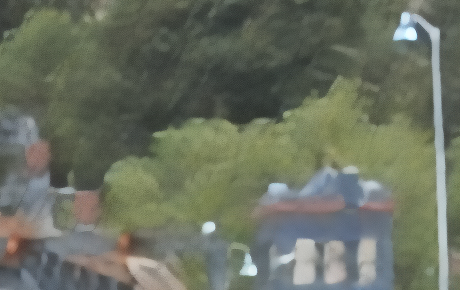}& \hspace{-4mm}
\includegraphics[width = 0.245\linewidth]{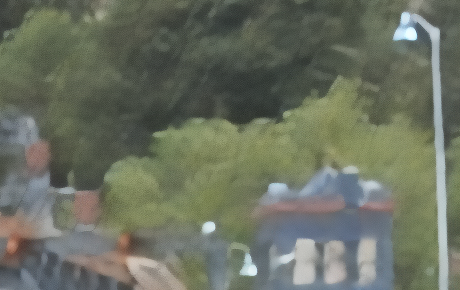}& \\
\multicolumn{4}{c}{(a) Visualization of the LKPN results during the diffusion process in the baseline "w/o SD for LKPN".} \\
\includegraphics[width = 0.245\linewidth]{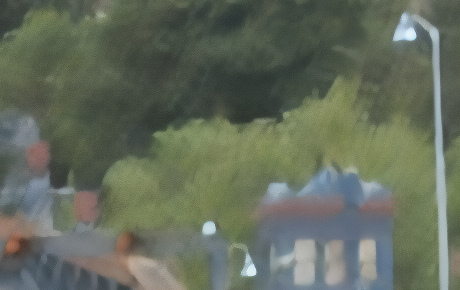}& \hspace{-4mm}
\includegraphics[width = 0.245\linewidth]{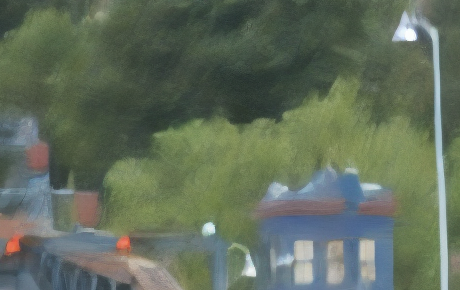}& \hspace{-4mm}
\includegraphics[width = 0.245\linewidth]{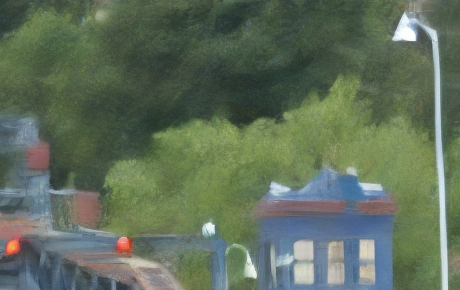}& \hspace{-4mm}
\includegraphics[width = 0.245\linewidth]{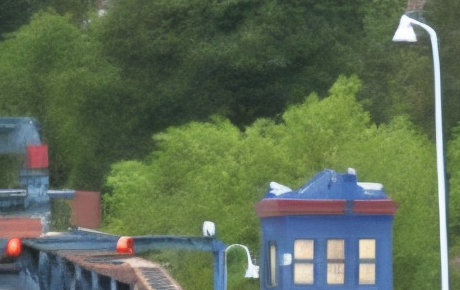}& \\
\multicolumn{4}{c}{(b) Visualization of the LKPN results during the diffusion process in the baseline "w/o EAC".} \\
\includegraphics[width = 0.245\linewidth]{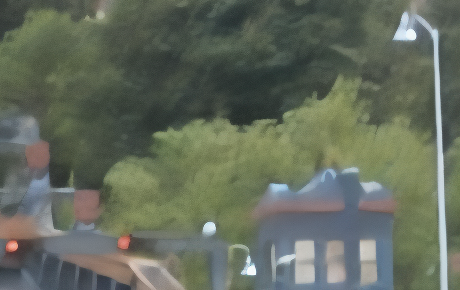}& \hspace{-4mm}
\includegraphics[width = 0.245\linewidth]{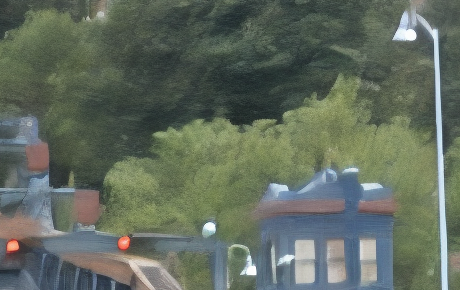}& \hspace{-4mm}
\includegraphics[width = 0.245\linewidth]{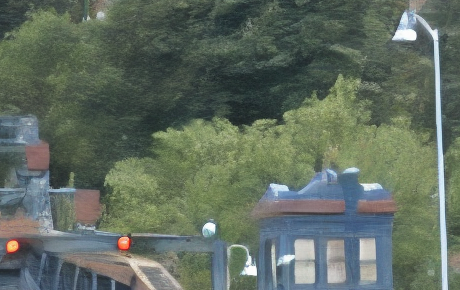}& \hspace{-4mm}
\includegraphics[width = 0.245\linewidth]{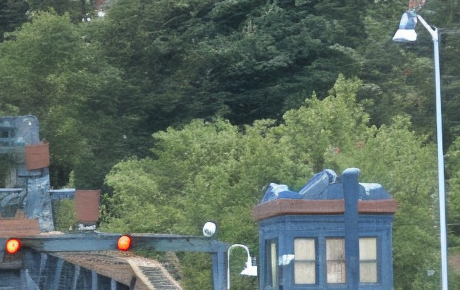}& \\
\multicolumn{4}{c}{(c) Visualization of the LKPN results during the diffusion process in the DeblurDiff.} \\
\end{tabular}
\vspace{-4mm}

\includegraphics[width = 0.95\linewidth]{result/kpn_step/1.png}
\vspace{-7mm}
\caption{Iterative results of LKPN. The arrow represents the iterative diffusion process. To visualize this process, we decode
the features deblurred by the LKPN to the image space using the VAE decoder in each time step. }
\label{fig: ablation_mov_supp}

\end{figure*}

\begin{figure*}[!ht]
\footnotesize
\centering
\vspace{-2mm}
\begin{tabular}{cccccc}

\includegraphics[width = 0.195\linewidth]{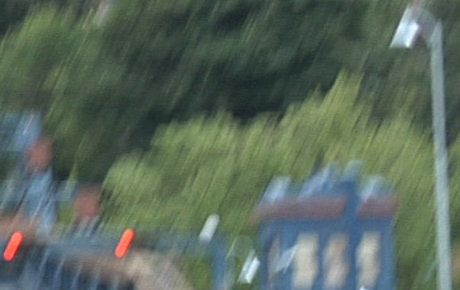}& \hspace{-4mm}
\includegraphics[width = 0.195\linewidth]{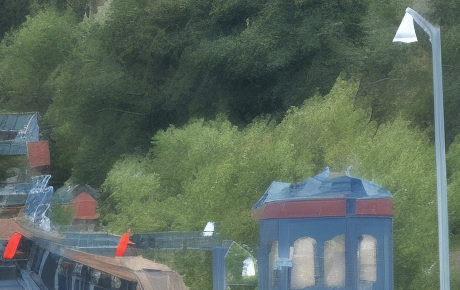}& \hspace{-4mm}
\includegraphics[width = 0.195\linewidth]{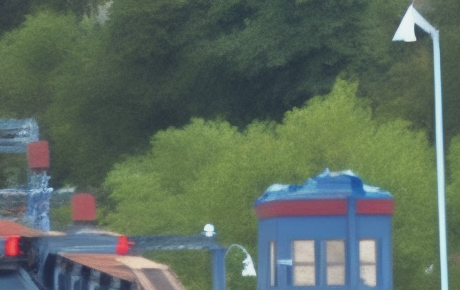}& \hspace{-4mm}
\includegraphics[width = 0.195\linewidth]{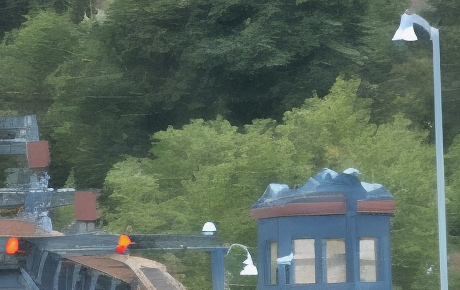}& \hspace{-4mm}
\includegraphics[width = 0.195\linewidth]{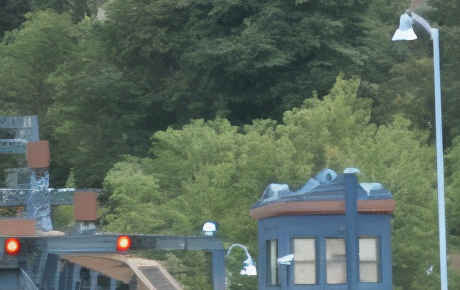}& 
 \\
(a) Input&\hspace{-4mm}(b) ControlNet&\hspace{-4mm}(c) w/o EAC &\hspace{-4mm}(d)w/o SD for LKPN  &\hspace{-4mm}(e)DeblurDiff\\

\vspace{-5mm}
\end{tabular}

\caption{Effectiveness of the proposed DeblurDiff on image deblurring. ControlNet struggles to recover clear structures due to the lack of explicit structural guidance, resulting in blurred outputs. 
Without the Stable Diffusion (SD) priors (referred to as w/o SD for LKPN), the LKPN fails to leverage intermediate clear priors for deblurring, leading to artifacts and inconsistencies in the generated results. 
When the LKPN directly predicts the deblurred result without using EAC (referred to as w/o EAC), the outputs tend to be overly smooth, losing important details and structural information.
In contrast, our method effectively recovers sharp structures and fine details while preserving the input information. By leveraging intermediate clear priors from the diffusion process and adaptively estimating spatially variant kernels through EAC, our approach achieves better deblurring performance, generating visually appealing and realistic results.}
\label{fig: ablation_supp}

\end{figure*}

\onecolumn
Since the LKPN lacks the guidance of intermediate clear priors, its predictions remain static and do not improve over time (Figure~\ref{fig: ablation_mov_supp}(a)). 
As a result, the deblurred results at each step are identical, leading to suboptimal performance and the inability to recover sharp structures or fine details effectively (Figure~\ref{fig: ablation_supp}(d)).
When the LKPN directly predicts the deblurred result in the latent space~\ref{fig: Network_without_EAC} instead of estimating a spatially variant kernel and applying it through the EAC (referred to as w/o EAC), the generated results tend to be overly smooth (Figure~\ref{fig: ablation_mov_supp}(b)). 
This approach fails to adaptively address distinct blur characteristics at each pixel location, resulting in the loss of important structural details and the inability to effectively preserve the input information. 
As a consequence, the deblurred images lack sharpness and fine details, leading to suboptimal visual quality (Figure~\ref{fig: ablation_supp}(c)).
In contrast, our method leverages intermediate clear priors from the diffusion process to iteratively refine the spatially variant kernels estimated by the LKPN. 
This iterative refinement enables the LKPN to progressively improve its predictions, generating increasingly accurate deblurring results (Figure~\ref{fig: ablation_mov_supp}(c)).  
The refined kernels are then applied through the Element-wise Adaptive Convolution (EAC), which adaptively addresses distinct blur characteristics at each pixel location, effectively preserving the input information and recovering sharp structures and fine details (Figure~\ref{fig: ablation_supp}(e)). 

\newpage
\section{Qualitative Comparisons.}
\begin{figure*}[!t]
\footnotesize
\centering
    \begin{tabular}{c c c c c c c}
            \multicolumn{3}{c}{\multirow{5}*[45.6pt]{
            \hspace{-4mm} 
            \vspace{-25mm}
            \includegraphics[width=0.325\linewidth,height=0.236\linewidth]{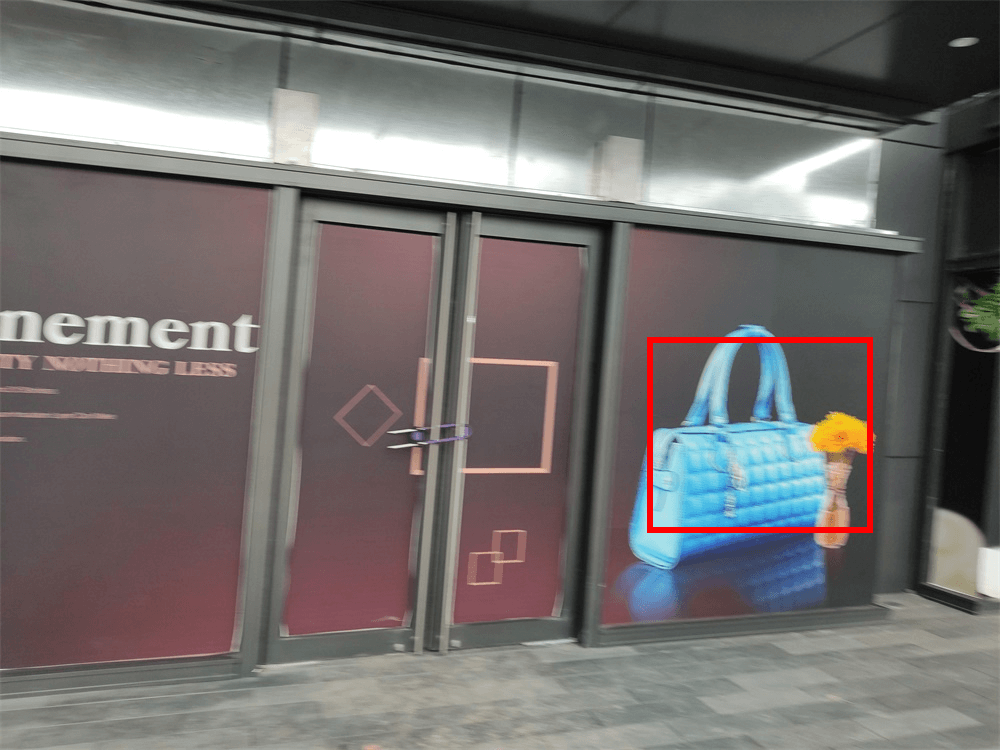}}}
            & \hspace{-4.0mm} \includegraphics[width=0.16\linewidth,height=0.105\linewidth]{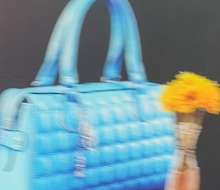}
            & \hspace{-4.0mm} \includegraphics[width=0.16\linewidth,height=0.105\linewidth]{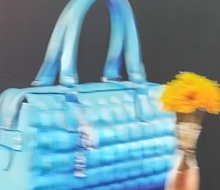}
            & \hspace{-4.0mm} \includegraphics[width=0.16\linewidth,height=0.105\linewidth]{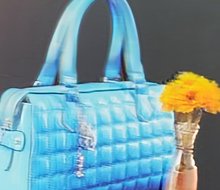}
            & \hspace{-4.0mm} \includegraphics[width=0.16\linewidth,height=0.105\linewidth]{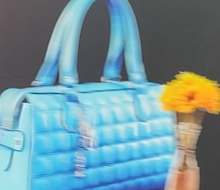}
              \\
    		\multicolumn{3}{c}{~}
            & \hspace{-4.0mm} (a) Blurred patch
            & \hspace{-4.0mm} (b) DBGAN
            & \hspace{-4.0mm} (c) FFTformer
            & \hspace{-4.0mm} (d) HI-Diff \\		
    	\multicolumn{3}{c}{~}
            & \hspace{-4.0mm} \includegraphics[width=0.16\linewidth,height=0.105\linewidth]{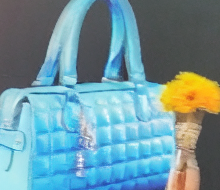}
            & \hspace{-4.0mm} \includegraphics[width=0.16\linewidth,height=0.105\linewidth]{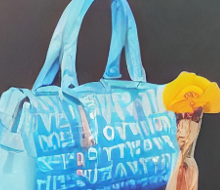}
            & \hspace{-4.0mm} \includegraphics[width=0.16\linewidth,height=0.105\linewidth]{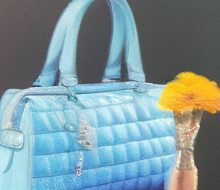}
            & \hspace{-4.0mm} \includegraphics[width=0.16\linewidth,height=0.105\linewidth]{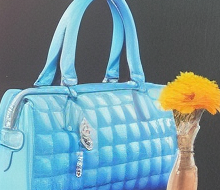}
            \\
    	\multicolumn{3}{c}{\hspace{-4.0mm} Blurred image from RWBI dataset}
            & \hspace{-4.0mm} (e) ResShift
            & \hspace{-4.0mm} (f) PASD
            & \hspace{-4.0mm} (g) DiffBIR
            & \hspace{-4.0mm} (h) Ours\\

    \end{tabular}
\vspace{-4mm}
\caption{Visual comparison with state-of-the-art image deblurring methods. (b) to (g) fail to preserve the input information well while deblurring. In contrast our method generates clear results while effectively retaining the input information.}
\label{fig: supp_result_1}
\vspace{-2mm}
\end{figure*}

\begin{figure*}[!t]
\footnotesize
\centering

    \begin{tabular}{c c c c c c c}
            \multicolumn{3}{c}{\multirow{5}*[45.6pt]{
            \hspace{-4mm} 
            \vspace{-25mm}
            \includegraphics[width=0.325\linewidth,height=0.236\linewidth]{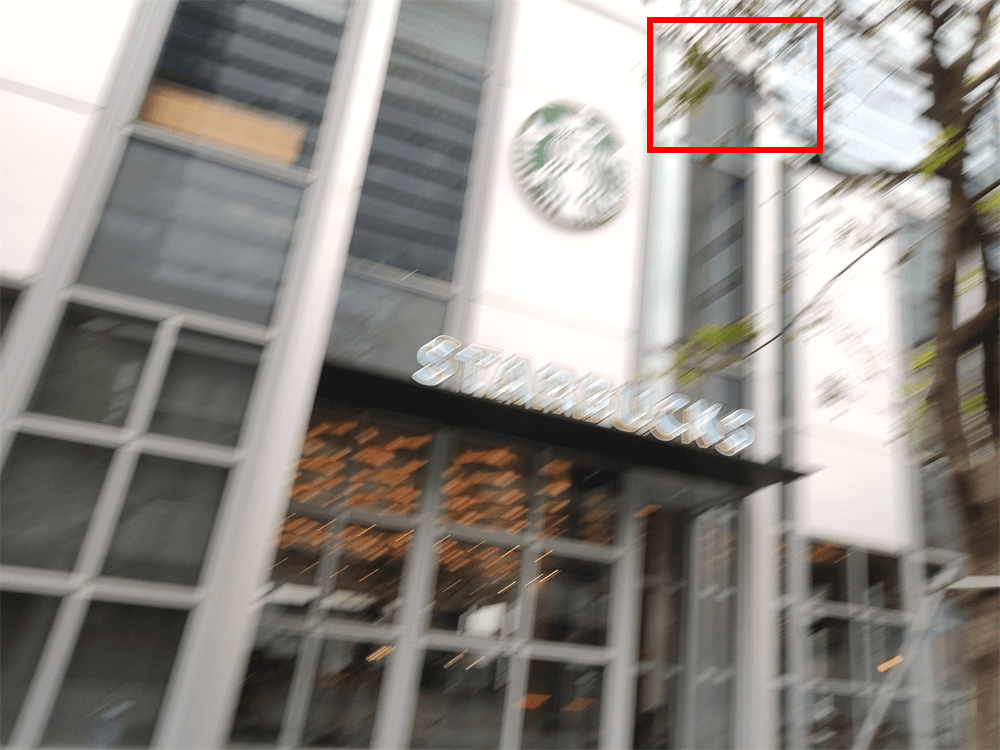}}}
            & \hspace{-4.0mm} \includegraphics[width=0.16\linewidth,height=0.105\linewidth]{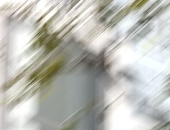}
            & \hspace{-4.0mm} \includegraphics[width=0.16\linewidth,height=0.105\linewidth]{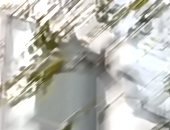}
            & \hspace{-4.0mm} \includegraphics[width=0.16\linewidth,height=0.105\linewidth]{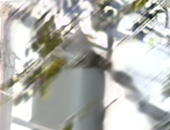}
            & \hspace{-4.0mm} \includegraphics[width=0.16\linewidth,height=0.105\linewidth]{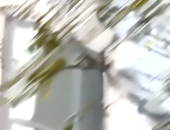}
              \\
    		\multicolumn{3}{c}{~}
            & \hspace{-4.0mm} (a) Blurred patch
            & \hspace{-4.0mm} (b) DBGAN
            & \hspace{-4.0mm} (c) FFTformer
            & \hspace{-4.0mm} (d) HI-Diff \\		
    	\multicolumn{3}{c}{~}
            & \hspace{-4.0mm} \includegraphics[width=0.16\linewidth,height=0.105\linewidth]{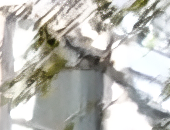}
            & \hspace{-4.0mm} \includegraphics[width=0.16\linewidth,height=0.105\linewidth]{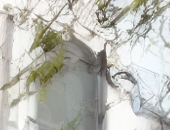}
            & \hspace{-4.0mm} \includegraphics[width=0.16\linewidth,height=0.105\linewidth]{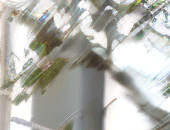}
            & \hspace{-4.0mm} \includegraphics[width=0.16\linewidth,height=0.105\linewidth]{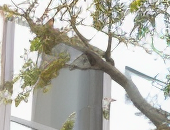}
            \\
    	\multicolumn{3}{c}{\hspace{-4.0mm} Blurred image from RWBI dataset}
            & \hspace{-4.0mm} (e) ResShift
            & \hspace{-4.0mm} (f) PASD
            & \hspace{-4.0mm} (g) DiffBIR
            & \hspace{-4.0mm} (h) Ours\\

    \end{tabular}
\vspace{-4mm}
\caption{Visual comparison with state-of-the-art image deblurring methods. he deblurred results in (b)-(g) still contain significant blur effects. The proposed method generates a clearer image, where the structure of the branches and the leaves are much clearer.}
\label{fig: supp_result_2}
\vspace{-2mm}
\end{figure*}

\begin{figure*}[!t]
\footnotesize
\centering
    \begin{tabular}{c c c c c c c}
            \multicolumn{3}{c}{\multirow{5}*[45.6pt]{
            \hspace{-4mm} 
            \vspace{-25mm}
            \includegraphics[width=0.325\linewidth,height=0.236\linewidth]{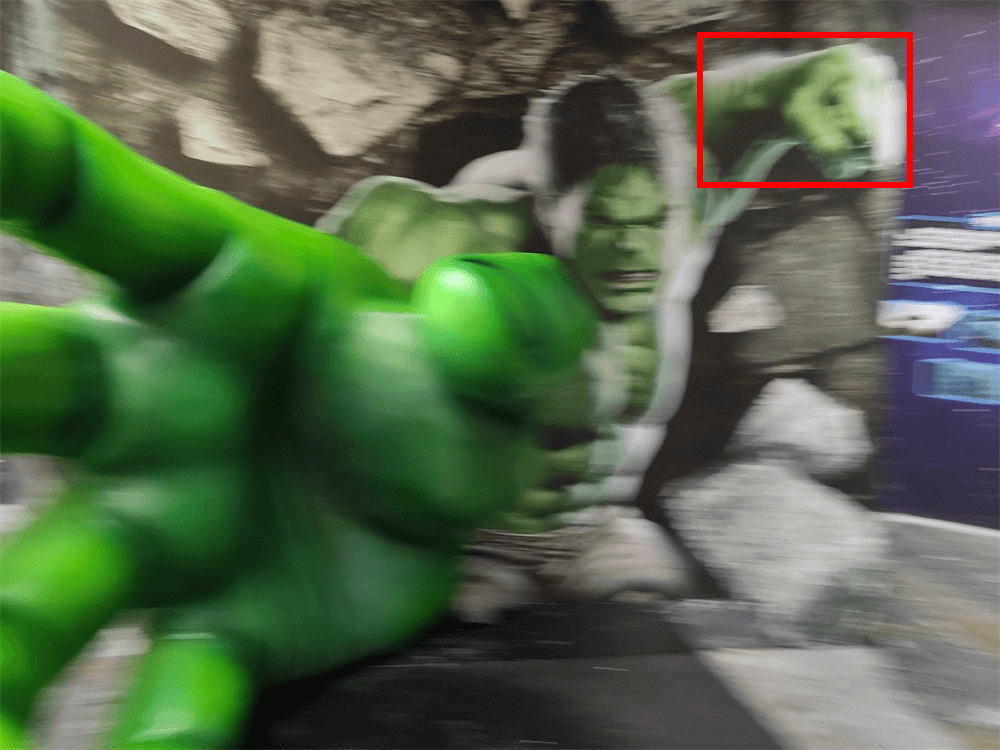}}}
            & \hspace{-4.0mm} \includegraphics[width=0.16\linewidth,height=0.105\linewidth]{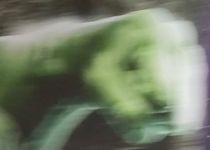}
            & \hspace{-4.0mm} \includegraphics[width=0.16\linewidth,height=0.105\linewidth]{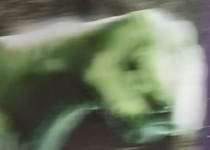}
            & \hspace{-4.0mm} \includegraphics[width=0.16\linewidth,height=0.105\linewidth]{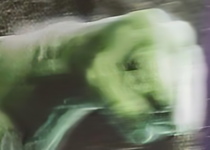}
            & \hspace{-4.0mm} \includegraphics[width=0.16\linewidth,height=0.105\linewidth]{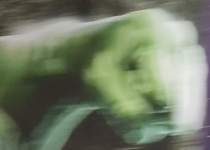}
              \\
    		\multicolumn{3}{c}{~}
            & \hspace{-4.0mm} (a) Blurred patch
            & \hspace{-4.0mm} (b) DBGAN
            & \hspace{-4.0mm} (c) FFTformer
            & \hspace{-4.0mm} (d) HI-Diff \\		
    	\multicolumn{3}{c}{~}
            & \hspace{-4.0mm} \includegraphics[width=0.16\linewidth,height=0.105\linewidth]{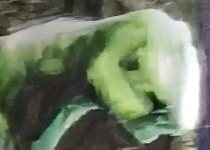}
            & \hspace{-4.0mm} \includegraphics[width=0.16\linewidth,height=0.105\linewidth]{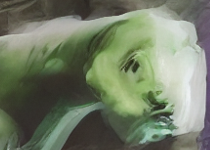}
            & \hspace{-4.0mm} \includegraphics[width=0.16\linewidth,height=0.105\linewidth]{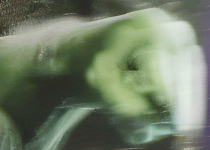}
            & \hspace{-4.0mm} \includegraphics[width=0.16\linewidth,height=0.105\linewidth]{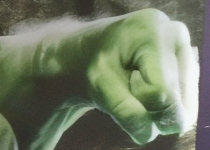}
            \\
    	\multicolumn{3}{c}{\hspace{-4.0mm} Blurred image from RWBI dataset}
            & \hspace{-4.0mm} (e) ResShift
            & \hspace{-4.0mm} (f) PASD
            & \hspace{-4.0mm} (g) DiffBIR
            & \hspace{-4.0mm} (h) Ours\\

    \end{tabular}
\vspace{-4mm}
\caption{Visual comparison with state-of-the-art image deblurring methods. (b) to (g) leave severe artifacts. In contrast, our method generates a clear and realistic result.}
\label{fig: supp_result_3}
\vspace{-2mm}
\end{figure*}

\begin{figure*}[!t]
\footnotesize
\centering
    \begin{tabular}{c c c c c c c}
            \multicolumn{3}{c}{\multirow{5}*[45.6pt]{
            \hspace{-4mm} 
            \vspace{-25mm}
            \includegraphics[width=0.325\linewidth,height=0.236\linewidth]{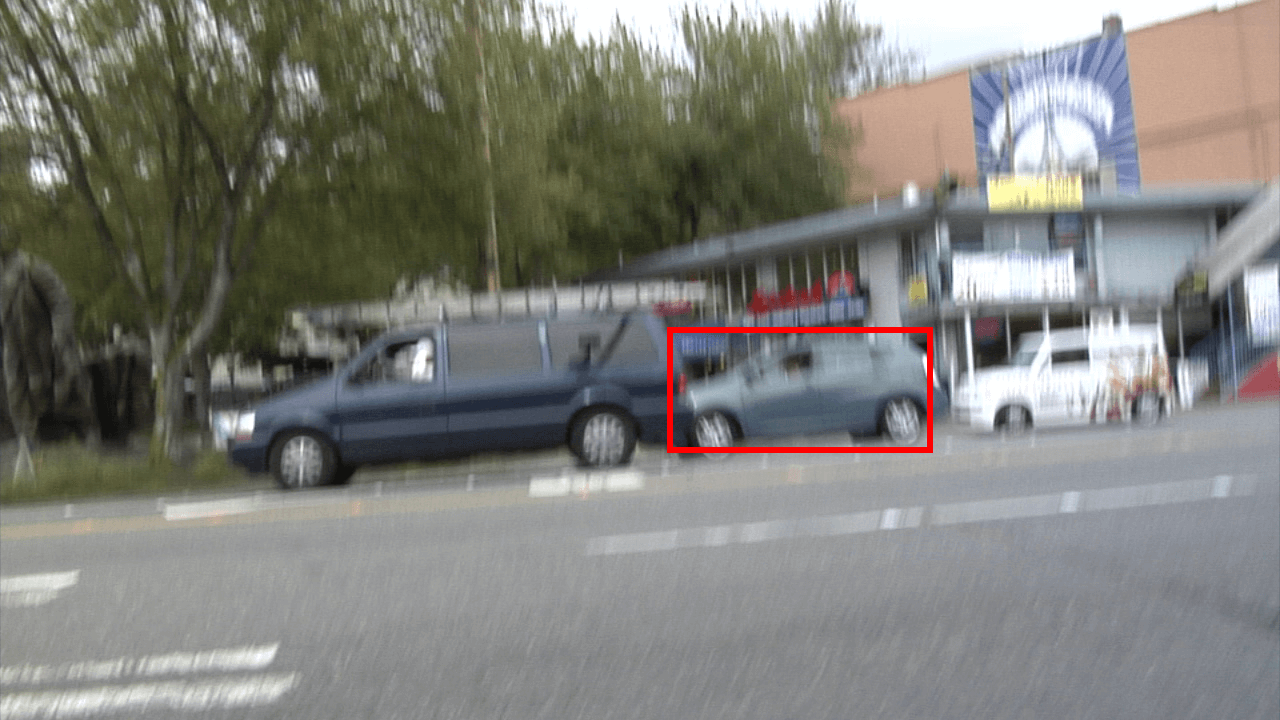}}}
            & \hspace{-4.0mm} \includegraphics[width=0.16\linewidth,height=0.105\linewidth]{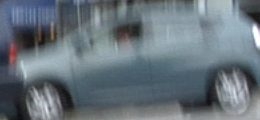}
            & \hspace{-4.0mm} \includegraphics[width=0.16\linewidth,height=0.105\linewidth]{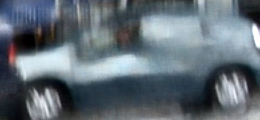}
            & \hspace{-4.0mm} \includegraphics[width=0.16\linewidth,height=0.105\linewidth]{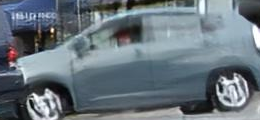}
            & \hspace{-4.0mm} \includegraphics[width=0.16\linewidth,height=0.105\linewidth]{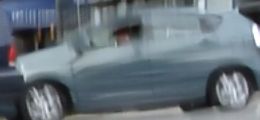}
              \\
    		\multicolumn{3}{c}{~}
            & \hspace{-4.0mm} (a) Blurred patch
            & \hspace{-4.0mm} (b) DBGAN
            & \hspace{-4.0mm} (c) FFTformer
            & \hspace{-4.0mm} (d) HI-Diff \\		
    	\multicolumn{3}{c}{~}
            & \hspace{-4.0mm} \includegraphics[width=0.16\linewidth,height=0.105\linewidth]{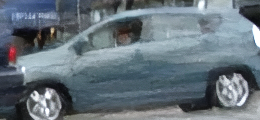}
            & \hspace{-4.0mm} \includegraphics[width=0.16\linewidth,height=0.105\linewidth]{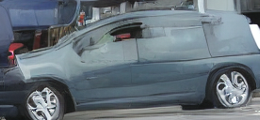}
            & \hspace{-4.0mm} \includegraphics[width=0.16\linewidth,height=0.105\linewidth]{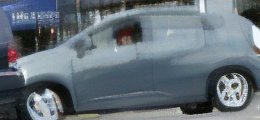}
            & \hspace{-4.0mm} \includegraphics[width=0.16\linewidth,height=0.105\linewidth]{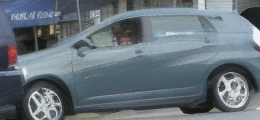}
            \\
    	\multicolumn{3}{c}{\hspace{-4.0mm} Blurred image from Real Blurry Images.}
            & \hspace{-4.0mm} (e) ResShift
            & \hspace{-4.0mm} (f) PASD
            & \hspace{-4.0mm} (g) DiffBIR
            & \hspace{-4.0mm} (h) Ours\\

    \end{tabular}
\vspace{-4mm}
\caption{Visual comparison with state-of-the-art image deblurring methods. The structures in the results restored by methods (b) to (g) are all distorted. In contrast, the structures in our results are clear and accurate.}
\label{fig: supp_result_4}
\vspace{-2mm}
\end{figure*}

In this section, we provide more visual comparisons of the proposed method with state-of-the-art ones on both synthetic and real-world benchmarks in ~\cref{fig: supp_result_1,fig: supp_result_2,fig: supp_result_3,fig: supp_result_4}  

\end{document}